\documentclass[conference]{IEEEtran}
\IEEEoverridecommandlockouts

\usepackage{cite}
\usepackage{amsmath,amssymb,amsfonts}
\usepackage{graphicx}
\usepackage{textcomp}
\usepackage{xcolor}
\usepackage{booktabs}
\usepackage{multirow}
\usepackage{array}
\usepackage{url}
\usepackage{enumitem}
\usepackage{bm}

\usepackage{algorithm}
\usepackage{algpseudocode}

\def\BibTeX{{\rm B\kern-.05em{\sc i\kern-.025em b}\kern-.08em
    T\kern-.1667em\lower.7ex\hbox{E}\kern-.125emX}}

\usepackage{orcidlink}
\usepackage[T1]{fontenc}
\usepackage[utf8]{inputenc}
\usepackage{microtype}
\usepackage{amsmath,amssymb,mathtools}
\usepackage{bm}
\usepackage{booktabs}
\usepackage{array}
\usepackage{multirow}
\usepackage{graphicx}
\usepackage{xcolor}
\usepackage{colortbl}
\usepackage{enumitem}
\usepackage{caption}
\usepackage{subcaption}
\usepackage{algorithm}
\usepackage{algpseudocode}
\usepackage{cite}
\usepackage{tikz}
\usepackage{pgfplots}
\pgfplotsset{compat=1.18}
\usetikzlibrary{arrows.meta,positioning,fit,backgrounds,shapes.geometric,
                decorations.pathreplacing,calc,matrix}
 
\definecolor{teachergreen}{RGB}{52,168,83}
\definecolor{studentblue}{RGB}{66,133,244}
\definecolor{pseudogold}{RGB}{251,188,4}
\definecolor{activered}{RGB}{234,67,53}
\definecolor{darkgray}{RGB}{60,60,60}
\definecolor{lightgraybox}{RGB}{240,242,245}
\definecolor{lossviolet}{RGB}{103,58,183}
 
\DeclareMathOperator{\KLdiv}{KL}
\DeclareMathOperator{\ReLU}{ReLU}
 
\providecommand{\Dl}{\mathcal{D}_l}
\providecommand{\Du}{\mathcal{D}_u}
\providecommand{\Pt}{P_t}
\providecommand{\Ps}{P_s}

\providecommand{\Aw}{\mathcal{A}_w}
\providecommand{\As}{\mathcal{A}_s}
\providecommand{\etal}{\textit{et~al.}}
\providecommand{\eg}{\textit{e.g.}}

\algrenewcommand\algorithmicrequire{\textbf{Input:}}
\algrenewcommand\algorithmicensure{\textbf{Output:}}
\usepackage{xcolor}
\usepackage{float}
\definecolor{myred}{RGB}{220,50,47}
\definecolor{mygreen}{RGB}{0,150,80}
\definecolor{myblue}{RGB}{38,139,210}
\definecolor{rowblue}{RGB}{232,242,255}
\definecolor{rowgreen}{RGB}{235,248,235}

\definecolor{goodgreen}{RGB}{0,150,80}
\definecolor{badred}{RGB}{200,50,50}
\begin{document}
\newcommand\HUGE{\fontsize{19.6}{25}\selectfont}

\title{\HUGE{Label Less, Learn More: Resource-Efficient Active Semi-Supervised Learning for Onboard Satellite Image Annotation}}

\author{\IEEEauthorblockN{\textsuperscript{1}Ahmed Abdelnaby, \textsuperscript{1}Mohamed Elmahallawy, \textsuperscript{2}Marius Bernahrndt, \textsuperscript{2}Tobias Hecking}
\IEEEauthorblockA{\textit{\textsuperscript{1} School of Engineering and Applied Science, Washington State University, Richland, WA 99354, USA} \\
\textit{\textsuperscript{2} Institute for Software Technology, German Aerospace Center (DLR), Porz, 51147 Cologne, Germany} \\
Email:\{ahmedabelnaby, mohamed.elmahallawy\}@wsu.edu, \{marius.bernahrndt, Tobias.Hecking\}@dlr.de}\vspace{-5mm}}

\maketitle

\begin{abstract}

Large-scale pervasive sensing increasingly relies on high-resolution satellite
imagery, yet task-specific {\em onboard vision} is constrained by costly
annotation and limited computation, memory, energy, and communication
resources. Existing approaches largely rely on either data-hungry supervised
learning or large vision-language foundation models, limiting efficient
adaptation and deployment under these constraints. We present {\sc SatLabel},
a resource-aware learning framework that transforms limited satellite labels
into progressively refined onboard models through \emph{adaptive sample
acquisition and semi-supervised model adaptation}. Rather than repeatedly
training on uniformly sampled labels, {\sc SatLabel} closes the loop between
model uncertainty, class imbalance, and pseudo-label quality to selectively
acquire informative samples while exploiting abundant unlabeled imagery.
This enables a compact student to adapt to target sensing domains with reduced
annotation and inference costs. We further introduce an optional
Mixture-of-Experts (MoE) student with graph-based feature refinement to enhance
representation capacity while retaining a lightweight footprint. We evaluate
{\sc SatLabel} on 11 remote-sensing datasets spanning core, extended, and
unseen domains against RemoteCLIP zero-shot inference. {\sc SatLabel} improves
Macro-F1 on most core and extended datasets while maintaining strong
cross-dataset transfer to unseen domains. More importantly, the Balanced
student contains only 11.2M parameters and occupies approximately 42.8~MB,
compared with 151.3M parameters and 577~MB for RemoteCLIP, while requiring
3.65 versus 5.89 GFLOPs. Across four efficiency benchmarks, it achieves
approximately $2\times$ higher GPU-forward throughput and reduces energy per
image on datasets. These results demonstrate that
{\sc SatLabel} enables accurate and resource-efficient satellite image
annotation and adaptation for constrained onboard deployment.


 \end{abstract}

\section{Introduction} \label{sec:intro}

High-resolution satellite imagery has become foundational for large-scale pervasive sensing applications, from environmental monitoring and precision agriculture to disaster response~\cite{karwowska2022using,li2024glh}. These domains demand timely, in-situ analysis of remotely sensed data. However, conventional workflows that downlink all raw imagery for ground-based processing face serious bottlenecks. Satellites have severely {\em limited resources} – including computing power, memory, and energy budgets – and only brief communication windows to ground stations (GSs) \cite{capogrosso2026tinymllunar,li2025grace}. For example, CubeSats and small LEO platforms simply cannot offload all collected data each pass, {\em motivating a shift towards onboard intelligence}. Recent research echoes this vision, treating LEO satellite constellations as a type of global edge network~\cite{elmahallawy2024stitching,li2025grace}, where processing is pushed to the ``edge'' (i.e.\ the satellites) rather than relying solely on GSs.  This paradigm transforms satellites from {\em passive data collectors} into {\em pervasive sensing and edge-computing platforms}, reducing raw-data downlink, communication overhead, and latency for time-sensitive applications.

Enabling task-specific onboard vision brings {\em two} key challenges. First, modern deep learning methods are typically data-hungry, yet satellites have very few labeled examples. Second, satellites have extremely constrained computation and power. State-of-the-art vision-language foundation models (e.g. RemoteCLIP \cite{liu2023remoteclip}) and large vision transformers (ViTs) remain challenging to deploy efficiently on resource-constrained satellite hardware due to their substantial computational and memory requirements~\cite{le2026onboard}. Recent works have begun to address these challenges by compressing or distributing learning. For instance, Le \etal{} \cite{le2024semantic} distill two large ViT teachers into a tiny ResNet-8 student, achieving over 90\% accuracy while using 97.5\% fewer parameters and 86\% less power. Similarly, Östman \etal{} \cite{ostman2023decentralised} demonstrate that a distributed semi-supervised learning scheme can train a constellation of satellites to $\sim91\%$ classification accuracy on the EuroSAT dataset within one day. These results show that on-board learning is feasible even under stringent resource limits.

Nevertheless, existing approaches have critical limitations for practical satellite deployment. Many methods assume uniform data sampling or fixed model architectures, ignoring the rich feedback available onboard. Uniform sampling wastes precious annotation budget on easy examples, while static student models cannot adapt their capacity when new types of images arrive. By contrast, vision-language models (VLMs) like RemoteCLIP \cite{liu2023remoteclip} achieve broad generalization but {\em require billions of parameters and extensive fine-tuning}, making them {\em impractical} for satellites. Even recent satellite–ground collaborative systems (e.g.\ Grace \cite{li2025grace}) that deploy small VLMs on satellites {\em must rely on the ground for heavy lifting}, highlighting the persistent resource gap. TinyML efforts (e.g.\ the $\Phi$-Sat-1 mission) focus on compressing convolutional neural networks (CNNs)  to save downlink (e.g.\ removing cloudy images in orbit saved $\approx$30\% of data \cite{capogrosso2026tinymllunar}), but {\em do not address how to select or label new data autonomously}. In short, prior work overlooks the combined problem of {\em which satellite images to label and how to update a model under strict onboard constraints.}

To address these gaps, we propose {\sc SatLabel}, a resource-aware learning framework for satellite edge intelligence. {\sc SatLabel} closes the loop between {\em data acquisition and model adaptation:} it uses the current model’s uncertainty and the dataset’s class balance to actively select informative samples for labeling, and then performs semi-supervised adaptation using both these labels and the remaining unlabeled images. This adaptive strategy ensures that each annotation maximally improves the model, avoiding redundant labeling of easy or over-represented examples. Furthermore, {\sc SatLabel}’s student model is compact and adaptive: we introduce a Mixture-of-Experts (MoE) architecture with graph-based feature refinement to {\em boost representation power without bloating computation}. In our evaluation on 11 remote-sensing datasets (including core, extended, and unseen domains), {\sc SatLabel} significantly improves classification performance (e.g.\ macro-F1) over baseline models, while using a {\em fraction of the resources}. In particular, our adaptive student runs $\approx$2$\times$ faster and has $\approx$13.5$\times$ fewer parameters than RemoteCLIP, making on-board annotation and model refinement far more practical under satellite constraints. These results demonstrate that {\sc SatLabel} advances pervasive satellite computing by enabling autonomous on-board image labeling and learning in a highly resource-aware manner. We summarize our key contributions as follows:
 
 \begin{itemize}[leftmargin=*]

\item We introduce {\sc SatLabel}, the first onboard semi-supervised learning framework that adaptively acquires labels from limited satellite data and refines a model with unlabeled imagery, all under stringent satellite resource constraints.

\item We develop a closed-loop sample-selection strategy that jointly exploits model uncertainty, class imbalance, and pseudo-label quality to acquire the most informative samples for onboard adaptation.

\item We design a compact adaptive student with MoE routing and graph-based feature refinement to improve representation quality under a limited compute budget, leveraging multiple experts for richer features while keeping {\em inference lightweight for resource-constrained onboard deployment}.

\item We evaluate {\sc SatLabel} across 11 remote-sensing datasets spanning
core, extended, and unseen domains, where it achieves competitive or higher
Macro-F1 than state-of-the-art remote-sensing models on most datasets.
The Balanced student uses approximately $13.5\times$ fewer parameters,
requires $38\%$ fewer FLOPs, and delivers approximately $2\times$ higher
GPU-forward throughput than RemoteCLIP, while reducing energy consumption per image on three of four efficiency benchmarks. These gains support resource-constrained onboard image annotation and adaptation.


\end{itemize}

\section{Related Work} \label{sec:RW}

Earth-observation satellites generate massive imagery volumes, but costly labeling and stringent onboard resource constraints hinder continuous model adaptation. Existing approaches address foundation models, semi-supervised learning (SSL), active learning (AL), and distillation largely separately, whereas {\sc SatLabel} integrates these capabilities for \emph{label-efficient and resource-efficient onboard learning}.

\noindent{\underline{\bf \em Remote sensing scene classification.}}
Early remote-sensing scene classification methods relied on hand-crafted representations, including bag-of-visual-words features~\cite{yang2010bag}. The emergence of deep CNNs substantially improved feature learning and classification performance on remote-sensing benchmarks~\cite{cheng2017remote}, while more recent ViTs have further improved the ability to capture long-range spatial and semantic relationships in satellite imagery~\cite{roy2023multimodal}.
Despite these advances, conventional supervised approaches still rely on large manually labeled datasets, making adaptation increasingly costly as satellite archives expand across new regions, conditions, and categories. Thus, although these models achieve strong classification performance, {\em they do not address the recurring labeling cost required to adapt to newly acquired imagery.}

\vspace{1mm}
\noindent{\underline{\bf \em Foundation models for remote sensing.}}
VLMs offer an alternative by transferring semantic knowledge learned from large-scale image-text corpora to previously unseen visual concepts. CLIP~\cite{radford2021learning} demonstrated that aligning visual and textual representations enables effective zero-shot recognition without task-specific labeled training data. Building on this paradigm, RemoteCLIP~\cite{liu2024remoteclip} adapts CLIP to remote-sensing imagery through domain-specific pre-training, substantially improving zero-shot recognition of aerial scenes. Other efforts, including RS-CLIP~\cite{li2023rs} and GeoRSCLIP~\cite{zhang2024rs5m}, further specialize vision-language features for Earth observation using large remote-sensing data, pseudo-labeling, and geospatial information.

These models substantially reduce the dependence on task-specific annotation, but their primary role remains \emph{recognition}: they provide predictions for individual images rather than a {\em mechanism for efficiently adapting a lightweight onboard model} to a continuously evolving satellite data stream. Moreover, directly deploying a large foundation model onboard can be prohibitively expensive under satellite compute and memory constraints. {\sc SatLabel} instead exploits a remote-sensing vision-language model as a \emph{frozen semantic teacher}, using its transferable knowledge to supervise adaptation of a substantially smaller student model. This formulation separates the semantic capability of the foundation model from the resource requirements of the onboard classifier.

\vspace{1mm}
\noindent{\underline{\bf \em Semi-supervised learning.}} SSL reduces annotation requirements by exploiting large collections of unlabeled data together with a small labeled set. FixMatch~\cite{sohn2020fixmatch}, for example, combines confidence-based pseudo-labeling with weak-to-strong consistency regularization, while Mean Teacher~\cite{tarvainen2017mean} improves pseudo-label stability through exponential moving-average (EMA) teacher updates. Such approaches are attractive for satellite learning because the vast majority of newly acquired imagery can be exploited without requiring manual annotation.

However, conventional SSL methods typically derive pseudo-labels from the model being trained or from an evolving EMA teacher. Consequently, pseudo-label quality can be limited during the early stages of adaptation, precisely when the student has little task-specific knowledge. {\sc SatLabel} addresses this limitation by obtaining initial semantic supervision from a pretrained and frozen remote-sensing foundation model rather than relying exclusively on self-generated predictions. The resulting supervision provides a stronger starting point for adaptation while allowing the student to progressively learn from the much larger unlabeled satellite archive.

\vspace{1mm}
\noindent{\underline{\bf \em Active learning for remote sensing.}}
AL complements SSL by addressing a different question: rather than labeling data indiscriminately, \emph{which} samples should receive human annotation? Existing remote-sensing studies have investigated uncertainty-based sampling~\cite{cacciarelli2024active} and query-by-committee strategies for selecting informative samples from hyperspectral~\cite{thoreau2022active} and RGB satellite imagery. These approaches can substantially reduce annotation effort by prioritizing samples where the current model is uncertain or where competing models disagree.

For onboard satellite learning, however, sample selection is especially important because transmitting every newly acquired image to the GSs for annotation is {\em constrained by communication bandwidth and energy}. AL alone does not provide a mechanism for exploiting the remaining unlabeled imagery, while SSL alone does not determine which samples are most valuable for human supervision. {\sc SatLabel} combines the two: {\em uncertainty-driven querying identifies informative images for labeling, while foundation-model-guided semi-supervised adaptation exploits the remaining unlabeled data}, enabling the annotation budget to be concentrated on samples that provide the greatest value for improving the onboard model.

\vspace{1mm}
\noindent{\underline{\bf \em Knowledge distillation (KD).}} It provides a complementary mechanism for transferring the capabilities of large models to resource-efficient models. Hinton~\etal~\cite{hinton2015distilling} showed that soft predictions from a larger teacher encode inter-class relationships that can be transferred to a compact student. Subsequent approaches, such as FitNets~\cite{dietmuller2024fitnets}, extended distillation to intermediate representations, while recent Earth-observation studies have applied KD using supervised ViT teachers~\cite{le2025semantic}.

Unlike these approaches, {\sc SatLabel} uses a \emph{zero-shot remote-sensing VLM} as the {\em teacher} rather than a {\em task-specific classifier} trained on the target label space. This distinction is important for satellite adaptation: the teacher can provide transferable semantic supervision without requiring retraining whenever the target scene categories or deployment environment changes. More importantly, distillation in {\sc SatLabel} is integrated with active and semi-supervised learning rather than used solely as a one-time model compression step. The student therefore acquires semantic knowledge from the foundation model while progressively adapting to locally acquired satellite imagery, yielding a compact model suitable for resource-constrained onboard inference and continued learning.

 \section{Methodology} \label{sec:Method}



We present {\sc SATLabel}, a resource-aware active semi-supervised framework that enables satellites to autonomously annotate their continuously acquired imagery, providing locally generated training data for downstream onboard AI learning. The framework addresses the key challenge of performing this process under stringent onboard computation, memory, energy, and communication constraints, where transmitting large image volumes to GSs for annotation can be costly or impractical. To achieve this goal, {\sc SATLabel} addresses three coupled challenges: \emph{obtaining reliable supervision from few labels, adapting a lightweight model for onboard deployment, and efficiently allocating scarce human annotation effort to the most informative imagery}.

\begin{figure*}[t]
  \centering\vspace{-2mm}
  \includegraphics[width=0.8\linewidth]{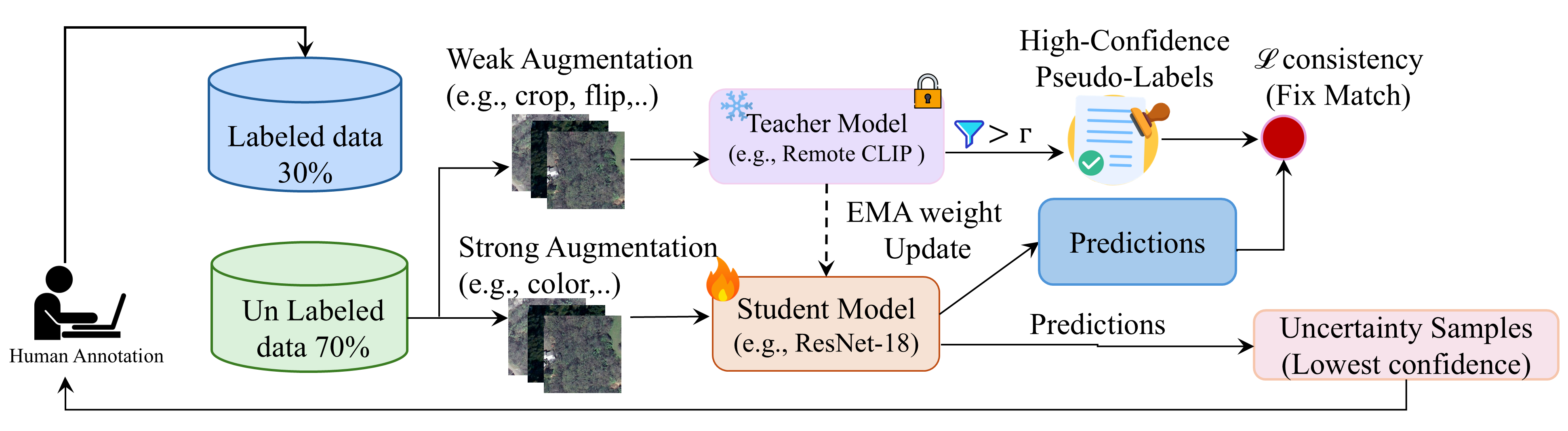}
\caption{Overview of the proposed {\sc SatLabel} methodology. The teacher--student framework integrates pseudo-labeling, consistency regularization, and uncertainty-guided active learning to progressively improve the student model.}

  \label{fig:framework}
\end{figure*}

As illustrated in Fig.~\ref{fig:framework}, {\sc SATLabel} innovates by offering three interconnected components. First, a frozen remote-sensing foundation model acts as a \emph{teacher}, providing semantic guidance for unlabeled imagery and reducing reliance on manual labels. This knowledge is transferred to a \emph{compact student} through semi-supervised adaptation, yielding a lightweight model suitable for resource-constrained onboard deployment. To offset its reduced capacity, an \emph{adaptive feature-refinement module} enriches the student's representations through expert specialization and inter-sample relationships. Finally, \emph{uncertainty-guided active acquisition} selects informative unlabeled samples for human annotation, which are added to the training pool for subsequent adaptation. This closed learning loop progressively improves the onboard model while jointly reducing annotation demand and computational overhead. The following subsections detail each component of the framework.

\subsection{Teacher-Guided Semi-Supervised Learning}
\label{sec:teacher}

{\sc SatLabel} employs a frozen {RemoteCLIP}~\cite{liu2024remoteclip} as a semantic teacher to provide stable supervision for the lightweight student. This design avoids relying solely on potentially unreliable student predictions when only a small labeled set is available. For each unlabeled image $x \in \Du$, {\sc SatLabel} generates two views using weak and strong augmentations:
\begin{equation}
x^{w} = \Aw(x), \qquad
x^{s} = \As(x),
\end{equation}
where $\Aw(\cdot)$ applies semantics-preserving transformations such as random cropping and flipping, while $\As(\cdot)$ introduces stronger appearance variations, including color jittering, blur, and random erasing (Fig.~\ref{fig:framework}). The weak view is used by the frozen teacher to obtain semantic supervision, whereas the strongly augmented view is used to train the student to preserve the same semantics under appearance variations.

\begin{figure*}[t]
  \centering
  \includegraphics[width=0.8\linewidth]{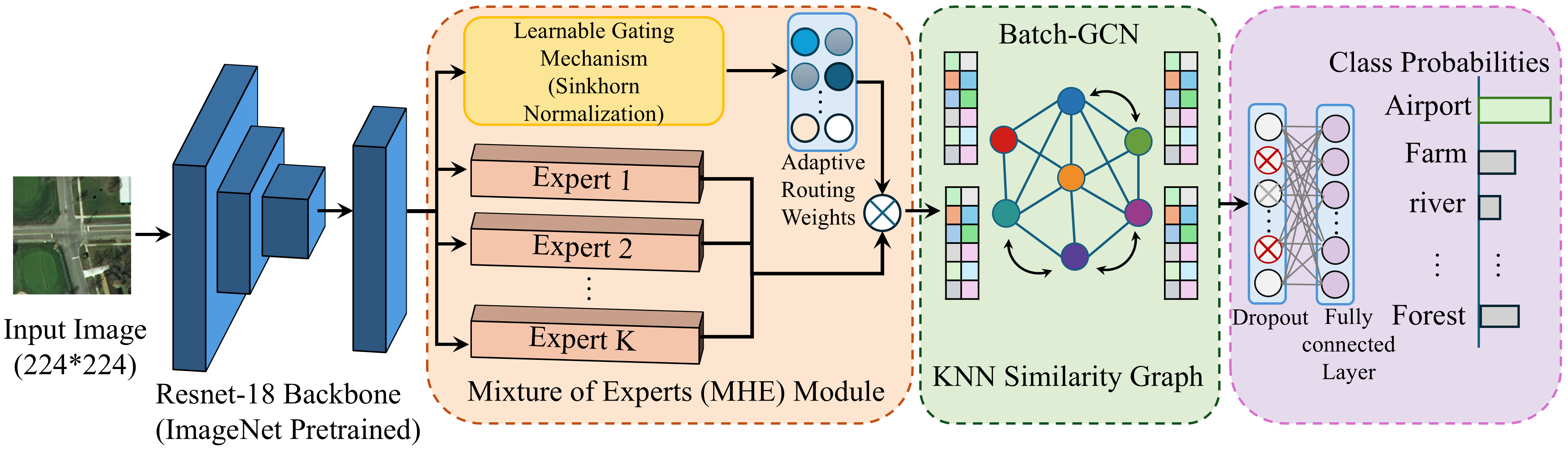}
\caption{An illustration of the adaptive student model architecture. The ResNet-18 backbone is augmented with a Mixture-of-Heads module, adaptive routing, and BatchGCN for efficient feature refinement, followed by a classification head for final prediction.}

  \label{fig:student}\vspace{-3mm}
\end{figure*}

\vspace{1mm}
\noindent\underline{{\bf\em Frozen teacher supervision.}}
Given a weakly augmented image $x^{w}$, RemoteCLIP computes an image embedding
$v(x^{w})$ and compares it with the text embedding $\bm{t}_{c}$ of each target
class $c \in \{1,\ldots,C\}$. The teacher produces a soft class distribution:
\begin{equation}
P_t(y=c \mid x^{w})
=
\frac{
\exp\!\left(\cos\!\left(v(x^{w}),\bm{t}_{c}\right)/T_t\right)
}{
\sum_{j=1}^{C}
\exp\!\left(\cos\!\left(v(x^{w}),\bm{t}_{j}\right)/T_t\right)
},
\label{eq:teacher}
\end{equation}
where $T_t$ denotes the teacher temperature. Freezing the teacher provides
stable supervision throughout training, avoids back-propagation through its
approximately 307M parameters, and preserves its zero-shot capability.
Adapting to a new target label space therefore requires only changing the
class prompts rather than retraining the teacher. Moreover, the resulting
soft distribution captures relative semantic relationships among classes
that are not represented by one-hot labels.
\vspace{1mm}


\noindent\underline{{\bf\em Soft-target KD for student adaptation.}} The semantic knowledge encoded by the frozen teacher is transferred to a compact student model, such as \textit{ResNet-18}, through KD and consistency regularization. Using a lightweight student is essential for onboard deployment, as directly executing a large model such as RemoteCLIP can impose substantial computational, memory, and energy overhead on resource-constrained satellite platforms. The student approximates the teacher's class posterior distribution, retaining its semantic knowledge while substantially reducing inference cost.

For an input $x^{w}$, we compute temperature-scaled teacher and student distributions as
\begin{equation}
\mathcal{L}_{\mathrm{KD}}
=
T^{2}
\operatorname{KL}
\left(
\Pt(y \mid x^{w};T)
,\middle|,
\Ps(y \mid x^{w};T)
\right),
\label{eq:lkd}
\end{equation}
where $T$ denotes the distillation temperature. Temperature scaling produces a softer probability distribution, exposing relative similarities among non-dominant classes and providing richer supervision than hard one-hot targets. For example, the student can learn that a \emph{forest} scene is more closely related to \emph{farmland} than to an \emph{airport}. We assign a larger weight to the distillation objective than to the supervised classification loss, i.e., $\lambda_{\mathrm{KD}} > \lambda_{\mathrm{CE}}$, to emphasize semantic knowledge transferred from the foundation-model teacher.

\vspace{1mm}
\noindent\underline{{\bf\em Pseudo-labeling and consistency regularization.}}
In addition to soft supervision, {\sc SatLabel} generates hard pseudo-labels from confident teacher predictions. An unlabeled sample is selected when
\begin{equation}
\max_{y} \Pt(y \mid x^{w}) > \tau,
\end{equation}
and its pseudo-label is given by
\begin{equation}
\hat{y}
=
\arg\max_{y} \Pt(y \mid x^{w}).
\end{equation}
These pseudo-labels enlarge the effective training set without requiring additional human annotation. For each selected sample, the student is trained on the strongly augmented view $x^{s}$ to match the teacher's prediction:
\begin{equation}
\mathcal{L}_{\mathrm{cons}}
=
\mathbb{E}*{x \in \Du'}
\left[
H
\left(
\Pt(y \mid x^{w}),
\Ps(y \mid x^{s})
\right)
\right],
\label{eq:lcons}
\end{equation}
where
\begin{equation}
\Du'={x \in \Du :
\max_{y} \Pt(y \mid x^{w}) > \tau}.
\end{equation}
This FixMatch-style consistency objective~\cite{sohn2020fixmatch} encourages the student to learn semantic features that remain invariant to substantial appearance changes.  To avoid reinforcing unstable or accidentally correct predictions during early adaptation, the consistency objective is activated from round~3, where each round denotes one complete AL iteration consisting of sample selection, annotation/pseudo-labeling, and student model update. By this stage, the student has undergone two rounds of supervised and teacher-guided training, providing more reliable predictions for consistency regularization.

\vspace{1mm}
\noindent\underline{{\bf\em EMA stabilization.}}
To further stabilize the evolving student during iterative adaptation, {\sc SatLabel} maintains an EMA copy of the student~\cite{tarvainen2017mean}:
\begin{equation}
\theta_{\mathrm{EMA}}
\leftarrow
\alpha \theta_{\mathrm{EMA}}
+
(1-\alpha)\theta_{S},
\qquad
\alpha = 0.999.
\label{eq:ema}
\end{equation}
Gradients propagate only through the online student parameters $\theta_{S}$. The EMA parameters are used as a temporally smoothed version of the student for prediction filtering, reducing sensitivity to short-term fluctuations during iterative learning.


\subsection{Adaptive Student Architecture}\label{sec:student}

This stage enhances the representational capacity of the lightweight student while preserving its suitability for resource-constrained onboard deployment. We first employ a ResNet-18 backbone~\cite{he2016deep} as the \emph{base student} because of its compact architecture and efficient inference. Although this lightweight backbone is attractive for onboard deployment, its limited capacity may hinder discrimination among the {\em visually diverse and semantically similar scenes} common in satellite imagery.

To address this limitation, {\sc SatLabel} introduces an \emph{adaptive feature-refinement module} composed of two complementary components: \emph{(i) a MoE module} that adaptively specializes feature transformation across heterogeneous scene characteristics, and \emph{(ii) a Batch Graph Convolutional Network (BatchGCN)} that exploits feature-space relationships among samples within a mini-batch. The MoE increases the effective representational capacity through expert specialization, while BatchGCN further refines these features by propagating information across related samples. Together, these modules enhance the base student's representations without requiring a substantially larger backbone. The resulting \emph{adaptive student} is shown in Fig.~\ref{fig:student}. Given an input image $x$, its feature transformation is expressed as:

\begin{equation}
x
\xrightarrow{f_{\mathrm{bb}}}
z
\xrightarrow{f_{\mathrm{MoE}}}
z_{\mathrm{MoE}}
\xrightarrow{f_{\mathrm{GCN}}}
z'
\xrightarrow{f_{\mathrm{cls}}}
p(y \mid x),
\label{eq:student_architecture}
\end{equation}
where $z$ denotes the ResNet-18 backbone representation, $z_{\mathrm{MoE}}$ is the expert-refined representation, and $z'$ is the relationally refined feature provided to the classification head.

 \vspace{1mm}

\noindent\underline{{\bf\em MoE with Sinkhorn routing.}} To enable adaptive representation learning for satellite imagery, where scenes exhibit substantial visual heterogeneity and differ in spatial structures, textures, and semantic characteristics, we employ a MoE layer comprising $K$ learnable expert networks. Given the backbone feature $z$, a gating network produces routing scores that determine which experts process the input. The resulting representation is
\begin{equation}
z_{\mathrm{MoE}}
=
\sum_{k=1}^{K} g_k(z),E_k(z),
\label{eq:feature}
\end{equation}
where $E_k(\cdot)$ denotes the $k$-th expert and $g_k(z)$ is its corresponding routing weight.

A conventional softmax gate may assign a disproportionately large fraction of samples to a small subset of experts, leading to \emph{expert collapse} and inefficient utilization of the available capacity. {\sc SatLabel} therefore applies Sinkhorn--Knopp normalization~\cite{cuturi2013sinkhorn,peyre2019computational} to the batch-level routing scores. The resulting approximately doubly stochastic assignment encourages a more balanced allocation of samples across experts while preserving input-dependent routing. This allows different experts to specialize in distinct visual characteristics without permitting a few experts to dominate training. To maintain computational efficiency, only the top-2 experts selected by the balanced routing mechanism are activated for each sample.
 

\vspace{1mm}
\noindent\underline{{\bf\em Batch Graph Convolution for Relational Refinement.}}
While the MoE module provides input-adaptive feature extraction, it processes samples independently and therefore does not explicitly exploit relationships among images observed in the same training batch. This is suboptimal for remote-sensing scenes, where visually related categories can exhibit similar textures, shapes, or spatial layouts. For example, cropland, grassland, and farmland may share similar texture patterns, whereas airports and industrial areas may exhibit related geometric structures.

To capture such inter-sample relationships, {\sc SatLabel} constructs a $k$-nearest-neighbor (KNN) similarity graph over the MoE features within each mini-batch. Each sample is treated as a graph node, and edges connect visually similar samples according to their feature-space similarity. Let $Z\in\mathbb{R}^{B\times d}$ denote the matrix of MoE features for a batch of $B$ samples and $A$ denote the resulting adjacency matrix. BatchGCN performs graph-based feature propagation using
\begin{equation}
Z'
=
\operatorname{ReLU}
\left(
\hat{A} Z W
\right),
\qquad
\hat{A}
=
\tilde{D}^{-\frac{1}{2}}
\tilde{A}
\tilde{D}^{-\frac{1}{2}},
\label{eq}
\end{equation}
where $\tilde{A}=A+I$ adds self-loops, $\tilde{D}$ is the corresponding degree matrix, and $W$ is a learnable projection matrix. This operation allows each sample to incorporate complementary information from its nearest neighbors while retaining its own representation. We use a single graph-convolution layer to provide relational refinement while limiting additional computation and avoiding the over-smoothing associated with deeper graph propagation~\cite{li2018deeper}.

The refined representation $Z'$ is subsequently regularized with dropout and passed through a lightweight fully connected classification head to obtain the student's prediction $\Ps(y\mid x)$. In this design, the MoE captures \emph{heterogeneous visual semantics} through adaptive expert specialization, whereas BatchGCN captures \emph{relational semantics} by propagating information among visually related samples. Together, they increase the representational capacity of the compact student while preserving its suitability for resource-constrained onboard deployment.



\subsection{Uncertainty-Guided Active Sample Acquisition} \label{sec:al}

To efficiently allocate the limited human-annotation budget, {\sc SatLabel} employs uncertainty-guided AL after each student adaptation round. Rather than querying samples uniformly from the unlabeled pool, the framework prioritizes images for which the current student exhibits the greatest predictive uncertainty. This strategy concentrates annotation effort on ambiguous or difficult examples that are more likely to expose weaknesses in the current decision boundary and provide useful supervision for subsequent model updates.

For each unlabeled sample $x\in\Du^{r}$, the student computes a least-confidence score based on its maximum posterior probability:
\begin{equation}
c(x)=\max_{y}\Ps(y\mid x),
\label{eq:confidence}
\end{equation}
where $\Ps(y\mid x)$ is the predicted probability of class $y$. A lower $c(x)$ indicates greater uncertainty, as the model assigns less probability mass to its most likely class. Given an annotation budget of $B$ samples per round, {\sc SatLabel} selects the subset with the smallest confidence scores:
\begin{equation}
\mathcal{S}^{r}=
\underset{
\mathcal{S}\subseteq\Du^{r},;|\mathcal{S}|=B
}{\arg\min}
\sum_{x\in\mathcal{S}} c(x).
\label{eq:query}
\end{equation}
where $\mathcal{S}^{r}$ denotes the set of samples selected for annotation at round $r$. Low-confidence samples are typically associated with ambiguous or difficult instances for the current decision boundary and therefore provide greater potential value for model refinement than samples that are already classified with high confidence~\cite{cacciarelli2024active}. After annotation, $\mathcal{S}^{r}$ is removed from the unlabeled pool and incorporated into the labeled training set for the subsequent round. This criterion requires only a forward pass through the lightweight student and therefore introduces minimal computational overhead, which is important for onboard deployment.

After human annotation, the queried samples are transferred from the unlabeled pool to the labeled set:
\begin{equation}
\Dl^{r+1}
=
\Dl^{r}\cup\mathcal{S}^{r},
\qquad
\Du^{r+1}
=
\Du^{r}/ \mathcal{S}^{r},
\label{eq:update}
\end{equation}
The expanded labeled set is then used in the subsequent adaptation round. Thus, active acquisition progressively replaces a small number of unlabeled, uncertain samples with ground-truth supervision, enabling {\sc SatLabel} to improve the onboard student while avoiding the substantially higher annotation cost of labeling the entire imagery stream.

\subsection{Joint Optimization}
\label{sec:overall}

The preceding components are jointly integrated to optimize the lightweight student. Specifically, the student is trained using three complementary supervisory signals: {\em ground-truth and high-confidence pseudo-labels through $\mathcal{L}_{\mathrm{CE}}$, semantic knowledge transferred from the frozen RemoteCLIP teacher through $\mathcal{L}_{\mathrm{KD}}$, and consistency between weakly and strongly augmented views through $\mathcal{L}_{\mathrm{cons}}$}. The resulting objective is
\begin{equation} 
\mathcal{L} = \lambda_{\mathrm{CE}}\,\mathcal{L}_{\mathrm{CE}} + \lambda_{\mathrm{KD}}\,\mathcal{L}_{\mathrm{KD}} + \lambda_{\mathrm{cons}}\,\mathcal{L}_{\mathrm{cons}}, \label{eq:joint} 
\end{equation} 
where $\mathcal{L}_{\mathrm{CE}}$ anchors learning to human and pseudo-labels, $\mathcal{L}_{\mathrm{KD}}$ transfers semantic knowledge from the teacher, and $\mathcal{L}_{\mathrm{cons}}$ enforces prediction consistency across different augmentation strengths. The weights $\lambda_{\mathrm{CE}}$, $\lambda_{\mathrm{KD}}$, and $\lambda_{\mathrm{cons}}$ control the contributions of supervised/pseudo-label learning, teacher knowledge transfer, and consistency regularization, respectively. The RemoteCLIP teacher remains frozen throughout all training rounds, while only the student parameters are updated.

This joint objective enables {\sc SatLabel} to simultaneously exploit labeled and unlabeled imagery, transfer rich semantic knowledge from the foundation model, and improve robustness to appearance variations within a single lightweight student model. The complete {\sc SatLabel} framework is illustrated in Fig.~\ref{fig:framework} and summarized in Algorithm~\ref{alg:training}.



\begin{algorithm}[t]
\small
\caption{{\sc SatLabel} Training}
\label{alg:training}
\begin{algorithmic}[1]
\Require $\Dl^0$, $\Du^0$, frozen teacher $f_T$, rounds $R$, budget $B$
\Ensure Trained student $f_s$

\State Initialize $f_s$ (ResNet-18$\to$MoE$\to$BatchGCN);
       $\theta_{\mathrm{EMA}}\!\leftarrow\!\theta_S$

\For{$r = 1$ \textbf{to} $R$}

  \State \textbf{Teacher:}
  $\Pt(y|x^w)\!\leftarrow\!f_T(\Aw(x))$,
  $\forall\,x\!\in\!\Du^r$
  \hfill\Comment{Eq.~\eqref{eq:teacher}}

  \State \textbf{Distillation:}
  minimize $\mathcal{L}_{\mathrm{KD}}$
  \hfill\Comment{Eq.~\eqref{eq:lkd}}

  \If{$r \geq 3$}
    \State \textbf{Pseudo-labeling:}
    construct $\Du'=\{x\in\Du^r:
    \max_y \Pt(y|x^w)>\tau\}$

    \State \textbf{Consistency:}
    minimize $\mathcal{L}_{\mathrm{cons}}$ on $\Du'$
    \hfill\Comment{Eq.~\eqref{eq:lcons}}
  \EndIf

  \State \textbf{Student update:}
  minimize $\mathcal{L}=
  \lambda_{\mathrm{CE}}\mathcal{L}_{\mathrm{CE}}+
  \lambda_{\mathrm{KD}}\mathcal{L}_{\mathrm{KD}}+
  \lambda_{\mathrm{cons}}\mathcal{L}_{\mathrm{cons}}$
  \hfill\Comment{Eq.~\eqref{eq:joint}}

  \State \textbf{EMA:}
  $\theta_{\mathrm{EMA}}\!\leftarrow\!
  \alpha\theta_{\mathrm{EMA}}+(1-\alpha)\theta_S$
  \hfill\Comment{Eq.~\eqref{eq:ema}}

  \State \textbf{Query:}
  select $\mathcal{S}^r\subset\Du^r$ using uncertainty;
  obtain human labels
  \hfill\Comment{Eq.~\eqref{eq:query}}

  \State \textbf{Update sets:}
  $\Dl^{r+1}\!\leftarrow\!\Dl^r\cup\mathcal{S}^r$;
  $\Du^{r+1}\!\leftarrow\!\Du^r\setminus\mathcal{S}^r$
  \hfill\Comment{Eq.~\eqref{eq:update}}

\EndFor
\State \Return $f_s$
\end{algorithmic}
\end{algorithm}

\section{Performance Evaluation} \label{sec:experiments}

\subsection{Experimental Setup}

\noindent\underline{\bf\em Datasets.}
We evaluate {\sc SatLabel} on eleven remote-sensing scene-classification datasets grouped into three categories. The \textit{Core} group comprises AID~\cite{xia2017aid}, UC-Merced~\cite{yang2010bag}, and EuroSAT~\cite{helber2019eurosat}, forming the baseline training configuration. The \textit{Extended} group includes RESISC45~\cite{cheng2017remote}, PatternNet~\cite{zhou2018patternnet}, WHU-RS19~\cite{xia2010structural}, and RSI-CB256~\cite{li2020rsi}. The \textit{Unseen} group comprises OPTIMAL31 ~\cite{wang2018scene}, RSC11 ~\cite{zhao2016feature}, RS2800 ~\cite{liu2025efficient}, and SIRI-WHU ~\cite{zhao2016dirichlet}, which are excluded from training and used solely for out-of-domain evaluation using the canonical class mapping. Datasets that do not satisfy the single-label scene-classification setting are excluded from the benchmark.


\vspace{1mm}
\noindent\underline{\bf\em Balanced Sampling and Querying.} The datasets vary substantially in size, from 1,005 to 31,500 images, causing conventional pooled sampling to favor larger datasets during training. To mitigate this bias, {\sc SatLabel} employs \emph{dataset-balanced sampling}, assigning each sample from dataset $d$ a weight proportional to $1/N_d$, where $N_d$ is the dataset size. This promotes a more uniform contribution across domains. We further use \emph{stratified active querying}, allocating the annotation budget across datasets and selecting the least-confident samples within each dataset. This prevents larger datasets from dominating the annotation budget and maintains coverage of smaller domains, such as WHU-RS19.


For each experiment, 20\,\% of images are held out for testing, 30\,\% initialize $\Dl^0$, and the remaining 50\,\% form $\Du^0$. Training proceeds for $R\!=\!5$ AL rounds with budget $B\!=\!200$ per round.

\vspace{1mm}
\noindent\underline{\bf\em Student Variants.}
We compare two variants: (i)~\textit{Balanced}, using ResNet-18 with dataset-balanced sampling; and (ii)~\textit{Balanced~+~MoE+GCN}, augmenting ResNet-18 with Sinkhorn-routed MoE and BatchGCN. Both use the same sampling strategy, isolating the architectural contribution.

\vspace{1mm} 
\noindent\underline{\bf\em Implementation.}
We use Adam with cosine annealing ($3\times10^{-4}$ to $1.5\times10^{-5}$), gradient clipping (max norm~1.0), and FP16. The MoE uses $K=4$ experts with top-2 Sinkhorn routing, and BatchGCN uses $k=8$. Pseudo-labeling ($M=600$, $\tau=0.95$) starts from round~3. We set $\lambda_{\mathrm{CE}}=0.3$, $\lambda_{\mathrm{KD}}=0.7$, $\alpha=0.999$, and $T=2$. We report best-checkpoint accuracy and Macro-F1, with Macro-F1 computed over classes present in each dataset.


\vspace{1mm}
\noindent\underline{\bf\em Baselines.}
We compare against four recent remote-sensing foundation VLMs. \textit{RemoteCLIP}~\cite{liu2024remoteclip}, pre-trained on remote-sensing image--text pairs, serves as both our frozen teacher and primary baseline. \textit{RS-M-CLIP}~\cite{wang2023samrs} is trained on the large-scale SAMRS dataset, while \textit{GeoRSCLIP}~\cite{zhang2024rs5m} is trained on RS5M with broad remote-sensing coverage. \textit{SkyCLIP}~\cite{mall2024remote} learns ground--aerial image alignment without text supervision.

\begin{table}[!t]
\centering
\caption{Accuracy comparison of {\sc SatLabel} against baselines on the core and extended datasets. Best results are shown in \textbf{\em bold} and second-best results are \underline{\em underlined}. Values are reported in percent. The student variant used by {\sc SatLabel} is indicated in parentheses.}
\label{tab:comparison_models}
\scriptsize
\setlength{\tabcolsep}{2pt}
\renewcommand{\arraystretch}{1.3}
\resizebox{\linewidth}{!}{
\begin{tabular}{lcccc|l}
\toprule
Dataset & RemoteCLIP & RS-M-CLIP & GeoRSCLIP & SkyCLIP & {\sc SatLabel} \\
\midrule
AID
& \underline{86.83}
& 88.90
& 76.33
& 70.89
& \textbf{94.41} (MoE+GCN) \\

EuroSAT
& 32.46
& 25.85
& \underline{67.47}
& 71.70
& \textbf{95.37} (Balanced) \\

UC Merced
& 77.95
& \textbf{91.50}
& 74.50
& 74.80
& \underline{91.00} (MoE+GCN) \\

RESISC45
& 68.38
& \textbf{92.62}
& 73.83
& 70.94
& \underline{86.63} (Balanced) \\

PatternNet
& 57.52
& 50.95
& 74.80
& \underline{80.88}
& \textbf{95.90} (Balanced) \\

RSI-CB256
& 50.38
& 47.50
& \underline{61.00}
& 50.09
& \textbf{87.01} (Balanced) \\

WHU-RS19
& \textbf{94.33}
& \underline{93.80}
& 82.00
& 83.50
& 77.41 (MoE+GCN) \\
\bottomrule
\end{tabular}}
\vspace{-5mm}
\end{table}

\subsection{Results: Comparison with Baselines}
\noindent{\underline{\bf \em Accuracy and Macro-F1 Comparison.} Table~\ref{tab:comparison_models} compares the two {\sc SatLabel} student variants with our baselines across the core and extended datasets. {\sc SatLabel} achieves the highest accuracy on four of seven benchmarks, demonstrating the benefit of combining knowledge distillation with active semi-supervised adaptation. The largest gains occur on EuroSAT, PatternNet, and RSI-CB256, reaching 95.37\%, 95.90\%, and 87.01\%, respectively. On AID, the MoE+GCN variant achieves 94.41\%, compared with 88.90\% for RS-M-CLIP. These improvements suggest that the student effectively transfers foundation-model semantic knowledge while incorporating target-domain supervision and unlabeled data.

RS-M-CLIP remains strongest on UC Merced (91.50\%) and RESISC45 (92.62\%), where {\sc SatLabel} achieves 91.00\% and 86.63\%, respectively. RemoteCLIP performs best on WHU-RS19 (94.33\%), substantially exceeding the student (77.41\%). Despite these modest accuracy gaps, {\sc SatLabel} employs a {\em substantially lighter student architecture}, making it more suitable for resource-constrained onboard deployment while maintaining competitive predictive performance. Thus, the small loss in accuracy on these datasets represents a favorable accuracy--efficiency trade-off for satellite deployment. Overall, {\sc SatLabel} demonstrates the potential of task-specific student adaptation to approach or exceed large foundation models while substantially reducing deployment complexity.


Fig.~\ref{fig:remoteclip_vs_student} compares Macro-F1 for RemoteCLIP and the best {\sc SatLabel} student. The trends corroborate the accuracy results, with substantial gains on EuroSAT, PatternNet, and RSI-CB256 and competitive performance on AID and UC Merced. The WHU-RS19 gap further shows that the effectiveness of adaptation varies across domains, with pretrained semantic priors remaining advantageous for some datasets.

\begin{figure}[!t]
\centering
\includegraphics[width=\linewidth]{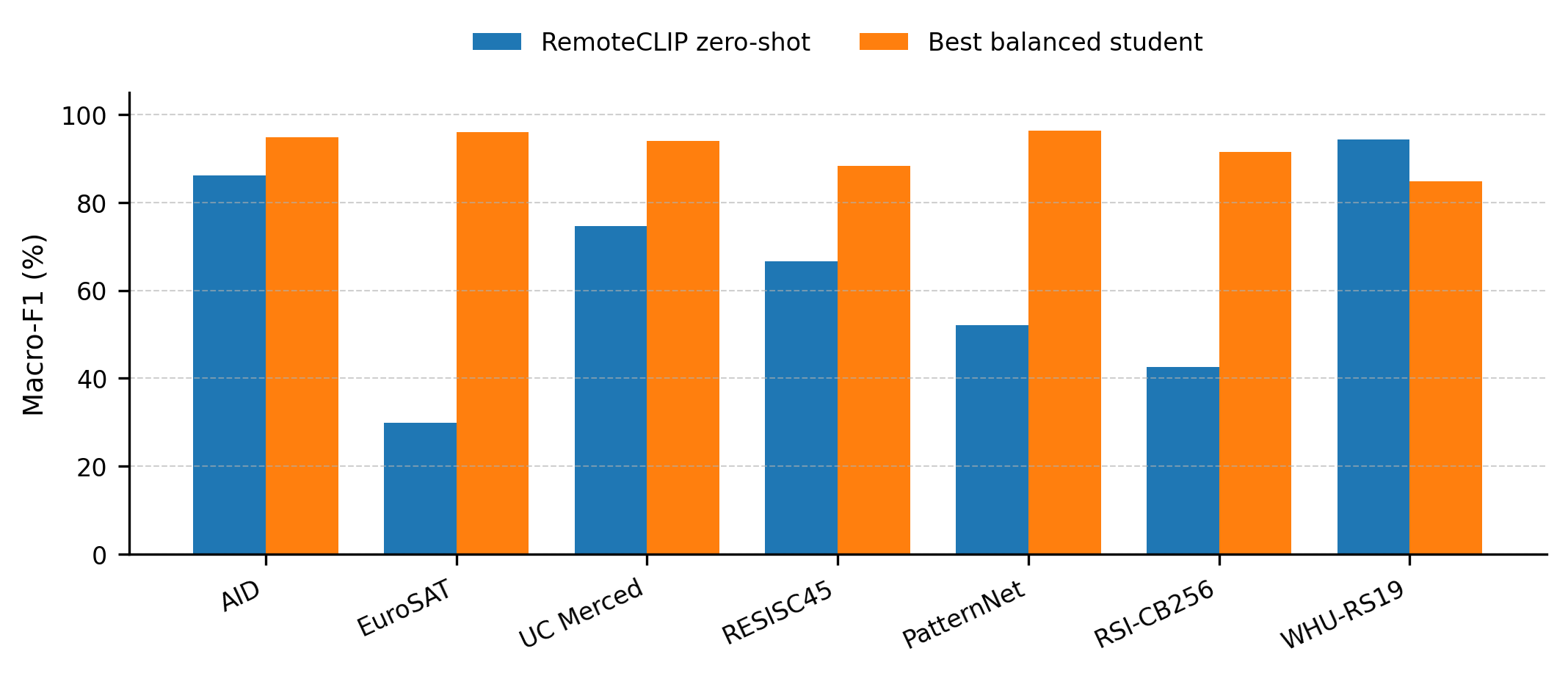}
\caption{RemoteCLIP zero-shot Macro-F1 compared with the best balanced student model on the core and extended datasets. The student outperforms RemoteCLIP on most datasets, while RemoteCLIP remains strongest on WHU-RS19.}
\label{fig:remoteclip_vs_student}
\end{figure}

 \begin{table}[!t]
\centering
\caption{Comparison of {\sc SatLabel} with baselines in terms of parameters, model size and computational complexity.}
\label{tab:eff_eurosat_complexity}
\setlength{\tabcolsep}{4pt}
\renewcommand{\arraystretch}{1.15}
\resizebox{\linewidth}{!}{
\begin{tabular}{@{}lccc@{}}
\toprule
\textbf{Model} &
\textbf{Params (M)$\downarrow$} &
\textbf{Size (MB)$\downarrow$} &
\textbf{FLOPs (G)$\downarrow$} \\
\midrule
RemoteCLIP ViT-B/32
& 151.28 & 577.21 & 5.89 \\

RS-M-CLIP
& 366.12 & 1396.81 & 5.89 \\

GeoRSCLIP ViT-B/32
& 151.28 & 577.21 & 5.89 \\

SkyCLIP ViT-B/32
& 151.28 & 577.21 & 5.89 \\
\midrule
\textbf{SatLabel (Balanced)}
& \textbf{11.21} & \textbf{42.83} & \textbf{3.65} \\

\textbf{SatLabel (MoE+GCN)}
& \underline{20.40} & \underline{77.93} & \underline{3.67} \\
\bottomrule
\end{tabular}}
\end{table}

\begin{table}[!t] 
\centering 
\caption{Comparison of {\sc SatLabel} with baselines in terms of runtime and resource efficiency on AID (10,000 images).} 
\label{tab:eff_aid_runtime} 
\setlength{\tabcolsep}{3pt} 
\renewcommand{\arraystretch}{1.15}
\resizebox{\linewidth}{!}{
\begin{tabular}{@{}lcccc@{}}
\toprule 
\textbf{Model} & \textbf{Images/s}$\uparrow$ & \textbf{Energy (J/img)$\downarrow$} & \textbf{Power (W)$\downarrow$} & \textbf{Peak GPU (MB)$\downarrow$} \\
\midrule
RemoteCLIP ViT-B/32 & 12128.4 & 0.128 & $143.19 \pm 62.97$ & 741.9 \\ 
RS-M-CLIP & 12211.2 & 0.124 & $140.00 \pm 59.80$ & 1560.4 \\

GeoRSCLIP ViT-B/32 & 11692.5 & 0.127 & $142.58 \pm 65.30$ & 760.7 \\ 

SkyCLIP ViT-B/32 & 12246.8 & 0.130 & $136.24 \pm 59.23$ & 741.9 \\
\midrule
\textbf{SatLabel (Balanced)} & \textbf{24353.8} & \textbf{0.106} & $\mathbf{110.55 \pm 30.50}$ & \textbf{381.1} \\ 
\textbf{SatLabel (MoE+GCN)} & \underline{12623.0} & \underline{0.123} & \underline{$132.34 \pm 47.20$} & \underline{575.4} \\
\bottomrule
\end{tabular}}\vspace{-5mm}
\end{table}

\begin{table}[t]
\centering
\caption{Comparison of {\sc SatLabel} with baselines in terms of runtime and resource efficiency on EuroSAT (27,000 images).}
\label{tab:eff_eurosat_runtime}
\setlength{\tabcolsep}{3pt}
\renewcommand{\arraystretch}{1.15}
\resizebox{\linewidth}{!}{
\begin{tabular}{@{}lcccc@{}}
\toprule
\textbf{Model} &
\textbf{Images/s}$\uparrow$ &
\textbf{Energy (J/img)$\downarrow$} &
\textbf{Power (W)$\downarrow$} &
\textbf{Peak GPU (MB)$\downarrow$} \\
\midrule
RemoteCLIP ViT-B/32
& 11966.9 & 0.055 & $291.88 \pm 46.42$ & 741.9 \\

RS-M-CLIP
& 12028.5 & 0.055 & $295.95 \pm 31.00$ & 1560.4 \\

GeoRSCLIP ViT-B/32
& 11712.4 & 0.056 & $299.99 \pm 46.62$ & 760.6 \\

SkyCLIP ViT-B/32
& 12111.9 & 0.055 & $297.14 \pm 40.29$ & 741.9 \\
\midrule
\textbf{SatLabel (Balanced)}
& \textbf{24143.0} & \textbf{0.036}
& $\mathbf{185.11 \pm 27.21}$ & \textbf{381.1} \\

\textbf{SatLabel (MoE+GCN)}
& \underline{12607.4} & \underline{0.052}
& \underline{$268.27 \pm 27.63$} & \underline{575.4} \\
\bottomrule
\end{tabular}}
\end{table}

 






%
%

\begin{table}[t]
\centering
\caption{Comparison of {\sc SatLabel} with baselines in terms of runtime and resource efficiency on PatternNet (30,400 images).}
\label{tab:eff_patternnet_runtime}
\setlength{\tabcolsep}{3pt}
\renewcommand{\arraystretch}{1.15}
\resizebox{\linewidth}{!}{
\begin{tabular}{@{}lcccc@{}}
\toprule
\textbf{Model} &
\textbf{Images/s}$\uparrow$ &
\textbf{Energy (J/img)$\downarrow$} &
\textbf{Power (W)$\downarrow$} &
\textbf{Peak GPU (MB)$\downarrow$} \\
\midrule
RemoteCLIP ViT-B/32
& 11992.8 & 0.064 & $232.96 \pm 50.63$ & 741.9 \\

RS-M-CLIP
& 11852.3 & 0.068 & \underline{$222.07 \pm 46.29$ }& 1560.4 \\

GeoRSCLIP ViT-B/32
& 11711.4 & 0.065 & $245.00 \pm 53.79$ & 760.7\\

SkyCLIP ViT-B/32
& 12012.8 & 0.064 & $241.46 \pm 47.66$ & 741.9 \\
\midrule
\textbf{SatLabel (Balanced)}
& \textbf{24282.9} & \textbf{0.044}
& $\mathbf{160.21 \pm 31.45}$ & \textbf{381.2} \\

\textbf{SatLabel (MoE+GCN)}
& \underline{12598.1} & \underline{0.059}
& $226.55 \pm 35.38$ & \underline{575.5} \\
\bottomrule
\end{tabular}}
\end{table}







%
%

\begin{table}[!t]
\centering
\caption{Comparison of {\sc SatLabel} with baselines in terms of runtime and resource efficiency on OPTIMAL31 (1,860 images).}
\label{tab:eff_optimal31_runtime}
\setlength{\tabcolsep}{3pt}
\renewcommand{\arraystretch}{1.15}
\resizebox{\linewidth}{!}{
\begin{tabular}{@{}lcccc@{}}
\toprule
\textbf{Model} &
\textbf{Images/s}$\uparrow$ &
\textbf{Energy (J/img)$\downarrow$} &
\textbf{Power (W)$\downarrow$} &
\textbf{Peak GPU (MB)$\downarrow$} \\
\midrule
RemoteCLIP ViT-B/32
& 12029.0 & \textbf{0.097} & $394.36 \pm 5.35$ & 741.9 \\

RS-M-CLIP
& 11891.9 & 0.103 & $408.85 \pm 2.68$ & 1560.4 \\

GeoRSCLIP ViT-B/32
& 11585.5 & 0.107 & $418.21 \pm 16.80$ & 760.7 \\

SkyCLIP ViT-B/32
& 12185.9 & 0.105 & $402.34 \pm 15.96$ & 741.9 \\
\midrule
\textbf{SatLabel (Balanced)}
& \textbf{24292.2} & \underline{0.100}
& $\mathbf{254.30 \pm 24.24}$ & \textbf{381.3} \\

{\bf SatLabel (MoE+GCN)}
& \underline{12677.1} & 0.129
& \underline{$351.10 \pm 10.88$} & \underline{575.6} \\
\bottomrule
\end{tabular}}\vspace{-5mm}
\end{table}

\vspace{1mm}
\noindent{\underline{\bf \em Computation and Efficiency Comparison.}}
Beyond classification accuracy, computational and energy efficiency are critical for onboard satellite image annotation, where limited compute, memory, and power constrain continuous large-scale processing. Table~\ref{tab:eff_eurosat_complexity} summarizes the model complexity of {\sc SatLabel} and the CLIP-based baselines in terms of parameter count, model size, and floating-point operations (FLOPs). The Balanced student requires only 11.21M parameters, 42.83~MB of storage, and 3.65~GFLOPs per inference, compared with 151.28--366.12M parameters, 577.21--1396.84~MB, and 5.89~GFLOPs for the CLIP-based baselines. Thus, relative to the smallest CLIP baseline, {\sc SatLabel} Balanced reduces the parameter count and model size by approximately $13.5\times$, while reducing the computational cost by approximately $1.6\times$. Relative to RS-M-CLIP, these reductions reach approximately $32.6\times$ for both parameters and model size. The MoE+GCN variant uses 20.4M parameters and 77.9~MB, while remaining substantially smaller than the CLIP-based baselines.


Tables~\ref{tab:eff_aid_runtime}--\ref{tab:eff_optimal31_runtime} further evaluate {\sc SatLabel} and the baselines across AID, EuroSAT, PatternNet, and OPTIMAL31 in terms of inference throughput, energy per image, average power, and peak GPU memory. All experiments use a batch size of 64, $224\times224$ input images, and automatic mixed precision (AMP). The CLIP-based baselines use ViT-B/32 implemented with OpenCLIP. For zero-shot classification, we use the prompt template \texttt{a satellite image of \{\}}, with class names derived from the corresponding folder names by replacing underscores and slashes with spaces. Class-text embeddings are computed once using \texttt{encode\_text} and cached before evaluation. Consequently, the measured inference path consists of \texttt{encode\_image} followed by cosine-similarity classification against the cached text embeddings. Images/s denotes GPU-forward throughput. Energy is obtained by integrating GPU power measurements from NVML over the evaluation period and normalizing by the number of processed images; power and peak GPU memory are measured under the same evaluation configuration.

Across all four datasets, {\sc SatLabel} Balanced student variant achieves roughly $2\times$ higher inference throughput than the CLIP-based baselines, processing 24.1K--24.4K images/s versus 11.6K--12.2K images/s. The MoE+GCN variant achieves approximately 12.6K images/s due to its additional expert and graph-based processing. The Balanced variant also consumes less energy on AID, EuroSAT, and PatternNet, with 0.106, 0.036, and 0.044~J/image, corresponding to reductions of 15--18\%, 35--36\%, and 31--35\%, respectively, relative to the lowest-energy CLIP baseline. On OPTIMAL31, RemoteCLIP uses slightly less energy (0.097 vs. 0.100~J/image). Nevertheless, {\sc SatLabel} Balanced consistently achieves lower average GPU power across all four datasets.

The compact student also substantially reduces peak GPU memory on AID (381.1~MB vs. 741.9--1560.4~MB for the baselines), while the MoE+GCN variant remains considerably smaller than the CLIP-based models. Overall, {\sc SatLabel} provides a favorable accuracy--efficiency trade-off, combining lower model complexity with approximately $2\times$ higher throughput and generally lower energy and power consumption.

\subsection{Results: Ablation Study}

\begin{table}[t]
\centering
\caption{Best student results on unseen datasets using canonical class mapping. Values are reported in percent.}
\label{tab:unseen_eval}
\scriptsize
\setlength{\tabcolsep}{3pt}
\renewcommand{\arraystretch}{1.15}
\resizebox{\linewidth}{!}{
\begin{tabular}{lrrrrl}
\toprule
Dataset & Cov. & RC F1 & Stud. Acc. & Stud. F1 & Best setup \\
\midrule
OPTIMAL31 & 100.00 & 75.98 & 92.63 & 93.56 & 06 Mix, Balanced \\
RSC11 & 90.91 & 60.42 & 56.97 & 71.97 & 06 Mix, MoE+GCN \\
RS2800 & 100.00 & 62.99 & 48.82 & 54.70 & 04 WHU, MoE+GCN \\
SIRI-WHU & 100.00 & 59.07 & 47.33 & 51.14 & 01 Base, MoE+GCN \\
\bottomrule
\end{tabular}}\vspace{-5mm}
\end{table}

\noindent{\underline{\bf \em Cross-Dataset Generalization with Canonical Class Mapping.}}
To evaluate cross-dataset generalization, we test the trained students on four unseen benchmarks: OPTIMAL31, RSC11, RS2800, and SIRI-WHU. Since their native label spaces differ from the student's training label space, we map semantically equivalent classes to a shared canonical label space. For example, \emph{airport} across OPTIMAL31, AID, RESISC45, and WHU-RS19 is mapped to the canonical class \emph{airport}. Evaluation is then performed only on classes with valid mappings, with coverage (Cov.) reporting the percentage of unseen classes retained after mapping.

Table~\ref{tab:unseen_eval} shows that the student models achieve strong transfer when sufficient semantic overlap exists. On OPTIMAL31, {\sc SatLabel} Balanced achieves 92.63\% accuracy and 93.56\% Macro-F1, exceeding RemoteCLIP by 17.58 percentage points in Macro-F1. On RSC11, the MoE+GCN student achieves 71.97\% Macro-F1, an 11.55-point improvement over RemoteCLIP, despite lower overall accuracy. In contrast, RemoteCLIP outperforms the students on RS2800 and SIRI-WHU by 8.29 and 7.93 Macro-F1 points, respectively. These results indicate that student transfer depends on the degree of semantic and domain alignment between the unseen benchmark and the training data, while RemoteCLIP's zero-shot representations can provide an advantage under greater distribution shift.

\vspace{1mm}
\noindent{\underline{\bf \em Comparison of Student Backbone Architectures.}}
Table~\ref{tab:kd_results} compares knowledge-distilled students using ResNet-18 and ResNet-50 across seven remote-sensing benchmarks. Both students substantially outperform zero-shot RemoteCLIP on six datasets, demonstrating effective transfer of foundation-model semantics into compact task-specific classifiers. ResNet-18 achieves the best student Macro-F1 on UC Merced, PatternNet, and RSI-CB256, while ResNet-50 provides modest gains on AID, EuroSAT, and RESISC45. On WHU-RS19, however, RemoteCLIP remains superior, achieving 94.33\% Macro-F1 versus 84.81\% and 85.93\% for ResNet-18 and ResNet-50, respectively. Overall, the results show that increasing backbone capacity does not consistently improve transfer performance, supporting ResNet-18 as a more efficient student architecture with competitive or superior accuracy across most benchmarks.


\begin{table}[t]
\centering
\caption{RemoteCLIP zero-shot performance compared with knowledge-distilled ResNet-18 and ResNet-50 student models. Best student results are shown in \textbf{bold}. Values are reported in percent.}
\label{tab:kd_results}
\small
\setlength{\tabcolsep}{4pt}
\renewcommand{\arraystretch}{1.15}
\resizebox{\linewidth}{!}{
\begin{tabular}{lcc cc cc}
\toprule
& \multicolumn{2}{c}{RemoteCLIP}
& \multicolumn{2}{c}{Student ResNet-18}
& \multicolumn{2}{c}{Student ResNet-50} \\
\cmidrule(lr){2-3}
\cmidrule(lr){4-5}
\cmidrule(lr){6-7}
Dataset & Acc. & F1 & Acc. & F1 & Acc. & F1 \\
\midrule
AID        & 86.83 & 86.04 & 94.41 & 94.76 & \textbf{94.53} & \textbf{95.02} \\
EuroSAT    & 32.46 & 29.87 & \textbf{95.37} & 95.95 & 94.87 & \textbf{96.29} \\
UC Merced  & 77.95 & 74.63 & \textbf{91.00} & \textbf{94.01} & 80.52 & 86.09 \\
RESISC45   & 68.38 & 66.53 & 86.63 & 88.24 & \textbf{88.38} & \textbf{90.34} \\
PatternNet & 57.52 & 52.15 & \textbf{95.90} & \textbf{96.26} & 93.24 & 93.41 \\
RSI-CB256  & 50.38 & 42.60 & \textbf{87.01} & \textbf{91.36} & 80.94 & 89.66 \\
WHU-RS19   & \textbf{94.33} & \textbf{94.33} & 77.41 & 84.81 & \textbf{78.52} & \textbf{85.93} \\
\bottomrule
\end{tabular}}
\end{table}

\begin{table}[t]
  \centering
  \caption{Cumulative component analysis on AID
(30\,\% labeled).}
  \label{tab:component_study}
  \small
  \setlength{\tabcolsep}{3pt}
  \renewcommand{\arraystretch}{1.1}
  \begin{tabular}{lcc}
    \toprule
    \textbf{Configuration} & \textbf{(Accuracy, \%)} & \textbf{$\Delta$}\\
    \midrule
    (1) ResNet-18 supervised                & 89.4 & ---\\
    (2) +KD from RemoteCLIP                 & 91.2 & +1.8\\
    (3) +Consistency (FixMatch)             & 92.3 & +1.1\\
    (4) +Active learning (random)$^\dagger$ & 92.6 & +0.3\\
    (5) +Active learning (uncertainty)      & 93.1 & +0.5\\
    (6) +MoE (softmax gating)              & 93.3 & +0.2\\
    (7) +MoE (Sinkhorn gating)             & 93.6 & +0.3\\
    (8) +BatchGCN \textbf{(full model)}    & \textbf{94.4} & +0.8\\
    \bottomrule
  \end{tabular}\vspace{-1mm}
\end{table}
 
\begin{figure}[t]
  \centering
  \begin{subfigure}[b]{0.49\linewidth}
    \includegraphics[width=\linewidth]{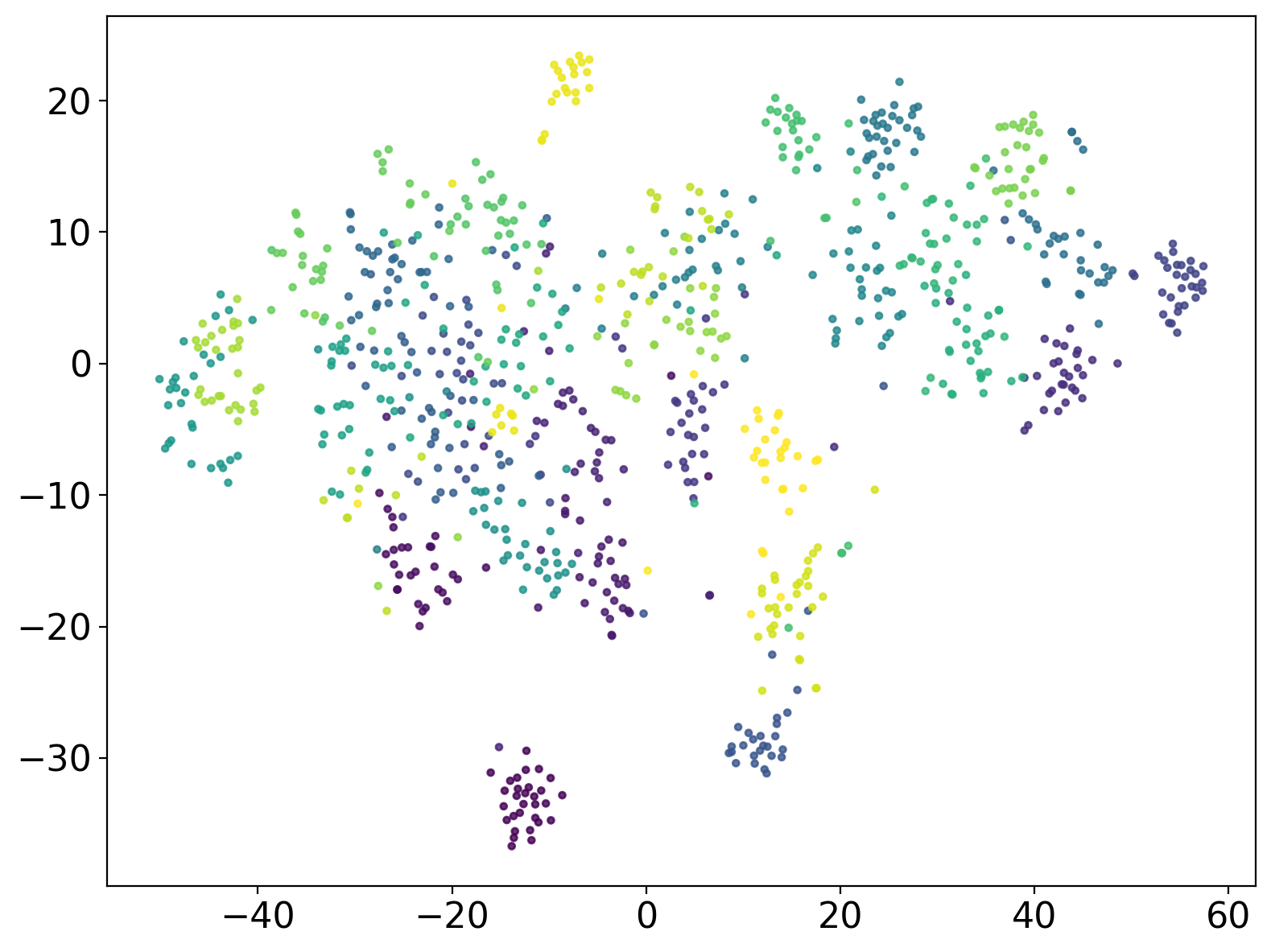}
    \caption{OPTIMAL31}
  \end{subfigure}
  \begin{subfigure}[b]{0.49\linewidth}
    \includegraphics[width=\linewidth]{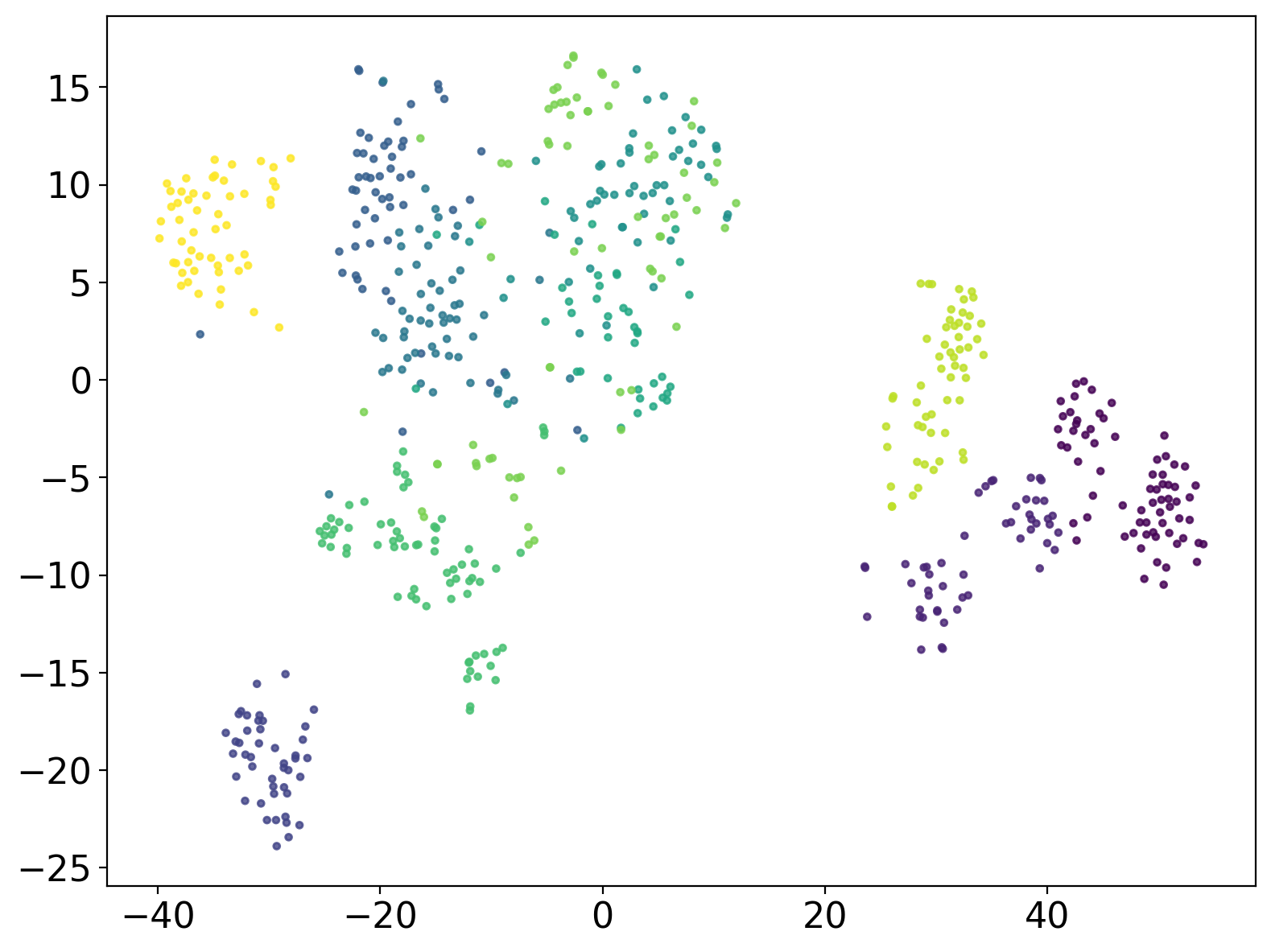}
    \caption{RSC11}
  \end{subfigure}
  \\
  \begin{subfigure}[b]{0.49\linewidth}
    \includegraphics[width=\linewidth]{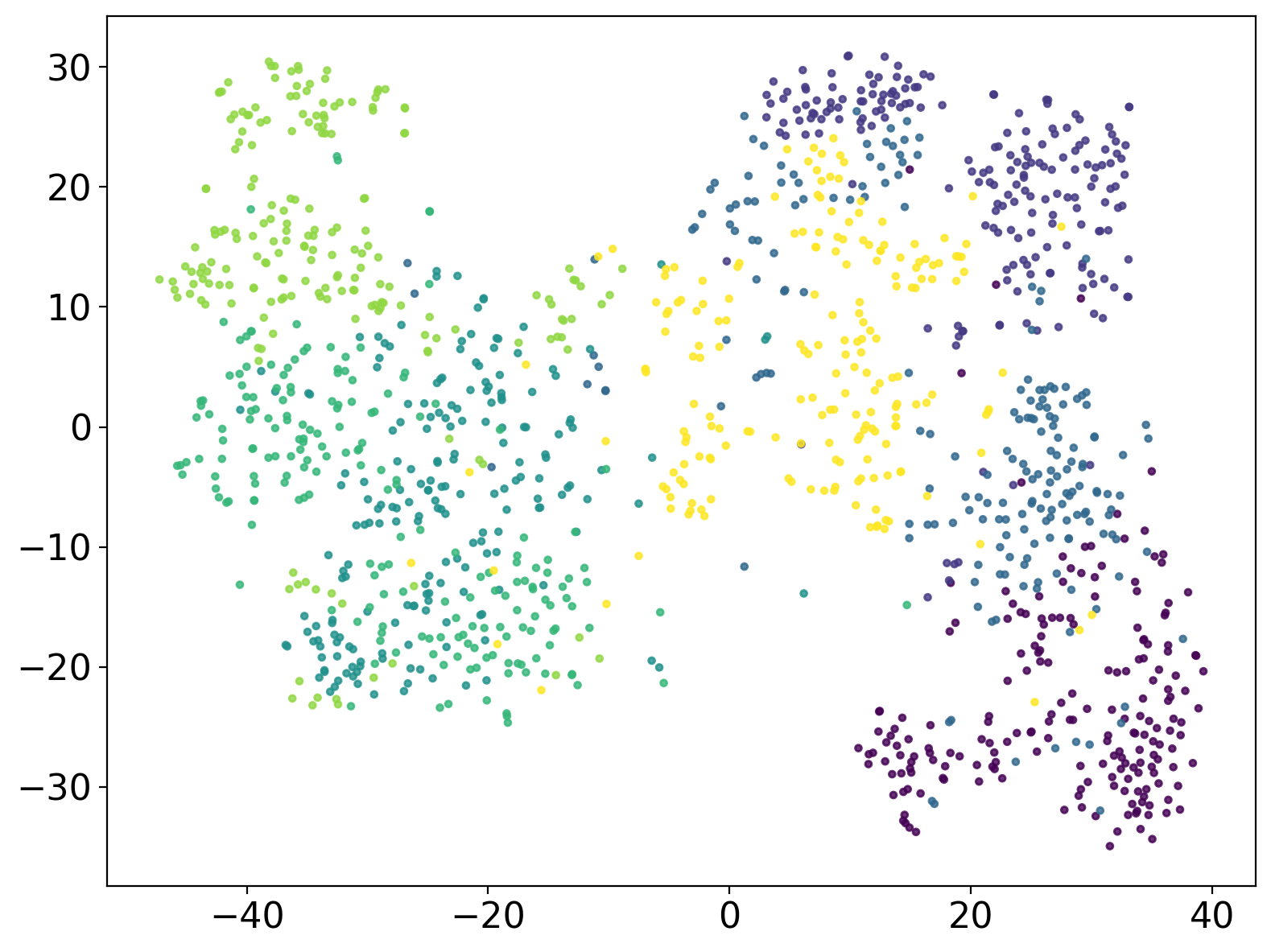}
    \caption{RS2800}
  \end{subfigure}
  \begin{subfigure}[b]{0.49\linewidth}
    \includegraphics[width=\linewidth]{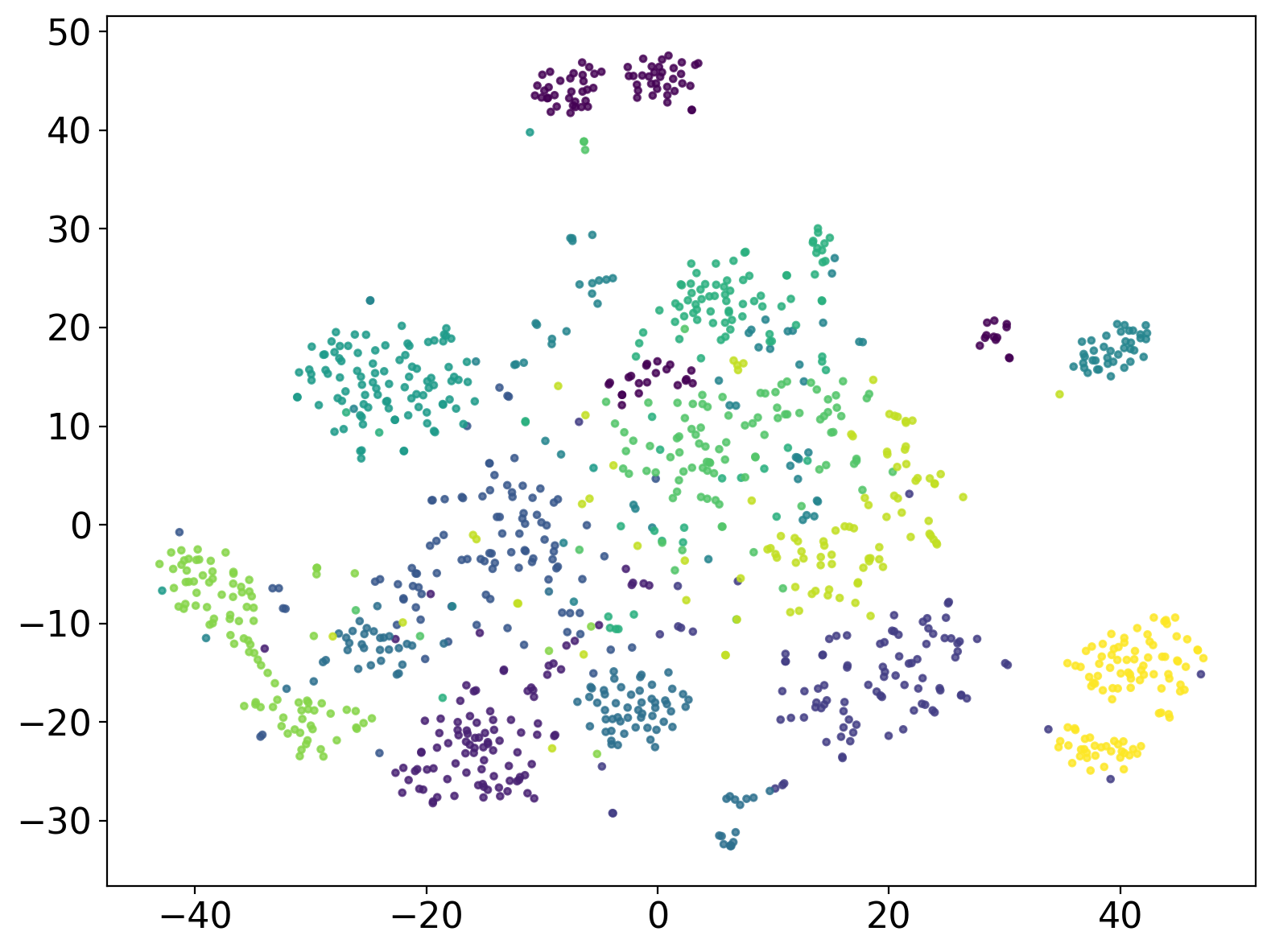}
    \caption{SIRI-WHU}
  \end{subfigure}
  \caption{t-SNE projections of student features on unseen datasets.
 }
  \label{fig:tsne_unseen}
\end{figure}

\vspace{1mm}
\noindent{\underline{\bf \em Model Component Analysis.}}
Table~\ref{tab:component_study} reports a cumulative ablation on AID with 30\% labeled data. Knowledge distillation provides the largest initial gain (+1.8 percentage points), followed by FixMatch consistency regularization (+1.1), demonstrating the complementary benefits of teacher-guided supervision and unlabeled-data utilization. Uncertainty-based active learning adds a further +0.5 points over random selection, while Sinkhorn routing improves the MoE student by +0.3 points over softmax gating. Finally, BatchGCN yields an additional +0.8 points, indicating the benefit of modeling within-batch feature relationships. Overall, the complete framework improves ResNet-18 accuracy from 89.4\% to 94.4\%, $5\times$ gain.


\begin{figure}[!t]
  \centering
  \begin{subfigure}[b]{0.32\linewidth}
    \includegraphics[
      width=\linewidth,
      height=2.5cm,
      keepaspectratio
    ]{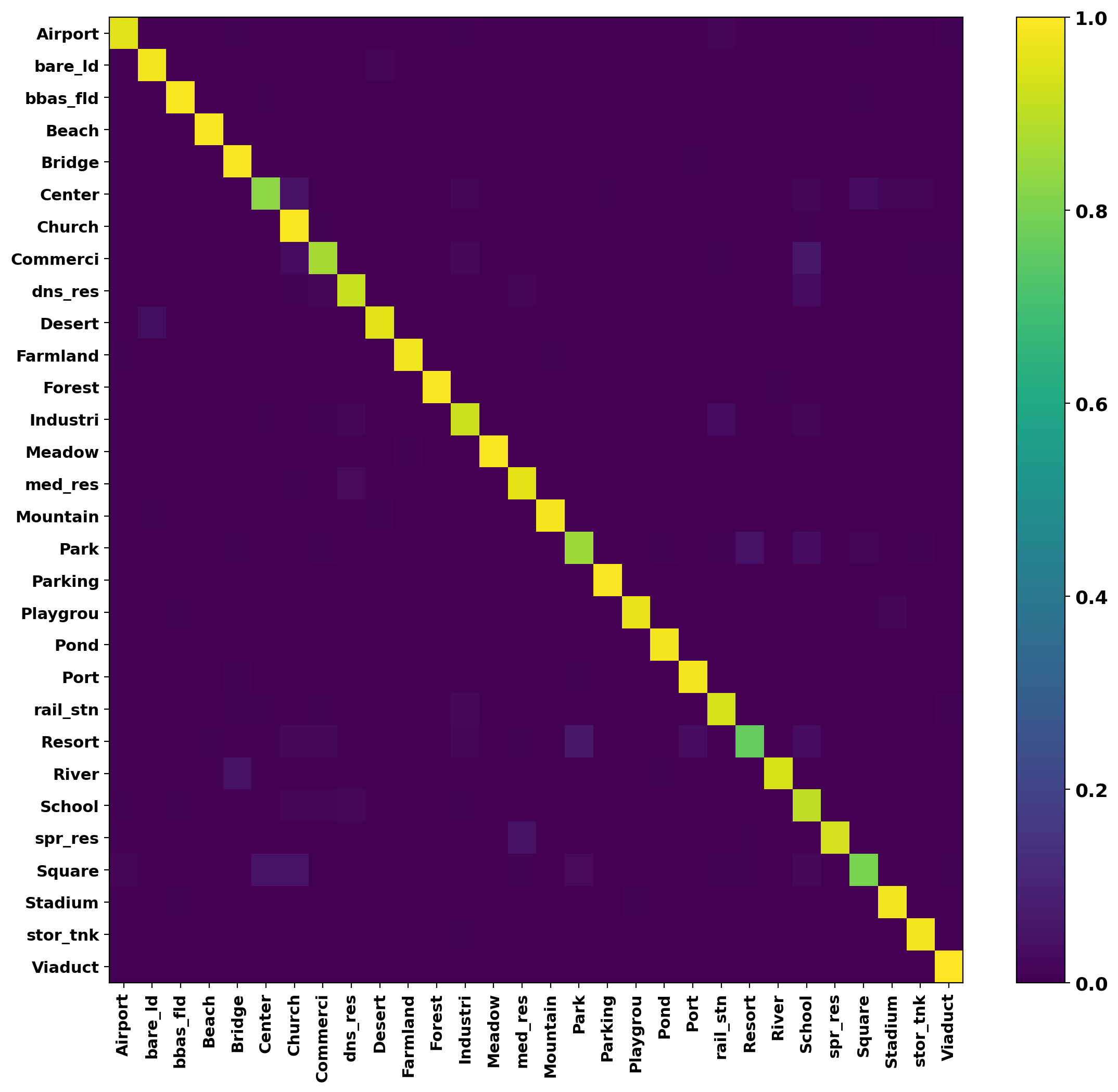}
    \caption{AID}
  \end{subfigure}
  \begin{subfigure}[b]{0.32\linewidth}
    \includegraphics[width=\linewidth]{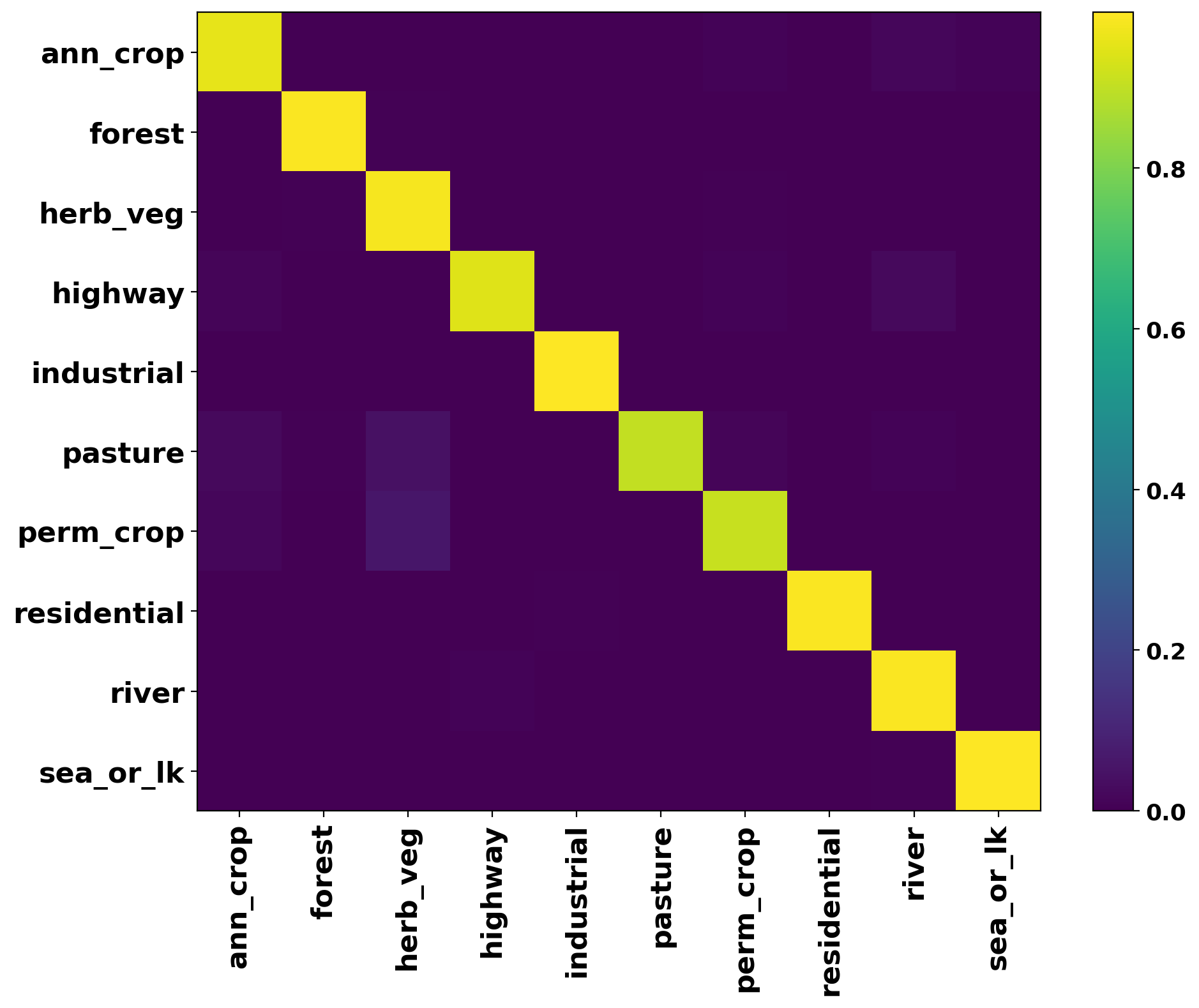}
    \caption{EuroSAT}
  \end{subfigure}
  \begin{subfigure}[b]{0.32\linewidth}
    \includegraphics[width=\linewidth]{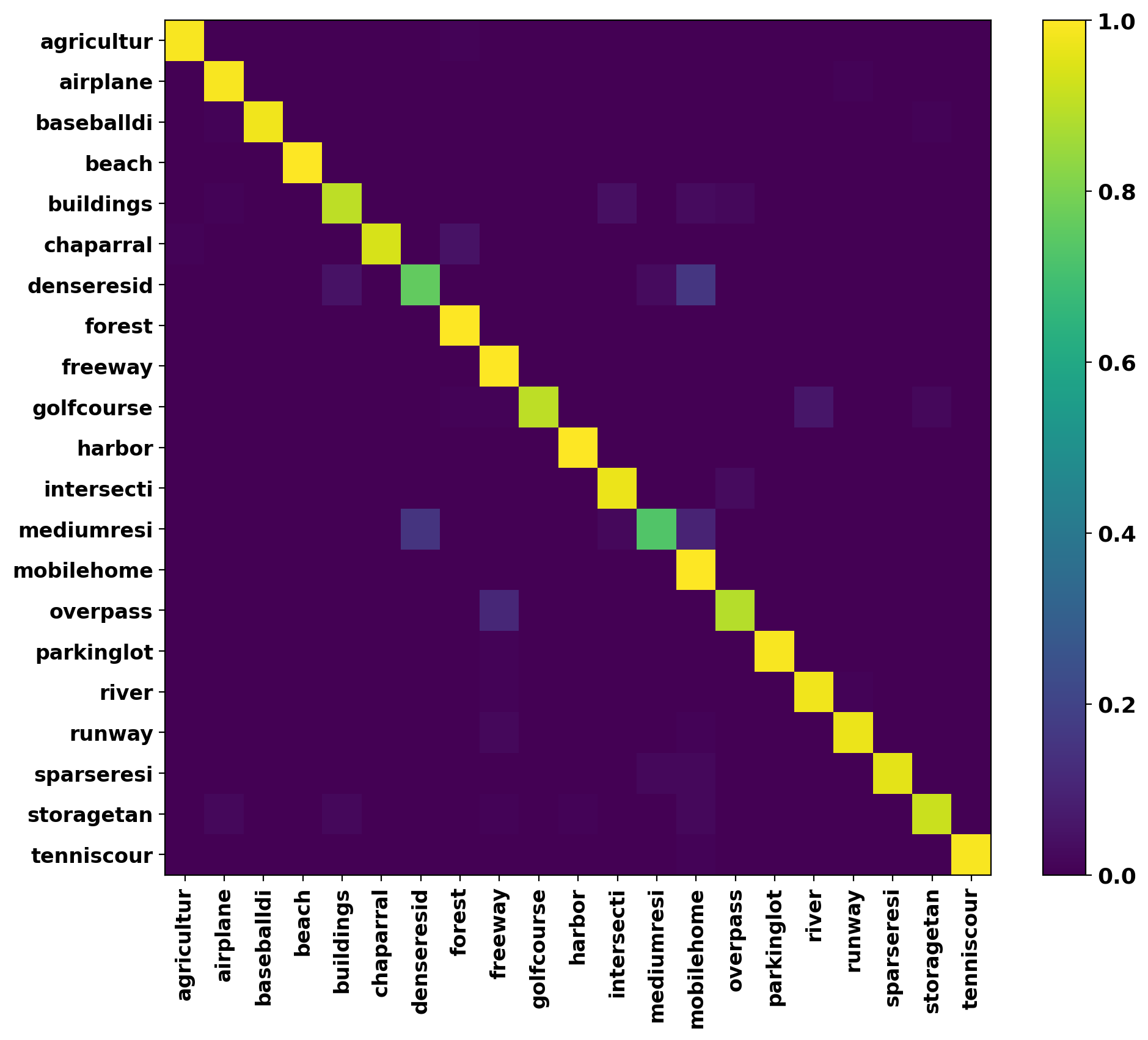}
    \caption{UC~Merced}
  \end{subfigure}
  \begin{subfigure}[b]{0.32\linewidth}
    \includegraphics[width=\linewidth]{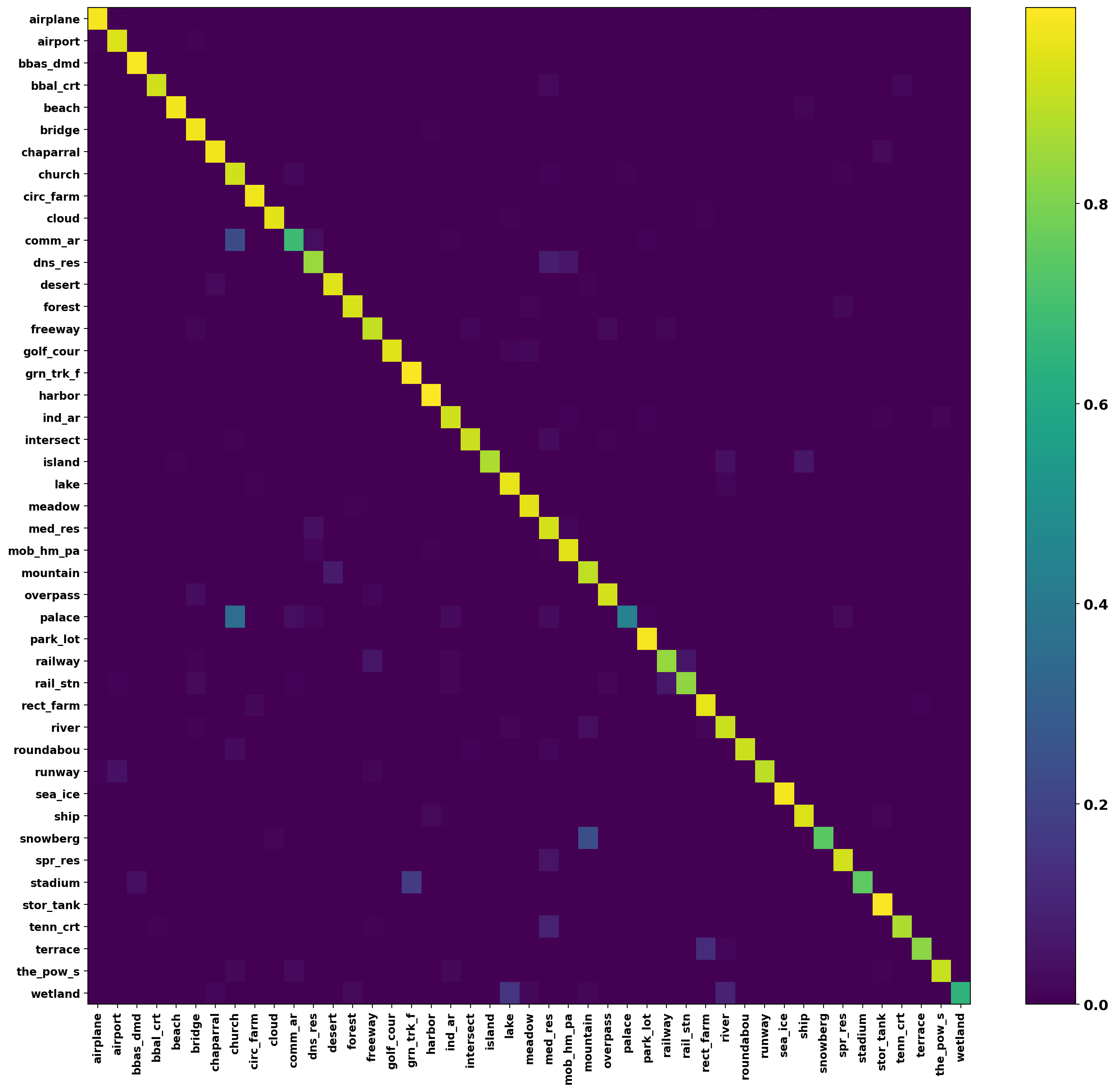}
    \caption{RESISC45}
  \end{subfigure}
  \begin{subfigure}[b]{0.32\linewidth}
    \includegraphics[width=\linewidth]{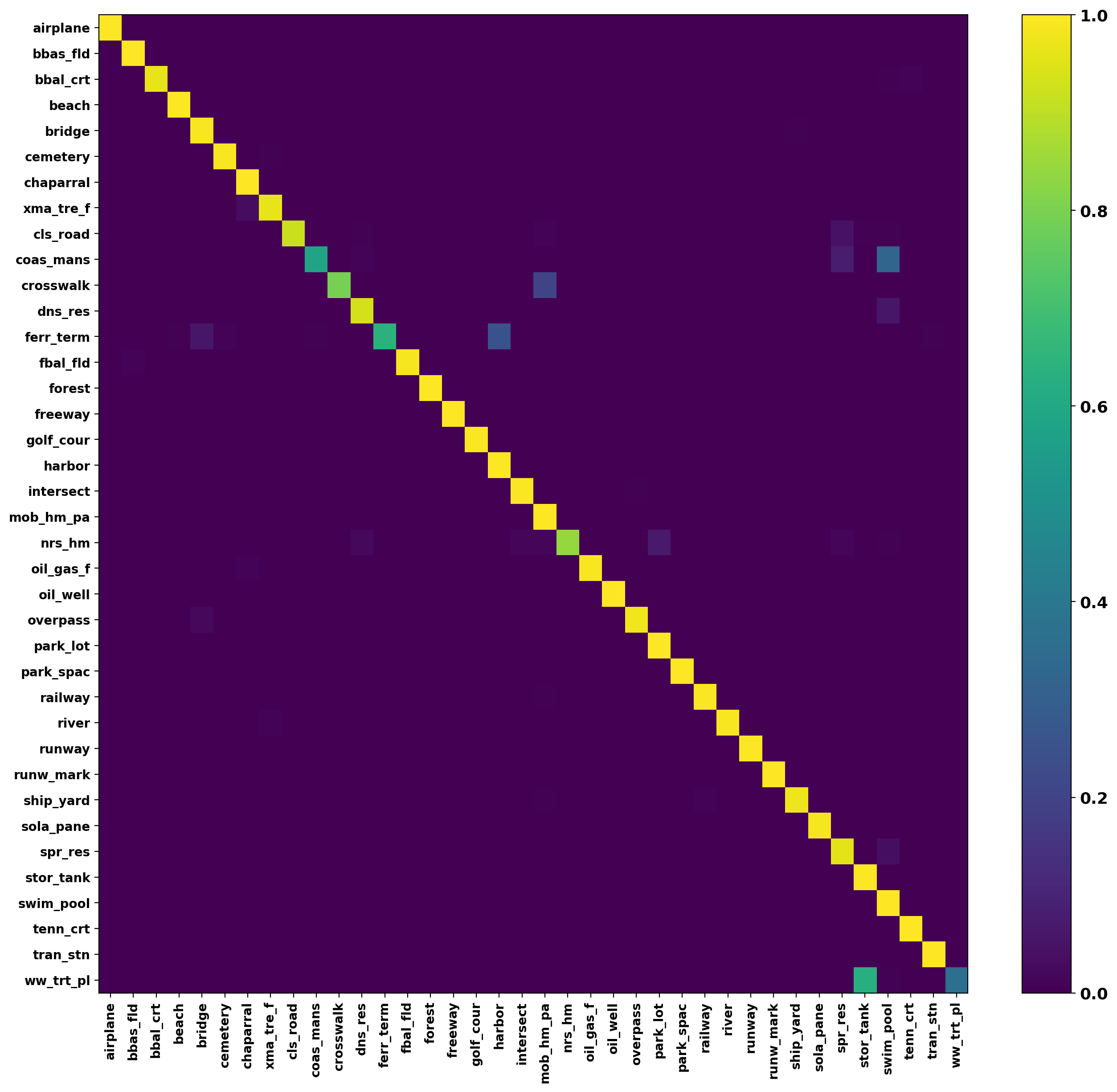}
    \caption{PatternNet}
  \end{subfigure}
  \begin{subfigure}[b]{0.32\linewidth}
    \includegraphics[width=\linewidth]{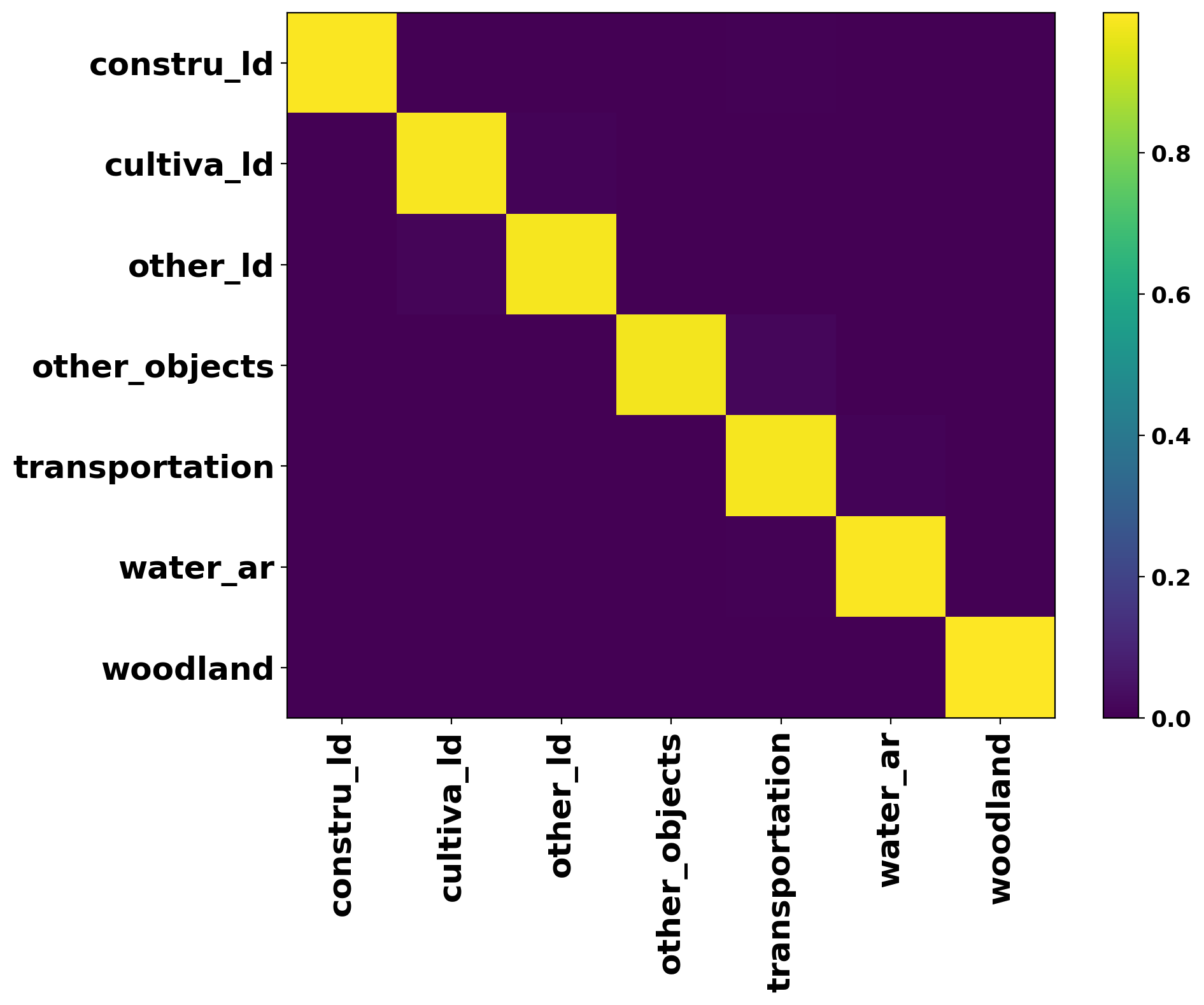}
    \caption{RSI-CB256}
  \end{subfigure}
  \begin{subfigure}[b]{0.24\linewidth}
    \includegraphics[width=\linewidth]{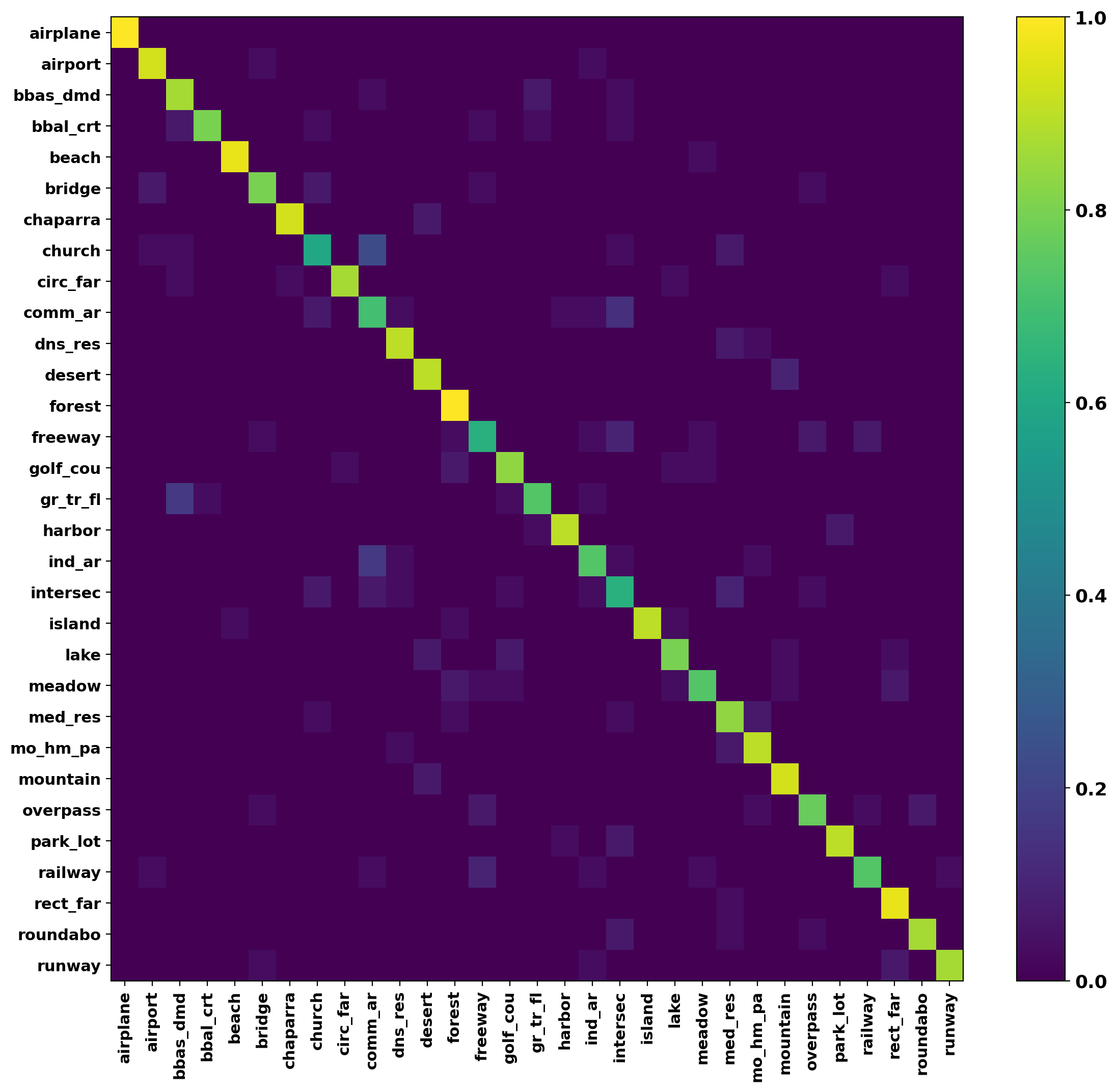}
    \caption{OPTIMAL31}
  \end{subfigure}
  \begin{subfigure}[b]{0.24\linewidth}
    \includegraphics[width=\linewidth]{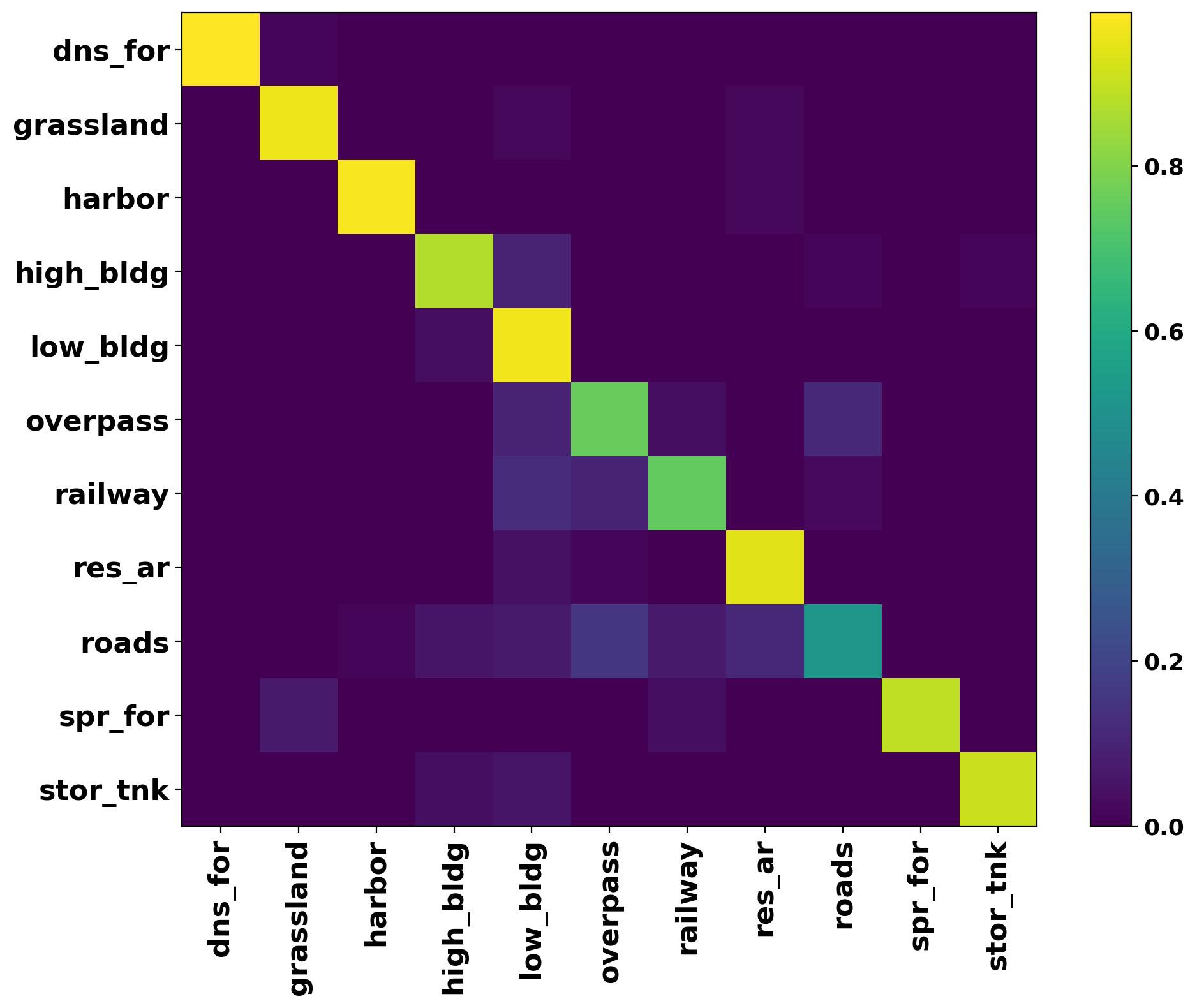}
    \caption{RSC11}
  \end{subfigure}
  \begin{subfigure}[b]{0.24\linewidth}
    \includegraphics[width=\linewidth]{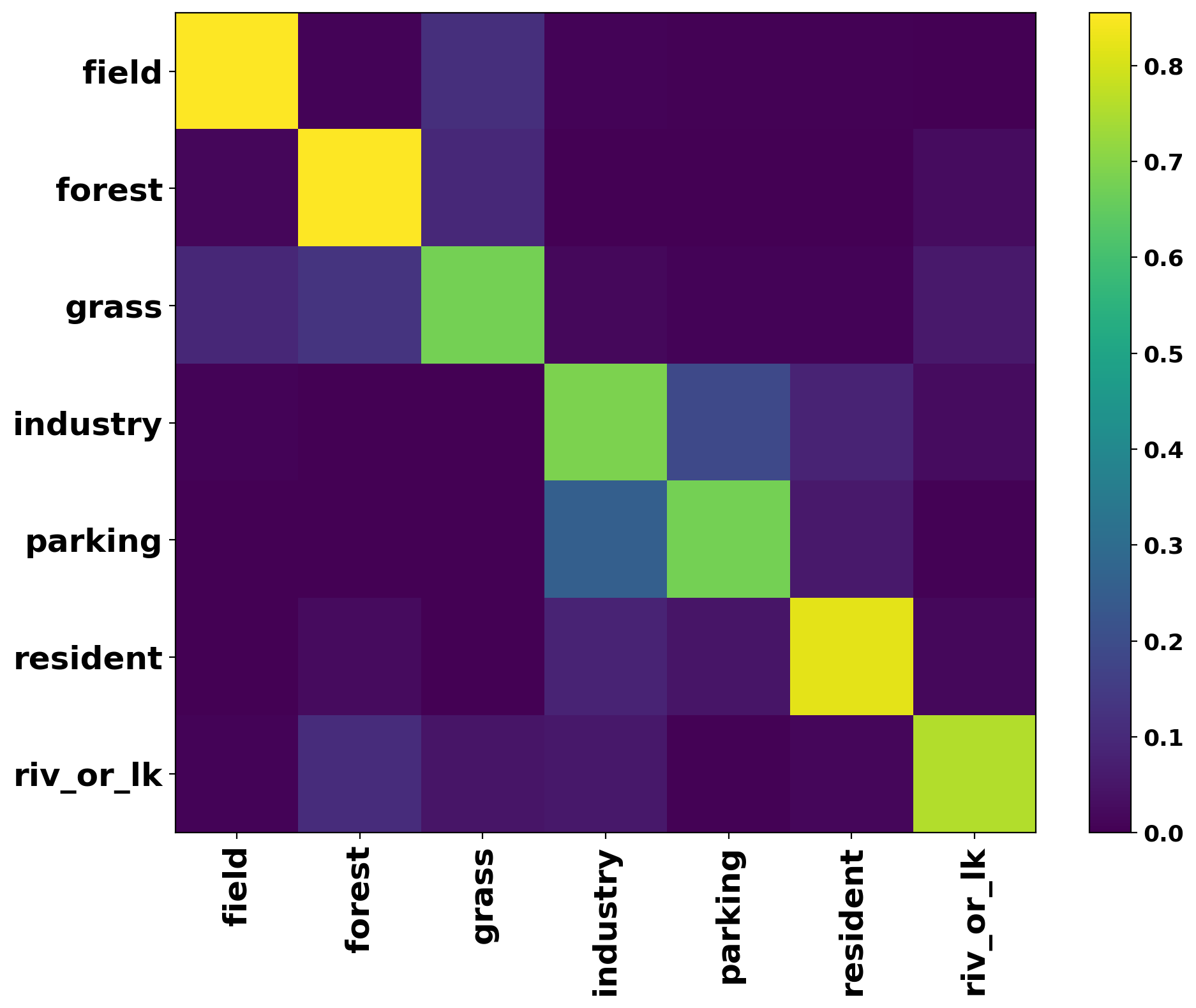}
    \caption{RS2800}
  \end{subfigure}
  \begin{subfigure}[b]{0.24\linewidth}
    \includegraphics[width=\linewidth]{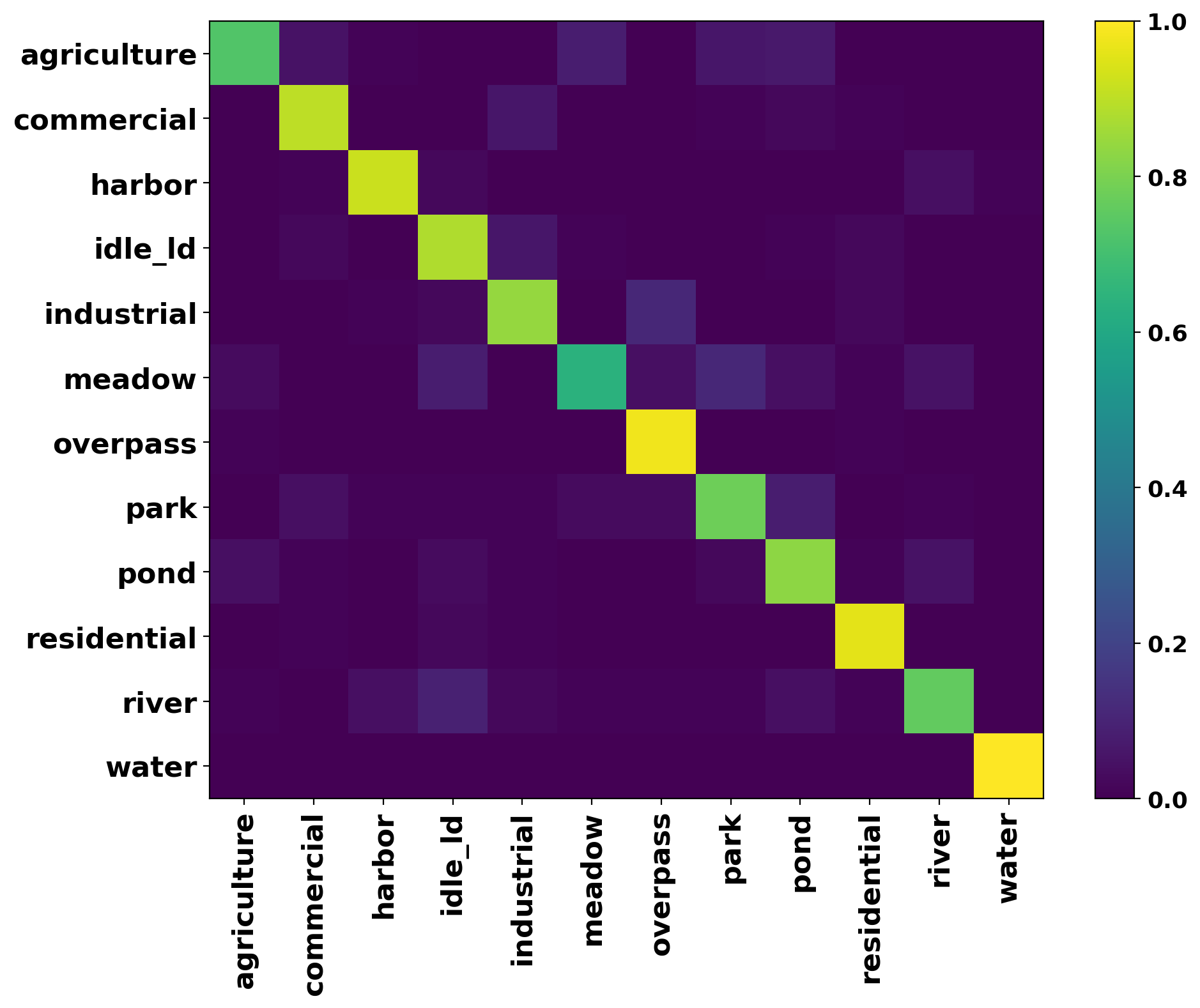}
    \caption{SIRI-WHU}
  \end{subfigure}
  \caption{Normalized confusion matrices on all benchmarks.}
  \label{fig:confusion_core}
\end{figure}

\vspace{1mm}
\noindent{\underline{\bf \em Feature-Space Analysis on Unseen Domains.}} To examine representation transfer, we visualize the learned student embeddings on four unseen datasets using t-SNE. As shown in Fig.~\ref{fig:tsne_unseen}, the embeddings exhibit distinct class-dependent structures across OPTIMAL31, RSC11, RS2800, and SIRI-WHU, despite these datasets being excluded from training. The observed inter-class separation and intra-class concentration indicate that the student preserves discriminative semantic structure under domain shift. Differences in cluster separability align with the quantitative results, indicating stronger transfer under greater semantic alignment.


\vspace{1mm}
\noindent{\underline{\bf \em Class-Wise Performance Analysis.}}
Fig.~\ref{fig:confusion_core} shows the normalized confusion matrices across the core, extended, and unseen datasets. The dominant diagonal structure indicates strong class-wise discrimination, while most errors occur between visually and semantically related categories that share similar spatial layouts, textures, or land-cover characteristics. The remaining confusion is particularly concentrated in fine-grained classes, such as residential-density and related vegetation or land-cover categories. These patterns suggest that the primary errors stem from subtle inter-class visual similarities rather than confusion across unrelated scene types.

\section{Conclusion}

We presented {\sc SatLabel}, a resource-aware framework that integrates adaptive sample acquisition with semi-supervised model adaptation to transform limited satellite labels and abundant unlabeled imagery into compact onboard vision models. By jointly exploiting model uncertainty, class balance, teacher-guided pseudo-labeling, and consistency learning, {\sc SatLabel} achieves competitive or higher Macro-F1 on most core and extended datasets while maintaining strong transfer to unseen domains. Beyond predictive performance, the Balanced student substantially reduces deployment overhead: compared with RemoteCLIP, it uses approximately $13.5\times$ fewer parameters, requires $38\%$ fewer FLOPs, and delivers approximately $2\times$ higher GPU-forward throughput, while reducing energy consumption per image on three of four efficiency benchmarks. The optional MoE+GCN variant increases representation capacity while remaining lightweight. Overall, {\sc SatLabel} provides label-efficient, resource-aware
onboard satellite annotation.

\clearpage

\bibliographystyle{IEEEtran}
\bibliography{references}

@inproceedings{he2016deep,
  title={Deep residual learning for image recognition},
  author={He, Kaiming and Zhang, Xiangyu and Ren, Shaoqing and Sun, Jian},
  booktitle={Proceedings of the IEEE conference on computer vision and pattern recognition},
  pages={770--778},
  year={2016}
}

@article{liu2024remoteclip,
  title={Remoteclip: A vision language foundation model for remote sensing},
  author={Liu, Fan and Chen, Delong and Guan, Zhangqingyun and Zhou, Xiaocong and Zhu, Jiale and Ye, Qiaolin and Fu, Liyong and Zhou, Jun},
  journal={IEEE Transactions on Geoscience and Remote Sensing},
  volume={62},
  pages={1--16},
  year={2024},
  publisher={IEEE}
}

@article{sohn2020fixmatch,
  title={Fixmatch: Simplifying semi-supervised learning with consistency and confidence},
  author={Sohn, Kihyuk and Berthelot, David and Carlini, Nicholas and Zhang, Zizhao and Zhang, Han and Raffel, Colin and Cubuk, Ekin Dogus and Kurakin, Alexey and Li, Chun-Liang},
  journal={Advances in neural information processing systems},
  volume={33},
  pages={596--608},
  year={2020}
}

@article{tarvainen2017mean,
  title={Mean teachers are better role models: Weight-averaged consistency targets improve semi-supervised deep learning results},
  author={Tarvainen, Antti and Valpola, Harri},
  journal={Advances in neural information processing systems},
  volume={30},
  year={2017}
}

@article{cuturi2013sinkhorn,
  title={Sinkhorn distances: Lightspeed computation of optimal transport},
  author={Cuturi, Marco},
  journal={Advances in neural information processing systems},
  volume={26},
  year={2013}
}

@inproceedings{li2018deeper,
  title={Deeper insights into graph convolutional networks for semi-supervised learning},
  author={Li, Qimai and Han, Zhichao and Wu, Xiao-Ming},
  booktitle={Proceedings of the AAAI conference on artificial intelligence},
  volume={32},
  number={1},
  year={2018}
}

@inproceedings{radford2021learning,
  title={Learning transferable visual models from natural language supervision},
  author={Radford, Alec and Kim, Jong Wook and Hallacy, Chris and Ramesh, Aditya and Goh, Gabriel and Agarwal, Sandhini and Sastry, Girish and Askell, Amanda and Mishkin, Pamela and Clark, Jack and others},
  booktitle={International conference on machine learning},
  pages={8748--8763},
  year={2021},
  organization={PmLR}
}

@article{li2023rs,
  title={RS-CLIP: Zero shot remote sensing scene classification via contrastive vision-language supervision},
  author={Li, Xiang and Wen, Congcong and Hu, Yuan and Zhou, Nan},
  journal={International Journal of Applied Earth Observation and Geoinformation},
  volume={124},
  pages={103497},
  year={2023},
  publisher={Elsevier}
}

@article{cacciarelli2024active,
  title={Active learning for data streams: a survey},
  author={Cacciarelli, Davide and Kulahci, Murat},
  journal={Machine Learning},
  volume={113},
  number={1},
  pages={185--239},
  year={2024},
  publisher={Springer}
}

@article{thoreau2022active,
  title={Active learning for hyperspectral image classification: A comparative review},
  author={Thoreau, Romain and Achard, Veronique and Risser, Laurent and Berthelot, Beatrice and Briottet, Xavier},
  journal={IEEE geoscience and remote sensing magazine},
  volume={10},
  number={3},
  pages={256--278},
  year={2022},
  publisher={IEEE}
}

@article{wang2023samrs,
  title={Samrs: Scaling-up remote sensing segmentation dataset with segment anything model},
  author={Wang, Di and Zhang, Jing and Du, Bo and Xu, Minqiang and Liu, Lin and Tao, Dacheng and Zhang, Liangpei},
  journal={Advances in Neural Information Processing Systems},
  volume={36},
  pages={8815--8827},
  year={2023}
}

@book{peyre2019computational,
  title={Computational optimal transport: With applications to data science},
  author={Peyr{\'e}, Gabriel and Cuturi, Marco},
  year={2019},
  publisher={Now Foundations and Trends}
}

@article{dietmuller2024fitnets,
  title={FitNets: An adaptive framework to learn accurate traffic distributions},
  author={Dietm{\"u}ller, Alexander and Alcoz, Albert Gran and Vanbever, Laurent},
  journal={arXiv preprint arXiv:2405.10931},
  year={2024}
}

@inproceedings{le2025semantic,
  title={Semantic knowledge distillation for onboard satellite earth observation image classification},
  author={Le, Thanh-Dung and Ha, Vu Nguyen and Nguyen, Ti Ti and Eappen, Geoffrey and Thiruvasagam, Prabhu and Chou, Hong-fu and Tran, Duc-Dung and Garces-Socarras, Luis M and Gonzalez-Rios, Jorge L and Merlano-Duncan, Juan Carlos and others},
  booktitle={2025 IEEE International Conference on Machine Learning for Communication and Networking (ICMLCN)},
  pages={1--6},
  year={2025},
  organization={IEEE}
}

@article{hinton2015distilling,
  title={Distilling the knowledge in a neural network},
  author={Hinton, Geoffrey and Vinyals, Oriol and Dean, Jeff},
  journal={arXiv preprint arXiv:1503.02531},
  year={2015}
}

@article{xia2017aid,
  title={AID: A benchmark data set for performance evaluation of aerial scene classification},
  author={Xia, Gui-Song and Hu, Jingwen and Hu, Fan and Shi, Baoguang and Bai, Xiang and Zhong, Yanfei and Zhang, Liangpei and Lu, Xiaoqiang},
  journal={IEEE Transactions on Geoscience and Remote Sensing},
  volume={55},
  number={7},
  pages={3965--3981},
  year={2017},
  publisher={IEEE}
}

@inproceedings{yang2010bag,
  title={Bag-of-visual-words and spatial extensions for land-use classification},
  author={Yang, Yi and Newsam, Shawn},
  booktitle={Proceedings of the 18th SIGSPATIAL international conference on advances in geographic information systems},
  pages={270--279},
  year={2010}
}

@article{helber2019eurosat,
  title={Eurosat: A novel dataset and deep learning benchmark for land use and land cover classification},
  author={Helber, Patrick and Bischke, Benjamin and Dengel, Andreas and Borth, Damian},
  journal={IEEE Journal of Selected Topics in Applied Earth Observations and Remote Sensing},
  volume={12},
  number={7},
  pages={2217--2226},
  year={2019},
  publisher={IEEE}
}

@article{roy2023multimodal,
  title={Multimodal fusion transformer for remote sensing image classification},
  author={Roy, Swalpa Kumar and Deria, Ankur and Hong, Danfeng and Rasti, Behnood and Plaza, Antonio and Chanussot, Jocelyn},
  journal={IEEE Transactions on Geoscience and Remote Sensing},
  volume={61},
  pages={1--20},
  year={2023},
  publisher={IEEE}
}

@article{zhang2024rs5m,
  title={RS5M and GeoRSCLIP: A large-scale vision-language dataset and a large vision-language model for remote sensing},
  author={Zhang, Zilun and Zhao, Tiancheng and Guo, Yulong and Yin, Jianwei},
  journal={IEEE Transactions on Geoscience and Remote Sensing},
  volume={62},
  pages={1--23},
  year={2024},
  publisher={IEEE}
}

@inproceedings{mall2024remote,
  title={Remote sensing vision-language foundation models without annotations via ground remote alignment},
  author={Mall, Utkarsh Kumar and Phoo, Cheng Perng and Liu, Meilin and Vondrick, Carl and Hariharan, Bharath and Bala, Kavita},
  booktitle={International Conference on Learning Representations},
  volume={2024},
  pages={49294--49314},
  year={2024}
}

@article{cheng2017remote,
  title={Remote sensing image scene classification: Benchmark and state of the art},
  author={Cheng, Gong and Han, Junwei and Lu, Xiaoqiang},
  journal={Proceedings of the IEEE},
  volume={105},
  number={10},
  pages={1865--1883},
  year={2017},
  publisher={IEEE}
}

@article{zhou2018patternnet,
  title={PatternNet: A benchmark dataset for performance evaluation of remote sensing image retrieval},
  author={Zhou, Weixun and Newsam, Shawn and Li, Congmin and Shao, Zhenfeng},
  journal={ISPRS journal of photogrammetry and remote sensing},
  volume={145},
  pages={197--209},
  year={2018},
  publisher={Elsevier}
}

@article{zhao2016dirichlet,
  title={Dirichlet-derived multiple topic scene classification model fusing heterogeneous features for high spatial resolution remote sensing imagery},
  author={Zhao, Bei and Zhong, Yanfei and Xia, GS and Zhang, Liangpei},
  journal={IEEE Trans. Geosci. Remote Sens},
  volume={54},
  number={4},
  pages={2108--2123},
  year={2016}
}

@article{liu2025efficient,
  title={Efficient remote sensing image classification using the novel STConvNeXt convolutional network},
  author={Liu, Bo and Zhan, Chenmei and Guo, Cheng and Liu, Xiaobo and Ruan, Shufen},
  journal={Scientific Reports},
  volume={15},
  number={1},
  pages={8406},
  year={2025},
  publisher={Nature Publishing Group UK London}
}

@article{zhao2016feature,
  title={Feature significance-based multibag-of-visual-words model for remote sensing image scene classification},
  author={Zhao, Lijun and Tang, Ping and Huo, Lianzhi},
  journal={Journal of Applied Remote Sensing},
  volume={10},
  number={3},
  pages={035004--035004},
  year={2016},
  publisher={Society of Photo-Optical Instrumentation Engineers}
}

@article{wang2018scene,
  title={Scene classification with recurrent attention of VHR remote sensing images},
  author={Wang, Qi and Liu, Shaoteng and Chanussot, Jocelyn and Li, Xuelong},
  journal={IEEE Transactions on Geoscience and Remote Sensing},
  volume={57},
  number={2},
  pages={1155--1167},
  year={2018},
  publisher={IEEE}
}

@article{li2020rsi,
  title={RSI-CB: A large-scale remote sensing image classification benchmark using crowdsourced data},
  author={Li, Haifeng and Dou, Xin and Tao, Chao and Wu, Zhixiang and Chen, Jie and Peng, Jian and Deng, Min and Zhao, Ling},
  journal={Sensors},
  volume={20},
  number={6},
  pages={1594},
  year={2020},
  publisher={MDPI}
}

@inproceedings{xia2010structural,
  title={Structural high-resolution satellite image indexing},
  author={Xia, Gui-Song and Yang, Wen and Delon, Julie and Gousseau, Yann and Sun, Hong and Ma{\^\i}tre, Henri},
  booktitle={ISPRS TC VII Symposium-100 Years ISPRS},
  volume={38},
  pages={298--303},
  year={2010}
}

@article{ostman2023decentralised,
  title={Decentralised Semi-supervised Onboard Learning for Scene Classification in Low-Earth Orbit},
  author={\O stman, Johan and G{\'o}mez, Pablo and Shreenath, Vinutha Magal and Meoni, Gabriele},
  journal={arXiv preprint arXiv:2305.04059},
  year={2023}
}

@article{karwowska2022using,
  title={Using super-resolution algorithms for small satellite imagery: A systematic review},
  author={Karwowska, Kinga and Wierzbicki, Damian},
  journal={IEEE Journal of Selected Topics in Applied Earth Observations and Remote Sensing},
  volume={15},
  pages={3292--3312},
  year={2022},
  publisher={IEEE}
}

@inproceedings{li2024glh,
  title={GLH-water: A large-scale dataset for global surface water detection in large-size very-high-resolution satellite imagery},
  author={Li, Yansheng and Dang, Bo and Li, Wanchun and Zhang, Yongjun},
  booktitle={Proceedings of the AAAI Conference on Artificial Intelligence},
  volume={38},
  number={20},
  pages={22213--22221},
  year={2024}
}

@inproceedings{capogrosso2026tinymllunar,
  title={TinyML Enhances CubeSat Mission Capabilities},
  author={Capogrosso, Luigi and Magno, Michele},
  booktitle={Proc.\ of the ACM/IEEE Int. Conf.\ on Cyber-Physical Systems (ICCPS)},
  year={2026}
}

@inproceedings{elmahallawy2024stitching,
  title={Stitching Satellites to the Edge: Pervasive and Efficient Federated {LEO} Satellite Learning},
  author={Elmahallawy, Mohamed and Luo, Tie},
  booktitle={Proc.\ IEEE Int.\ Conf.\ on Pervasive Computing and Communications (PerCom)},
  year={2024}
}

@inproceedings{le2024semantic,
  title={Semantic knowledge distillation for onboard satellite earth observation image classification},
  author={Le, Thanh-Dung and Ha, Vu Nguyen and Nguyen, Ti Ti and Eappen, Geoffrey and Thiruvasagam, Prabhu and Chou, Hong-fu and Tran, Duc-Dung and Garces-Socarras, Luis M and Gonzalez-Rios, Jorge L and Merlano-Duncan, Juan Carlos and others},
  booktitle={2025 IEEE International Conference on Machine Learning for Communication and Networking (ICMLCN)},
  pages={1--6},
  year={2025},
  organization={IEEE}
}

@article{liu2023remoteclip,
  title={RemoteCLIP: A Vision Language Foundation Model for Remote Sensing},
  author={Liu, Fan and Chen, Delong and Guan, Zhangqingyun and Zhou, Xiaocong and Zhu, Jiale and Ye, Qiaolin and Fu, Liyong and Zhou, Jun},
  journal={IEEE Trans.\ on Geoscience and Remote Sensing},
  note={Accepted},
  year={2024}
}

@article{li2025grace,
  title={Enabling Near-realtime Remote Sensing via Satellite–Ground Collaboration of Large Vision–Language Models},
  author={Li, Zihan and Yang, Jiahao and Zhang, Yuxin and Chen, Zhe and Gao, Yue},
  journal={arXiv preprint arXiv:2510.24242},
  year={2025}
}

@article{le2026onboard,
  title={Onboard satellite image classification for earth observation: A comparative study of ViT models},
  author={Le, Thanh-Dung and Ha, Vu Nguyen and Nguyen, Ti Ti and Tran, Duc-Dung and Nguyen-Kha, Hung and Garces-Socarras, Luis M and Merlano-Duncan, Juan Carlos and Chatzinotas, Symeon},
  journal={IEEE Transactions on Geoscience and Remote Sensing},
  year={2026},
  publisher={IEEE}
}
\clearpage
\appendix
\section{Additional Methodological Details}
\label{supp:method}
 
\subsection{Detailed Augmentation Strategy}
\label{supp:augmentation}
 
The weak/strong augmentation strategy is a central design choice in
{\sc SatLabel} and deserves elaboration beyond the main paper description.
 
\noindent{\bf{\textbf{Weak Augmentation $\Aw(\cdot)$.}}}
Weak augmentation applies mild, semantics-preserving transformations to
produce a view $x^w = \Aw(x)$ that closely resembles the original image
in terms of scene content.
The transformations applied are:
\begin{itemize}[leftmargin=*,itemsep=2pt]
  \item \textbf{Random resized crop}: images are cropped to a random region
        (scale 0.8--1.0 of the original area) and resized to $224\!\times\!224$.
  \item \textbf{Random horizontal flip}: applied with probability 0.5.
\end{itemize}
These transformations preserve the overall scene structure, ensuring that
the teacher's prediction on $x^w$ closely reflects its prediction on the
original image.
The weak view is used exclusively for teacher forward passes and never
directly for student training.
 
\noindent{\bf{\textbf{Strong Augmentation $\As(\cdot)$.}}}
Strong augmentation applies an aggressive sequence of appearance-level
transformations to produce $x^s = \As(x)$, substantially altering the
image's visual characteristics while preserving its semantic category:
\begin{itemize}[leftmargin=*,itemsep=2pt]
  \item \textbf{Random color jitter}: brightness, contrast, saturation,
        and hue each perturbed with strength 0.4 and probability 0.8.
  \item \textbf{Random grayscale conversion}: applied with probability 0.2.
  \item \textbf{Gaussian blur}: kernel size $23\!\times\!23$, $\sigma\!\in\![0.1, 2.0]$,
        applied with probability 0.5.
  \item \textbf{Random erasing}: a random rectangular region (area ratio
        0.02--0.33, aspect ratio 0.3--3.3) is replaced with the mean pixel
        value; applied with probability 0.5.
  \item \textbf{Random resized crop and horizontal flip}: same as weak.
\end{itemize}
The student is trained on $x^s$ and must reproduce the teacher's
prediction on $x^w$, enforcing invariance to appearance changes
irrelevant to scene category.
This asymmetric design---stable inputs for the teacher, distorted inputs
for the student---is the core mechanism of FixMatch-style consistency
regularization~\cite{sohn2020fixmatch}.
 
\subsection{Training Objectives in Full}
\label{supp:loss}
 
The main paper presents equations for $\mathcal{L}_{\mathrm{KD}}$,
$\mathcal{L}_{\mathrm{cons}}$, and the combined objective.
Here we give the complete formulation including the supervised loss.
 
\noindent{\bf{\textbf{Supervised Cross-Entropy.}}}
For images in the labeled set $\Dl^r$, standard cross-entropy anchors the
student to ground-truth human annotations and prevents drift toward
spurious patterns introduced by pseudo-label noise:
\begin{equation}
  \mathcal{L}_{\mathrm{CE}}
  = -\mathbb{E}_{(x,y)\sim\Dl}\!\Bigl[
      \textstyle\sum_c y_c \log\Ps(y_c \mid x)\Bigr].
  \label{eq:lce}
\end{equation}
Without this term, the student would depend entirely on the teacher's
pseudo-labels, which---while generally accurate---can be wrong for scenes
that are ambiguous even to the foundation model.
 
\noindent{\bf{\textbf{Knowledge Distillation.}}}
For all images (labeled and unlabeled), the student minimizes the
temperature-scaled KL divergence from the teacher's soft distribution:
\begin{equation}
  \mathcal{L}_{\mathrm{KD}}
  = T^2\,\KLdiv\!\bigl(
      \Pt(y \mid x;\,T)\;\|\;\Ps(y \mid x;\,T)\bigr).
  \label{eq:lkd}
\end{equation}
The temperature $T\!=\!2$ softens both distributions, amplifying
differences among non-dominant classes and encoding inter-class semantic
similarity (\eg, forest is closer to farmland than to airport).
The $T^2$ factor compensates for the reduced gradient magnitude at
higher temperatures, keeping $\mathcal{L}_{\mathrm{KD}}$ on a comparable
scale to $\mathcal{L}_{\mathrm{CE}}$.
 
\noindent{\bf\{\textbf{Consistency Regularization.}}
For unlabeled images whose teacher confidence exceeds $\tau$, the student
is penalized when its prediction on the strongly augmented view $x^s$
differs from the teacher's prediction on $x^w$:
\begin{equation}
  \mathcal{L}_{\mathrm{cons}}
  = \mathbb{E}_{x \in \Du'}\!\Bigl[
      H\!\bigl(\Pt(y \mid x^w),\,\Ps(y \mid x^s)\bigr)\Bigr],
  \label{eq:lcons}
\end{equation}
where $\Du'\!=\!\{x\!\in\!\Du:\max_y\Pt(y|x^w)\!>\!\tau\}$ and $H$
denotes cross-entropy.
 
\noindent{\bf{\textbf{Combined Objective.}}}
The student is trained with the weighted sum:
\begin{equation}
  \mathcal{L}
  = \lambda_{\mathrm{CE}}\,\mathcal{L}_{\mathrm{CE}}
  + \lambda_{\mathrm{KD}}\,\mathcal{L}_{\mathrm{KD}}
  + \lambda_{\mathrm{cons}}\,\mathcal{L}_{\mathrm{cons}},
  \label{eq:total}
\end{equation}
with $\lambda_{\mathrm{CE}}\!=\!0.3$, $\lambda_{\mathrm{KD}}\!=\!0.7$.
The higher weight on $\mathcal{L}_{\mathrm{KD}}$ reflects the richer
information content of the teacher's soft distributions.
$\lambda_{\mathrm{cons}}$ is activated from round~3 onward, once the
student has reached sufficient reliability to avoid reinforcing early
errors.
 
\section{MoE and BatchGCN Architecture Details}
\label{supp:moe_gcn}
 
\subsection{Mixture-of-Experts Routing}
\label{supp:moe}
 
The MoE module addresses the visual heterogeneity of satellite imagery by
learning $K$ specialized expert networks.
Each expert independently transforms the backbone feature $z$:
\begin{equation}
  e_k = E_k(z), \quad k = 1, \ldots, K,
  \label{eq:experts}
\end{equation}
where each $E_k$ is a two-layer MLP with ReLU activation.
Through gradient-driven learning, experts develop complementary
specializations: one may focus on the geometric regularity of urban
structures, another on the spectral texture of vegetation, and a third
on the smooth homogeneous appearance of water bodies.
 
A lightweight gating network predicts routing logits over the experts.
Standard softmax gating suffers from expert collapse---routing nearly all
inputs to one dominant expert---which is particularly harmful in remote
sensing where diverse scene categories require genuine expert diversity.
We apply Sinkhorn--Knopp normalization to the routing logit matrix
$G \in \mathbb{R}^{B \times K}$, alternating row and column normalizations
to produce a doubly-stochastic assignment that enforces balanced expert
utilization across the mini-batch.
Only the top-2 experts are activated per sample (sparse routing).
The final MoE output is:
\begin{equation}
  z_{\mathrm{moe}} = \textstyle\sum_{k=1}^{K} g_k\, E_k(z),
  \label{eq:moe}
\end{equation}
where $\{g_k\}$ are the Sinkhorn-normalized routing weights summing to 1.
In our experiments we use $K\!=\!4$ experts.

\begin{table}[t]
  \centering
  \caption{Dataset statistics. Support denotes images per class
  (min / max / mean).}
  \label{tab:dataset_stats}
  \small
  \setlength{\tabcolsep}{4pt}
  \renewcommand{\arraystretch}{1.12}
  \begin{tabular}{llrrrrr}
    \toprule
    \textbf{Group} & \textbf{Dataset}
    & \textbf{Classes} & \textbf{Images}
    & \textbf{Min} & \textbf{Max} & \textbf{Mean}\\
    \midrule
    \rowcolor{rowblue}
    Core     & AID        & 30 & 10{,}000 & 220 & 420 & 333\\
    \rowcolor{rowblue}
    Core     & EuroSAT    & 10 & 27{,}000 & 2000 & 3000 & 2700\\
    \rowcolor{rowblue}
    Core     & UC~Merced  & 21 &  2{,}100 & 100 & 100 & 100\\
    \midrule
    \rowcolor{rowgreen}
    Extended & RESISC45   & 45 & 31{,}500 & 700 & 700 & 700\\
    \rowcolor{rowgreen}
    Extended & PatternNet & 38 & 30{,}400 & 800 & 800 & 800\\
    \rowcolor{rowgreen}
    Extended & RSI-CB256  &  7 & 24{,}747 & 884 & 6258 & 3535\\
    \rowcolor{rowgreen}
    Extended & WHU-RS19   & 19 &  1{,}005 &  50 &   61 &   53\\
    \midrule
    Unseen   & OPTIMAL31  & 31 & 1{,}860 &  60 &  60 &  60\\
    Unseen   & RSC11      & 11 & 1{,}232 &  80 & 142 & 112\\
    Unseen   & RS2800     &  7 & 2{,}800 & 400 & 400 & 400\\
    Unseen   & SIRI-WHU   & 12 & 2{,}400 & 200 & 200 & 200\\
    \bottomrule
  \end{tabular}
\end{table}

\

\begin{table}[t]
\centering
\caption{Student training configurations used in the ablation study.}
\label{tab:student_configs}
\small
\begin{tabular}{p{0.23\linewidth}p{0.68\linewidth}}
\toprule
Experiment & Training domains \\
\midrule
01 Baseline & AID, UC Merced, EuroSAT \\
02 + RESISC45 & AID, UC Merced, EuroSAT, RESISC45 \\
03 + PatternNet & AID, UC Merced, EuroSAT, PatternNet \\
04 + WHU-RS19 & AID, UC Merced, EuroSAT, WHU-RS19 \\
05 + RSI-CB256 & AID, UC Merced, EuroSAT, RSI-CB256 \\
06 All Eval Mix & AID, UC Merced, EuroSAT, RESISC45, PatternNet, WHU-RS19, RSI-CB256 \\
\bottomrule
\end{tabular}
\end{table}
\begin{table}[!t]
  \centering
  \caption{ResNet-18 MoE+GCN student -- best checkpoint results on all
  eleven datasets. Acc: accuracy. F1: Macro-F1. $\Delta$: student minus RemoteCLIP (Accuracy).}
  \label{tab:r18_full_results}
  \small
  \setlength{\tabcolsep}{5pt}
  \resizebox{\linewidth}{!}{
  \renewcommand{\arraystretch}{1.15}
  \begin{tabular}{llrrrr}
    \toprule
    \textbf{Group} & \textbf{Dataset}
    & \textbf{RC Acc.} & \textbf{Stud.\ Acc.}
    & \textbf{Stud.\ F1} & \textbf{$\Delta$ Acc.}\\
    \midrule
    \rowcolor{rowblue}
    Core & AID
      & 86.83 & \textbf{94.41} & \textbf{94.76}
      & \textcolor{goodgreen}{+7.58}\\
    \rowcolor{rowblue}
    Core & EuroSAT
      & 32.46 & \textbf{95.37} & \textbf{95.95}
      & \textcolor{goodgreen}{+62.91}\\
    \rowcolor{rowblue}
    Core & UC~Merced
      & 77.95 & \textbf{91.00} & \textbf{94.01}
      & \textcolor{goodgreen}{+13.05}\\
    \midrule
    \rowcolor{rowgreen}
    Extended & RESISC45
      & 68.38 & \textbf{86.63} & \textbf{88.24}
      & \textcolor{goodgreen}{+18.25}\\
    \rowcolor{rowgreen}
    Extended & PatternNet
      & 57.52 & \textbf{95.90} & \textbf{96.26}
      & \textcolor{goodgreen}{+38.38}\\
    \rowcolor{rowgreen}
    Extended & RSI-CB256
      & 50.38 & \textbf{87.01} & \textbf{91.36}
      & \textcolor{goodgreen}{+36.63}\\
    \rowcolor{rowgreen}
    Extended & WHU-RS19
      & \textbf{94.33} & 77.41 & 84.81
      & \textcolor{badred}{$-$16.92}\\
    \midrule
    Unseen & OPTIMAL31
      & 77.47 & \textbf{83.66} & \textbf{83.74}
      & \textcolor{goodgreen}{+6.19}\\
    Unseen & RSC11
      & 62.91 & \textbf{86.20} & \textbf{86.10}
      & \textcolor{goodgreen}{+23.29}\\
    Unseen & RS2800
      & 63.61 & \textbf{76.00} & \textbf{76.03}
      & \textcolor{goodgreen}{+12.39}\\
    Unseen & SIRI-WHU
      & 61.54 & \textbf{85.17} & \textbf{84.96}
      & \textcolor{goodgreen}{+23.63}\\
    \bottomrule
  \end{tabular}}
\end{table}
\begin{table}[!t]
  \centering
  \caption{RemoteCLIP zero-shot vs.\ best ResNet-18 MoE+GCN student
  (Accuracy\,/\,Macro-F1, \%).
  $\Delta$: student minus RemoteCLIP (Accuracy).
  \textbf{Bold}: best per dataset.}
  \label{tab:r18_vs_rc}
  \small
  \setlength{\tabcolsep}{4pt}
  \renewcommand{\arraystretch}{1.15}
  \begin{tabular}{lrrr}
    \toprule
    \textbf{Dataset} & \textbf{RemoteCLIP} & \textbf{Student (R18)}
    & \textbf{$\Delta$ Acc.}\\
    \midrule
    AID        & 86.83 / 86.06 & \textbf{94.41} / \textbf{94.76}
               & \textcolor{goodgreen}{+7.58}\\
    EuroSAT    & 32.46 / 29.87 & \textbf{95.37} / \textbf{95.95}
               & \textcolor{goodgreen}{+62.91}\\
    UC~Merced  & 77.95 / 74.63 & \textbf{91.00} / \textbf{94.01}
               & \textcolor{goodgreen}{+13.05}\\
    RESISC45   & 68.38 / 66.53 & \textbf{86.63} / \textbf{88.24}
               & \textcolor{goodgreen}{+18.25}\\
    PatternNet & 57.52 / 52.15 & \textbf{95.90} / \textbf{96.26}
               & \textcolor{goodgreen}{+38.38}\\
    RSI-CB256  & 50.38 / 42.60 & \textbf{87.01} / \textbf{91.36}
               & \textcolor{goodgreen}{+36.63}\\
    WHU-RS19   & \textbf{94.33} / \textbf{94.33} & 77.41 / 84.81
               & \textcolor{badred}{$-$16.92}\\
    \bottomrule
  \end{tabular}
\end{table}
 \begin{table}[!t]
  \centering
  \caption{AID -- ResNet-18 student per-class results (\%). Sorted by F1-Score descending.}
  \label{tab:r18_perclass_aid}
  \scriptsize
  \setlength{\tabcolsep}{3pt}
  \renewcommand{\arraystretch}{1.1}
    \resizebox{\linewidth}{!}{
  \begin{tabular}{lrrrr}
    \toprule
    \textbf{Class} & \textbf{Support} & \textbf{Precision} & \textbf{Recall} & \textbf{F1-Score}\\
    \midrule
    Forest & 250 & 99.6 & 99.6 & 100.0\\
    Beach & 400 & 99.0 & 99.2 & 99.7\\
    Meadow & 280 & 99.3 & 98.9 & 99.6\\
    Parking & 390 & 98.5 & 99.2 & 99.4\\
    Mountain & 340 & 98.8 & 98.5 & 99.2\\
    Viaduct & 420 & 97.2 & 100.0 & 99.1\\
    Farmland & 370 & 98.6 & 97.8 & 98.8\\
    Pond & 420 & 97.6 & 98.1 & 98.4\\
    Playground & 370 & 98.6 & 97.0 & 98.4\\
    Storagetanks & 360 & 96.7 & 98.3 & 98.1\\
    Baseballfield & 220 & 95.6 & 99.1 & 97.9\\
    Stadium & 290 & 96.3 & 98.3 & 97.8\\
    Port & 380 & 95.6 & 98.2 & 97.4\\
    Desert & 300 & 97.3 & 96.0 & 97.2\\
    Bareland & 310 & 95.3 & 97.7 & 97.0\\
    River & 410 & 99.2 & 93.9 & 97.0\\
    Sparseresidential & 300 & 99.6 & 93.3 & 96.9\\
    Airport & 360 & 96.1 & 95.6 & 96.3\\
    Bridge & 360 & 91.5 & 98.9 & 95.6\\
    Denseresidential & 410 & 95.2 & 91.7 & 93.9\\
    Mediumresidential & 290 & 90.8 & 95.9 & 93.8\\
    Industrial & 390 & 91.8 & 92.3 & 92.6\\
    Railwaystation & 260 & 89.3 & 93.5 & 91.9\\
    Commercial & 350 & 91.3 & 86.6 & 89.3\\
    Church & 240 & 80.4 & 99.2 & 89.3\\
    Park & 350 & 89.8 & 85.4 & 88.0\\
    Square & 330 & 93.3 & 80.0 & 86.6\\
    Center & 260 & 89.6 & 82.7 & 86.5\\
    School & 300 & 77.9 & 90.3 & 84.1\\
    Resort & 290 & 89.8 & 76.2 & 82.9\\
    \midrule
    \textbf{Macro} & --- & 94.3 & 94.4 & \textbf{94.76}\\
    \bottomrule
  \end{tabular}}
\end{table}
\begin{table}[!t]
  \centering
  \caption{EuroSAT -- ResNet-18 student per-class results (\%). Sorted by F1-Score descending. }
  \label{tab:r18_perclass_eurosat}
  \scriptsize
  \setlength{\tabcolsep}{3pt}
  \renewcommand{\arraystretch}{1.1}
    \resizebox{\linewidth}{!}{
  \begin{tabular}{lrrrr}
    \toprule
    \textbf{Class} & \textbf{Support} & \textbf{Precision} & \textbf{Recall} & \textbf{F1-Score}\\
    \midrule
    Residential & 3000 & 99.5 & 99.0 & 98.5\\
    Industrial & 2500 & 98.9 & 99.4 & 98.5\\
    Sea Or Lake & 3000 & 98.8 & 99.0 & 98.2\\
    Forest & 3000 & 98.2 & 98.9 & 97.9\\
    Highway & 2500 & 98.4 & 94.6 & 95.8\\
    River & 2500 & 93.4 & 98.8 & 95.3\\
    Annual Crop & 3000 & 95.5 & 95.9 & 95.0\\
    Herbaceous Vegetation & 3000 & 91.3 & 98.0 & 93.8\\
    Pasture & 2000 & 98.8 & 90.1 & 93.6\\
    Permanent Crop & 2500 & 96.0 & 91.2 & 92.9\\
    \midrule
    \textbf{Macro} & --- & 96.9 & 96.5 & \textbf{95.95}\\
    \bottomrule
  \end{tabular}}
\end{table}
\begin{table}[!t]
  \centering
  \caption{UC~Merced -- ResNet-18 student per-class results (\%). Sorted by F1-Score descending. }
  \label{tab:r18_perclass_ucm}
  \scriptsize
  \setlength{\tabcolsep}{3pt}
  \renewcommand{\arraystretch}{1.1}
    \resizebox{\linewidth}{!}{
  \begin{tabular}{lrrrr}
    \toprule
    \textbf{Class} & \textbf{Support} & \textbf{Precision} & \textbf{Recall} & \textbf{F1-Score}\\
    \midrule
    Beach & 100 & 100.0 & 100.0 & 99.4\\
    Harbor & 100 & 99.0 & 100.0 & 98.9\\
    Parkinglot & 100 & 100.0 & 99.0 & 98.9\\
    Tenniscourt & 100 & 100.0 & 99.0 & 98.9\\
    Agricultural & 100 & 99.0 & 99.0 & 98.4\\
    Baseballdiamond & 100 & 100.0 & 98.0 & 98.4\\
    Sparseresidential & 100 & 100.0 & 96.0 & 97.4\\
    Airplane & 100 & 96.1 & 99.0 & 97.0\\
    Runway & 100 & 98.0 & 97.0 & 96.9\\
    Chaparral & 100 & 100.0 & 94.0 & 96.3\\
    Forest & 100 & 93.5 & 100.0 & 96.1\\
    River & 100 & 94.2 & 98.0 & 95.5\\
    Intersection & 100 & 94.2 & 97.0 & 95.0\\
    Golfcourse & 100 & 100.0 & 90.0 & 94.2\\
    Storagetanks & 100 & 96.8 & 92.0 & 93.8\\
    Freeway & 100 & 85.5 & 100.0 & 91.6\\
    Overpass & 100 & 94.7 & 89.0 & 91.2\\
    Buildings & 100 & 92.8 & 90.0 & 90.8\\
    Mobilehomepark & 100 & 74.1 & 100.0 & 84.6\\
    Mediumresidential & 100 & 93.6 & 73.0 & 81.5\\
    Denseresidential & 100 & 83.5 & 76.0 & 79.4\\
    \midrule
    \textbf{Macro} & --- & 95.0 & 94.6 & \textbf{94.01}\\
    \bottomrule
  \end{tabular}}
\end{table}
\begin{table}[!t]
  \centering
  \caption{RESISC45 -- ResNet-18 student per-class results (\%). Sorted by F1-Score descending.}
  \label{tab:r18_perclass_resisc}
  \scriptsize
  \setlength{\tabcolsep}{3pt}
  \renewcommand{\arraystretch}{1.1}
    \resizebox{\linewidth}{!}{
  \begin{tabular}{lrrrr}
    \toprule
    \textbf{Class} & \textbf{Support} & \textbf{Precision} & \textbf{Recall} & \textbf{F1-Score}\\
    \midrule
    Sea Ice & 700 & 99.7 & 98.3 & 97.0\\
    Airplane & 700 & 98.1 & 98.3 & 96.2\\
    Beach & 700 & 96.7 & 97.7 & 95.3\\
    Circular Farmland & 700 & 96.6 & 97.6 & 95.1\\
    Golf Course & 700 & 98.8 & 94.7 & 94.8\\
    Harbor & 700 & 93.7 & 99.6 & 94.6\\
    Chaparral & 700 & 95.4 & 97.3 & 94.4\\
    Cloud & 700 & 97.0 & 95.6 & 94.3\\
    Baseball Diamond & 700 & 93.9 & 98.6 & 94.2\\
    Parking Lot & 700 & 92.8 & 98.3 & 93.6\\
    Storage Tank & 700 & 91.0 & 99.1 & 93.0\\
    Basketball Court & 700 & 97.4 & 92.4 & 92.9\\
    Roundabout & 700 & 98.3 & 91.6 & 92.9\\
    Forest & 700 & 94.8 & 93.6 & 92.3\\
    Runway & 700 & 98.6 & 89.7 & 92.0\\
    Intersection & 700 & 95.7 & 91.6 & 91.7\\
    Airport & 700 & 92.6 & 94.4 & 91.6\\
    Meadow & 700 & 91.5 & 95.4 & 91.5\\
    Desert & 700 & 91.7 & 94.9 & 91.4\\
    Overpass & 700 & 92.4 & 93.3 & 90.9\\
    Thermal Power Station & 700 & 94.4 & 91.3 & 90.9\\
    Mobile Home Park & 700 & 90.0 & 95.3 & 90.7\\
    Island & 700 & 98.5 & 87.1 & 90.6\\
    Bridge & 700 & 87.8 & 97.6 & 90.6\\
    Ship & 700 & 89.6 & 94.3 & 90.0\\
    Tennis Court & 700 & 96.4 & 87.1 & 89.7\\
    Sparse Residential & 700 & 90.1 & 92.7 & 89.6\\
    Ground Track Field & 700 & 83.8 & 99.0 & 88.9\\
    Industrial Area & 700 & 88.1 & 92.3 & 88.3\\
    Rectangular Farmland & 700 & 84.3 & 95.7 & 87.8\\
    Terrace & 700 & 98.0 & 82.3 & 87.6\\
    Freeway & 700 & 87.8 & 90.4 & 87.3\\
    Lake & 700 & 82.7 & 95.7 & 86.9\\
    Railway Station & 700 & 91.6 & 82.9 & 85.3\\
    Railway & 700 & 89.6 & 84.0 & 85.0\\
    Dense Residential & 700 & 86.8 & 84.6 & 83.9\\
    River & 700 & 80.4 & 91.7 & 83.9\\
    Stadium & 700 & 98.9 & 75.0 & 83.6\\
    Snowberg & 700 & 96.7 & 74.3 & 82.3\\
    Medium Residential & 700 & 71.6 & 93.1 & 79.3\\
    Mountain & 700 & 73.4 & 90.1 & 79.3\\
    Wetland & 700 & 97.0 & 64.7 & 76.1\\
    Commercial Area & 700 & 86.1 & 68.1 & 74.5\\
    Church & 700 & 57.8 & 92.3 & 69.6\\
    Palace & 700 & 96.6 & 44.1 & 59.4\\
    \midrule
    \textbf{Macro} & --- & 91.2 & 90.2 & \textbf{88.24}\\
    \bottomrule
  \end{tabular}}
\end{table}
\begin{table}[!t]
  \centering
  \caption{PatternNet -- ResNet-18 student per-class results (\%). Sorted by F1-Score descending. }
  \label{tab:r18_perclass_patternnet}
  \scriptsize
  \setlength{\tabcolsep}{3pt}
  \renewcommand{\arraystretch}{1.1}
     \resizebox{\linewidth}{!}{
  \begin{tabular}{lrrrr}
    \toprule
    \textbf{Class} & \textbf{Support} & \textbf{Precision} & \textbf{Recall} & \textbf{F1-Score}\\
    \midrule
    Forest & 800 & 100.0 & 100.0 & 100.0\\
    Oil Well & 800 & 100.0 & 100.0 & 100.0\\
    Freeway & 800 & 100.0 & 99.7 & 100.0\\
    Airplane & 800 & 99.8 & 99.9 & 100.0\\
    Runway & 800 & 99.9 & 99.7 & 100.0\\
    Beach & 800 & 99.5 & 100.0 & 100.0\\
    Runway Marking & 800 & 99.6 & 99.9 & 100.0\\
    Parking Space & 800 & 99.4 & 100.0 & 100.0\\
    Golf Course & 800 & 100.0 & 99.4 & 100.0\\
    Oil Gas Field & 800 & 100.0 & 99.1 & 100.0\\
    Transformer Station & 800 & 99.3 & 99.7 & 100.0\\
    Baseball Field & 800 & 98.9 & 100.0 & 100.0\\
    River & 800 & 99.9 & 99.0 & 100.0\\
    Tennis Court & 800 & 99.1 & 99.6 & 100.0\\
    Football Field & 800 & 100.0 & 98.6 & 100.0\\
    Solar Panel & 800 & 99.5 & 98.6 & 100.0\\
    Cemetery & 800 & 98.9 & 99.1 & 100.0\\
    Railway & 800 & 98.6 & 99.2 & 100.0\\
    Intersection & 800 & 97.8 & 99.2 & 100.0\\
    Overpass & 800 & 99.1 & 97.7 & 100.0\\
    Shipping Yard & 800 & 98.7 & 97.6 & 100.0\\
    Basketball Court & 800 & 99.6 & 96.6 & 100.0\\
    Chaparral & 800 & 96.2 & 100.0 & 100.0\\
    Christmas Tree Farm & 800 & 98.2 & 96.5 & 99.7\\
    Parking Lot & 800 & 92.2 & 100.0 & 98.2\\
    Closed Road & 800 & 99.6 & 92.0 & 98.0\\
    Bridge & 800 & 92.1 & 98.9 & 97.7\\
    Dense Residential & 800 & 96.0 & 93.6 & 97.1\\
    Nursing Home & 800 & 99.9 & 84.2 & 93.6\\
    Sparse Residential & 800 & 85.5 & 95.9 & 92.6\\
    Harbor & 800 & 79.8 & 99.7 & 90.8\\
    Mobile Home Park & 800 & 79.7 & 99.7 & 90.8\\
    Crosswalk & 800 & 99.8 & 79.5 & 90.7\\
    Swimming Pool & 800 & 68.5 & 99.9 & 83.2\\
    Ferry Terminal & 800 & 100.0 & 64.1 & 81.4\\
    Storage Tank & 800 & 60.5 & 99.9 & 81.3\\
    Coastal Mansion & 800 & 98.9 & 58.2 & 81.3\\
    Wastewater Treatment Plant & 800 & 100.0 & 36.4 & 81.3\\
    \midrule
    \textbf{Macro} & --- & 95.6 & 94.3 & \textbf{96.26}\\
    \bottomrule
  \end{tabular}}
\end{table}
\begin{table}[!t]
  \centering
  \caption{RSI-CB256 -- ResNet-18 student per-class results (\%). Sorted by F1-Score descending. }
  \label{tab:r18_perclass_rsicb}
  \scriptsize
  \setlength{\tabcolsep}{3pt}
  \renewcommand{\arraystretch}{1.1}
    \resizebox{\linewidth}{!}{
  \begin{tabular}{lrrrr}
    \toprule
    \textbf{Class} & \textbf{Support} & \textbf{Precision} & \textbf{Recall} & \textbf{F1-Score}\\
    \midrule
    Woodland & 6258 & 99.8 & 99.8 & 92.0\\
    Construction Land & 3791 & 99.8 & 99.4 & 91.9\\
    Water Area & 4104 & 98.8 & 99.1 & 91.4\\
    Other Objects & 884 & 99.7 & 98.1 & 91.3\\
    Other Land & 3593 & 99.1 & 98.5 & 91.2\\
    Cultivated Land & 2817 & 98.1 & 98.8 & 90.9\\
    Transportation & 3300 & 98.2 & 98.6 & 90.8\\
    \midrule
    \textbf{Macro} & --- & 99.1 & 98.9 & \textbf{91.36}\\
    \bottomrule
  \end{tabular}}
\end{table}
\begin{table}[!t]
  \centering
  \caption{WHU-RS19 -- ResNet-18 student per-class results (\%). Sorted by F1-Score descending. }
  \label{tab:r18_perclass_whu}
  \scriptsize
  \setlength{\tabcolsep}{3pt}
  \renewcommand{\arraystretch}{1.1}
  \resizebox{\linewidth}{!}{
  \begin{tabular}{lrrrr}
    \toprule
    \textbf{Class} & \textbf{Support} & \textbf{Precision} & \textbf{Recall} & \textbf{F1-Score}\\
    \midrule
    Football Field & 50 & 100.0 & 100.0 & 87.6\\
    Meadow & 61 & 100.0 & 98.4 & 86.8\\
    River & 56 & 100.0 & 98.2 & 86.8\\
    Pond & 54 & 98.2 & 100.0 & 86.8\\
    Port & 53 & 100.0 & 98.1 & 86.7\\
    Mountain & 50 & 98.0 & 100.0 & 86.7\\
    Parking & 50 & 100.0 & 98.0 & 86.7\\
    Bridge & 52 & 100.0 & 96.2 & 85.9\\
    Desert & 50 & 96.2 & 100.0 & 85.9\\
    Farmland & 50 & 100.0 & 96.0 & 85.8\\
    Beach & 50 & 94.3 & 100.0 & 85.0\\
    Park & 50 & 100.0 & 94.0 & 84.9\\
    Forest & 53 & 96.2 & 96.2 & 84.3\\
    Railway Station & 50 & 92.5 & 98.0 & 83.3\\
    Commercial & 56 & 96.3 & 92.9 & 82.8\\
    Industrial & 53 & 96.1 & 92.5 & 82.5\\
    Residential & 54 & 88.5 & 100.0 & 82.2\\
    Viaduct & 58 & 87.9 & 100.0 & 81.9\\
    Airport & 55 & 100.0 & 81.8 & 78.8\\
    \midrule
    \textbf{Macro} & --- & 97.1 & 96.9 & \textbf{84.81}\\
    \bottomrule
  \end{tabular}}
\end{table}
\begin{table}[!t]
  \centering
  \caption{RSC11 -- ResNet-18 student per-class results (\%). Sorted by F1-Score descending. }
  \label{tab:r18_perclass_rsc11}
  \scriptsize
  \setlength{\tabcolsep}{3pt}
  \renewcommand{\arraystretch}{1.1}
    \resizebox{\linewidth}{!}{
  \begin{tabular}{lrrrr}
    \toprule
    \textbf{Class} & \textbf{Support} & \textbf{Precision} & \textbf{Recall} & \textbf{F1-Score}\\
    \midrule
    Dense Forest & 69 & 100.0 & 98.6 & 99.3\\
    Harbor & 44 & 97.7 & 97.7 & 97.7\\
    Sparse Forest & 56 & 100.0 & 89.3 & 94.3\\
    Storage Tanks & 53 & 98.0 & 90.6 & 94.1\\
    Grassland & 51 & 90.7 & 96.1 & 93.3\\
    Residential Area & 70 & 86.8 & 94.3 & 90.4\\
    High Buildings & 54 & 85.5 & 87.0 & 86.2\\
    Low Buildings & 55 & 66.2 & 96.4 & 78.5\\
    Railway & 40 & 76.9 & 75.0 & 75.9\\
    Overpass & 53 & 71.4 & 75.5 & 73.4\\
    Roads & 71 & 82.2 & 52.1 & 64.0\\
    \midrule
    \textbf{Macro} & --- & 86.9 & 86.6 & \textbf{86.10}\\
    \bottomrule
  \end{tabular}}
\end{table}
\begin{table}[!t]
  \centering
  \caption{RS2800 -- ResNet-18 student per-class results (\%). Sorted by F1-Score descending. }
  \label{tab:r18_perclass_rs2800}
  \scriptsize
  \setlength{\tabcolsep}{3pt}
  \renewcommand{\arraystretch}{1.1}
    \resizebox{\linewidth}{!}{
  \begin{tabular}{lrrrr}
    \toprule
    \textbf{Class} & \textbf{Support} & \textbf{Precision} & \textbf{Recall} & \textbf{F1-Score}\\
    \midrule
    Field & 200 & 86.8 & 85.5 & 86.0\\
    Resident & 200 & 82.4 & 82.0 & 82.2\\
    Forest & 200 & 75.7 & 85.5 & 80.3\\
    River Or Lake & 200 & 83.9 & 75.5 & 79.5\\
    Parking & 200 & 72.2 & 67.5 & 69.8\\
    Grass & 200 & 71.8 & 67.5 & 69.6\\
    Industry & 200 & 61.4 & 68.5 & 64.8\\
    \midrule
    \textbf{Macro} & --- & 76.3 & 76.0 & \textbf{76.03}\\
    \bottomrule
  \end{tabular}}
\end{table}
\begin{table}[!t]
  \centering
  \caption{SIRI-WHU -- ResNet-18 student per-class results (\%). Sorted by F1-Score descending. }
  \label{tab:r18_perclass_siriwhu}
  \scriptsize
  \setlength{\tabcolsep}{3pt}
  \renewcommand{\arraystretch}{1.1}
    \resizebox{\linewidth}{!}{
  \begin{tabular}{lrrrr}
    \toprule
    \textbf{Class} & \textbf{Support} & \textbf{Precision} & \textbf{Recall} & \textbf{F1-Score}\\
    \midrule
    Water & 100 & 99.0 & 100.0 & 99.4\\
    Residential & 100 & 91.4 & 96.0 & 93.7\\
    Harbor & 100 & 92.9 & 92.0 & 92.5\\
    Overpass & 100 & 83.8 & 98.0 & 90.3\\
    Commercial & 100 & 86.5 & 90.0 & 88.2\\
    Industrial & 100 & 83.2 & 84.0 & 83.6\\
    Idle Land & 100 & 76.5 & 88.0 & 81.9\\
    Agriculture & 100 & 89.0 & 73.0 & 80.2\\
    River & 100 & 83.5 & 76.0 & 79.6\\
    Pond & 100 & 76.1 & 83.0 & 79.4\\
    Park & 100 & 78.8 & 78.0 & 78.4\\
    Meadow & 100 & 83.1 & 64.0 & 72.3\\
    \midrule
    \textbf{Macro} & --- & 85.3 & 85.2 & \textbf{84.96}\\
    \bottomrule
  \end{tabular}}
\end{table}
\begin{table}[!t]
  \centering
  \caption{OPTIMAL31 -- ResNet-18 student per-class results (\%). Sorted by F1-Score descending. }
  \label{tab:r18_perclass_optimal}
  \scriptsize
  \setlength{\tabcolsep}{3pt}
  \renewcommand{\arraystretch}{1.1}
    \resizebox{\linewidth}{!}{
  \begin{tabular}{lrrrr}
    \toprule
    \textbf{Class} & \textbf{Support} & \textbf{Precision} & \textbf{Recall} & \textbf{F1-Score}\\
    \midrule
    Airplane & 30 & 100.0 & 100.0 & 100.0\\
    Beach & 30 & 96.7 & 96.7 & 96.7\\
    Chaparral & 30 & 96.6 & 93.3 & 94.9\\
    Island & 30 & 100.0 & 90.0 & 94.7\\
    Harbor & 30 & 93.1 & 90.0 & 91.5\\
    Parking Lot & 30 & 93.1 & 90.0 & 91.5\\
    Circular Farmland & 30 & 96.3 & 86.7 & 91.2\\
    Runway & 30 & 96.3 & 86.7 & 91.2\\
    Airport & 30 & 87.5 & 93.3 & 90.3\\
    Roundabout & 30 & 92.9 & 86.7 & 89.7\\
    Forest & 30 & 81.1 & 100.0 & 89.6\\
    Rectangular Farmland & 30 & 82.9 & 96.7 & 89.2\\
    Mountain & 30 & 84.8 & 93.3 & 88.9\\
    Dense Residential & 30 & 87.1 & 90.0 & 88.5\\
    Basketball Court & 30 & 96.0 & 80.0 & 87.3\\
    Mobile Home Park & 30 & 84.4 & 90.0 & 87.1\\
    Desert & 30 & 81.8 & 90.0 & 85.7\\
    Golf Course & 30 & 83.3 & 83.3 & 83.3\\
    Bridge & 30 & 85.7 & 80.0 & 82.8\\
    Lake & 30 & 85.7 & 80.0 & 82.8\\
    Baseball Diamond & 30 & 74.3 & 86.7 & 80.0\\
    Railway & 30 & 88.0 & 73.3 & 80.0\\
    Overpass & 30 & 82.1 & 76.7 & 79.3\\
    Ground Track Field & 30 & 84.6 & 73.3 & 78.6\\
    Meadow & 30 & 84.6 & 73.3 & 78.6\\
    Medium Residential & 30 & 69.4 & 83.3 & 75.8\\
    Industrial Area & 30 & 75.9 & 73.3 & 74.6\\
    Freeway & 30 & 70.4 & 63.3 & 66.7\\
    Church & 30 & 69.2 & 60.0 & 64.3\\
    Commercial Area & 30 & 56.8 & 70.0 & 62.7\\
    Intersection & 30 & 54.3 & 63.3 & 58.5\\
    \midrule
    \textbf{Macro} & --- & 84.3 & 83.7 & \textbf{83.74}\\
    \bottomrule
  \end{tabular}}
\end{table}

\subsection{BatchGCN Refinement}
\label{supp:gcn}
 
The BatchGCN module exploits relational structure within each mini-batch.
For a mini-batch of $B$ images, we extract MoE features and build a
$k$-nearest-neighbor ($k$-NN) graph in feature space.
Each image corresponds to one node; an edge connects $x_i$ to $x_j$ if
$x_j$ is among the $k\!=\!8$ nearest neighbors of $x_i$ by cosine
similarity:
\begin{equation}
  A_{ij} = \mathbf{1}\!\left[x_j \in \mathcal{N}_k(x_i)\right].
  \label{eq:adj}
\end{equation}
Constructing this graph per mini-batch rather than over the entire dataset
keeps computation tractable: for batch sizes up to 256 the $k$-NN
construction requires negligible time relative to the forward pass.
 
A single graph convolutional layer with symmetric degree normalization
propagates information between neighboring nodes:
\begin{equation}
  Z' = \ReLU\!\bigl(\hat{A}\,Z\,W\bigr),
  \quad
  \hat{A} = \tilde{D}^{-\frac{1}{2}}\tilde{A}\tilde{D}^{-\frac{1}{2}},
  \label{eq:gcn}
\end{equation}
where $\tilde{A}\!=\!A\!+\!I$ adds self-loops so each node retains its
own features, $\tilde{D}$ is the corresponding degree matrix, and $W$ is
a learnable weight matrix.
The normalization by $\tilde{D}$ prevents high-degree nodes from
dominating the aggregation.
A single layer is used deliberately to avoid over-smoothing, which would
collapse all node representations toward the graph mean and destroy
class-discriminative structure.
 \begin{figure*}[t]
  \centering
  \begin{subfigure}[b]{0.48\linewidth}
    \includegraphics[width=\linewidth]{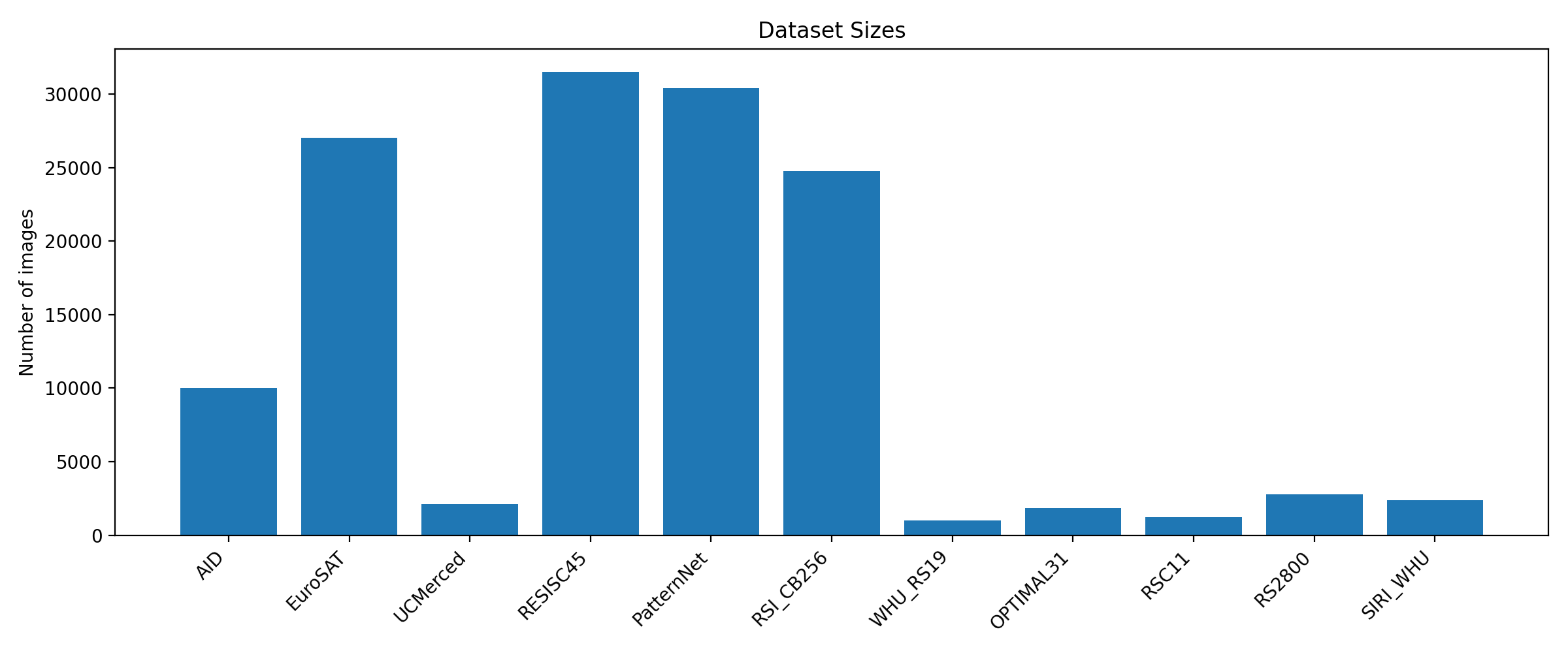}
    \caption{Total images per dataset.}
  \end{subfigure}
  \hfill
  \begin{subfigure}[b]{0.48\linewidth}
    \includegraphics[width=\linewidth]{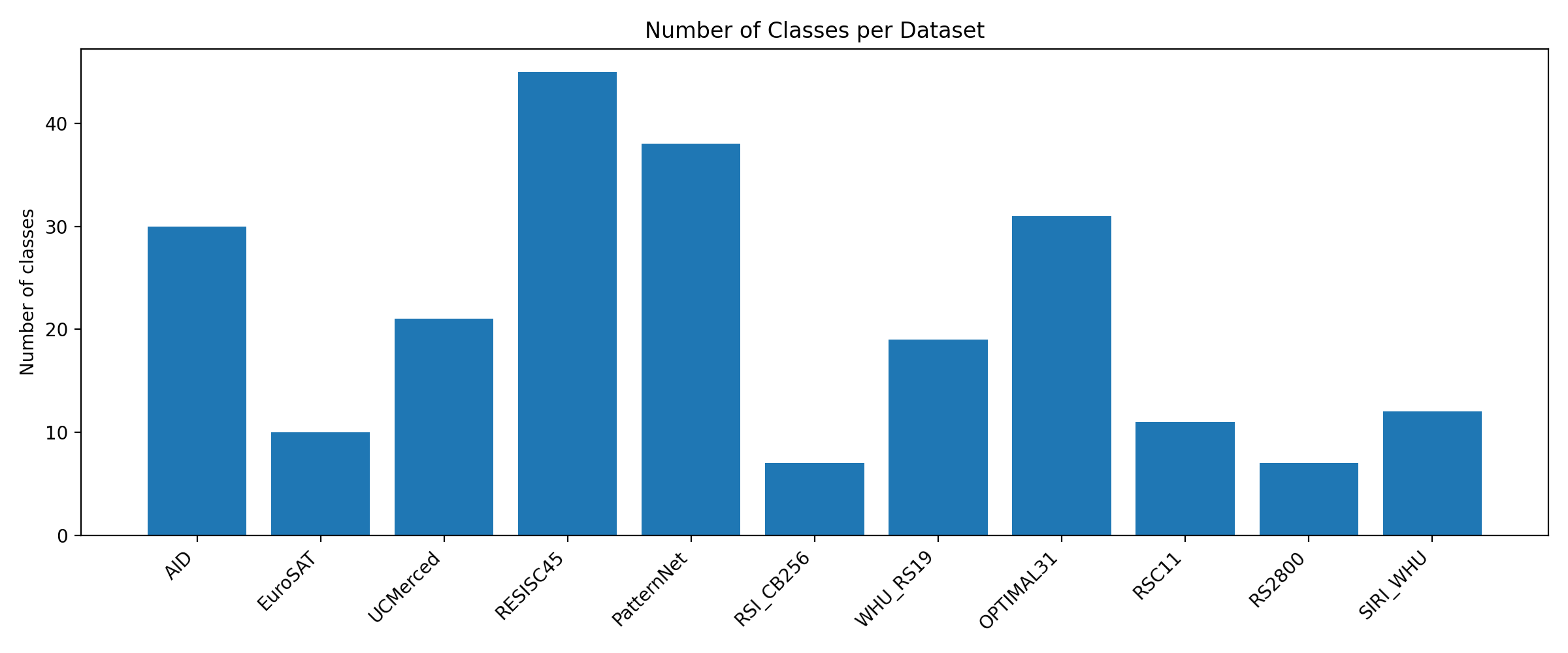}
    \caption{Number of classes per dataset.}
  \end{subfigure}
  \caption{Dataset overview. The $30\times$ size imbalance (1\,005 to
  31\,500 images) motivates dataset-balanced sampling and stratified
  active queries.}
  \label{fig:dataset_overview}
\end{figure*}
 \begin{figure}[!t]
  \centering
  \includegraphics[width=1\linewidth]{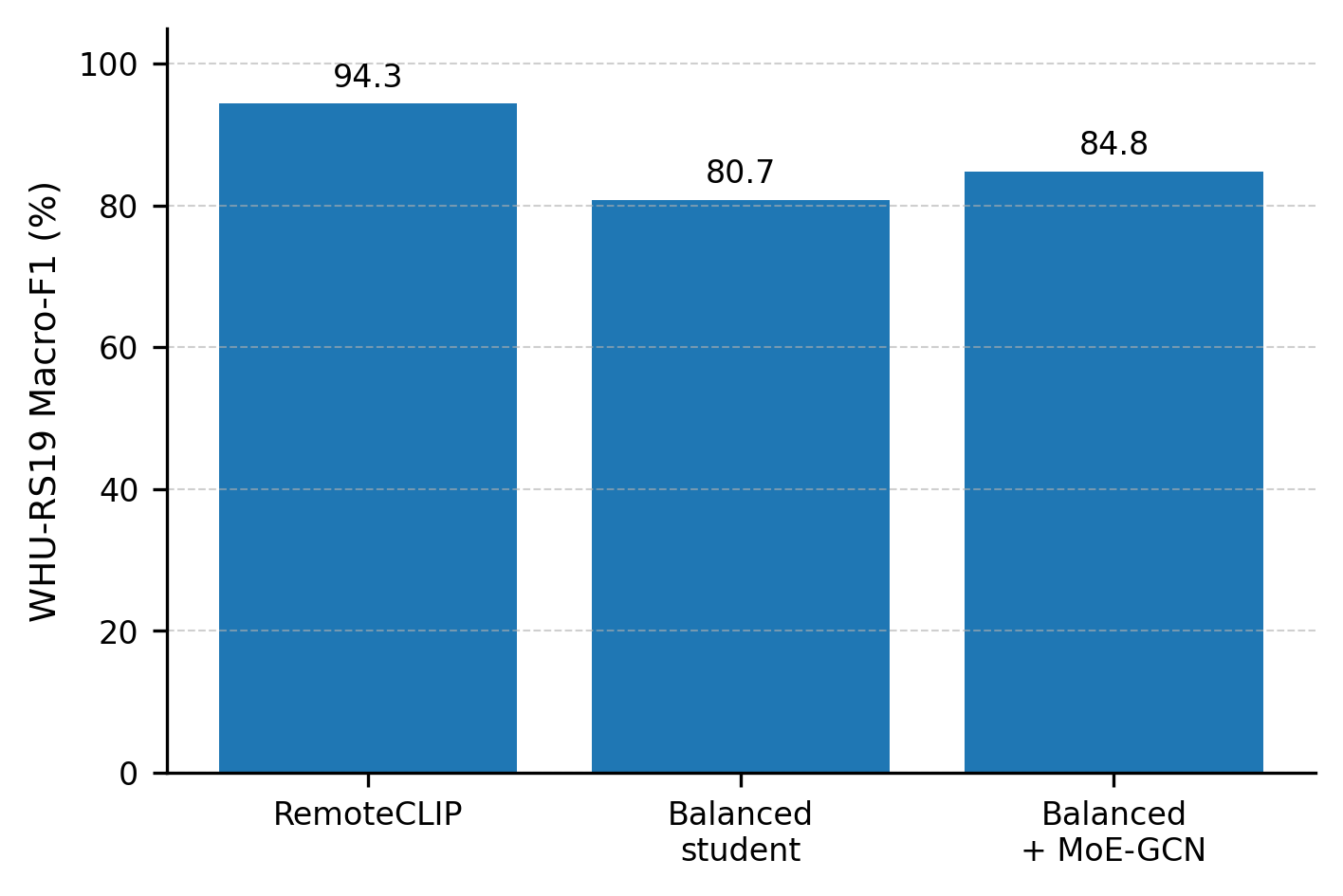}
  \caption{WHU-RS19 Macro-F1 comparison.
  RemoteCLIP remains strongest on this small dataset, while the balanced
  student substantially narrows the gap and MoE+GCN further improves the
  best checkpoint.}
  \label{fig:supp_whu_balanced_comparison}
\end{figure}
  
\begin{figure*}[!t]
  \centering
  \begin{subfigure}[b]{0.32\linewidth}
    \includegraphics[width=\linewidth]{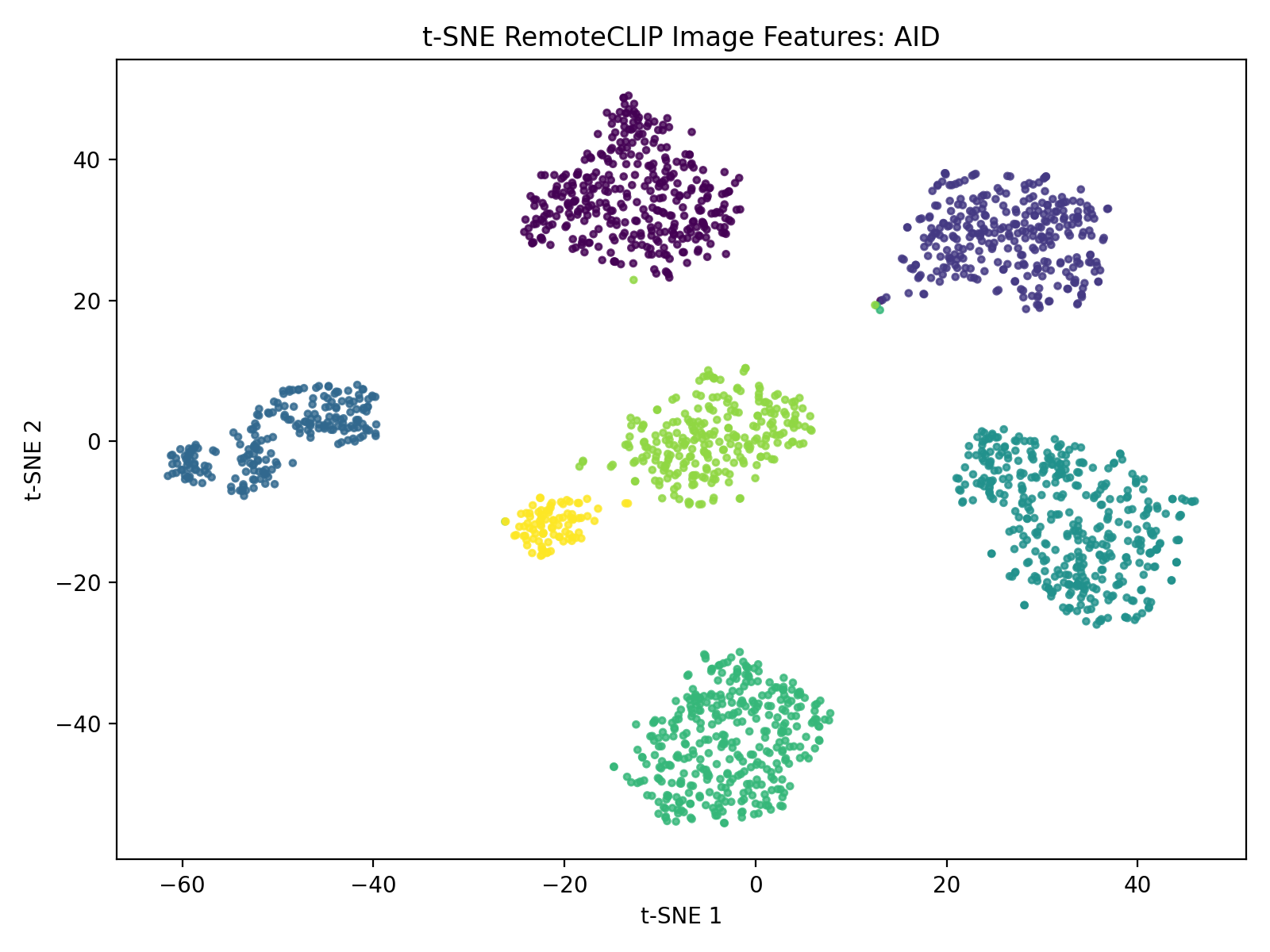}
    \caption{AID (RemoteCLIP)}
  \end{subfigure}
  \begin{subfigure}[b]{0.32\linewidth}
    \includegraphics[width=\linewidth]{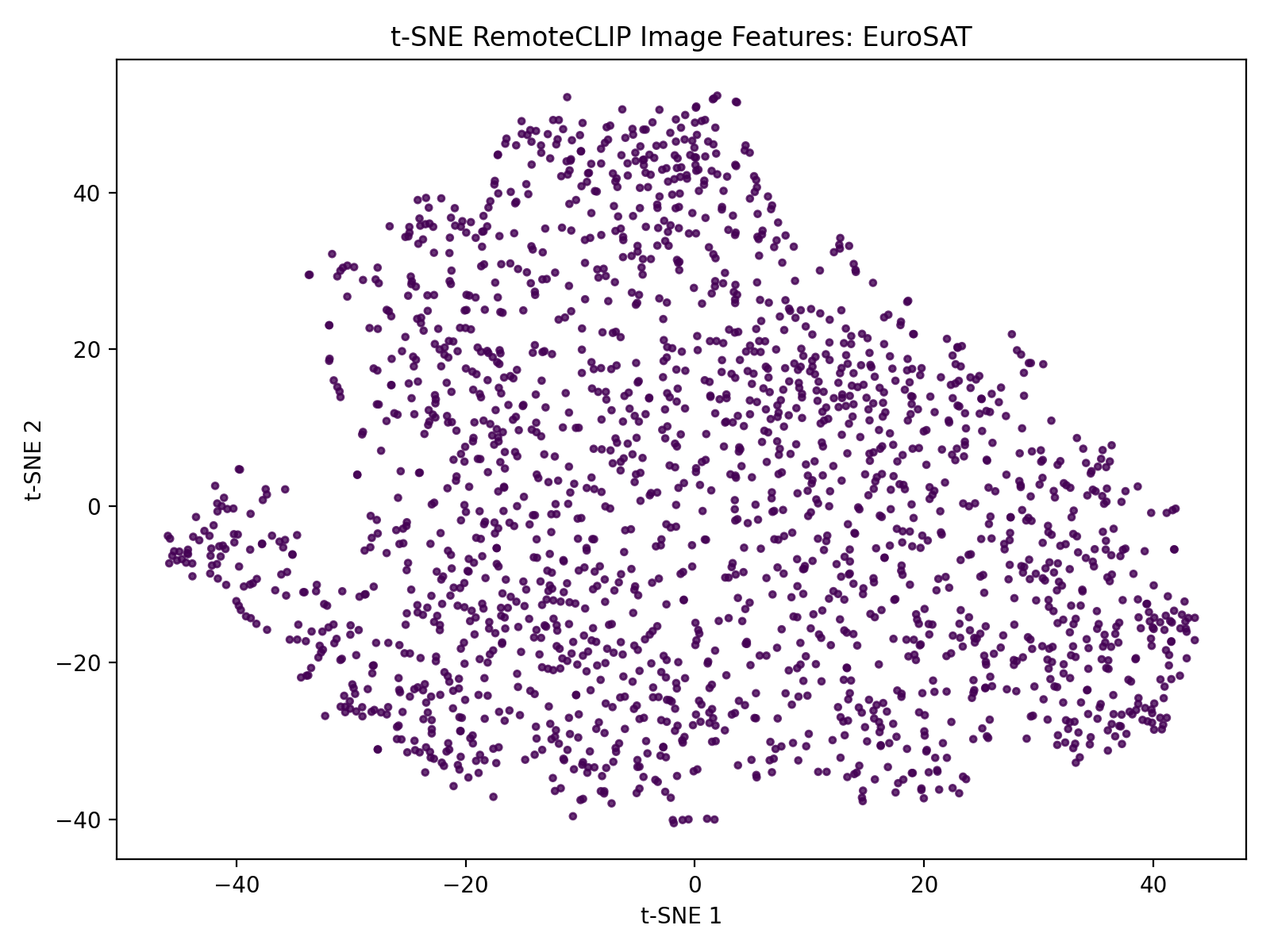}
    \caption{EuroSAT (RemoteCLIP)}
  \end{subfigure}
  \begin{subfigure}[b]{0.32\linewidth}
    \includegraphics[width=\linewidth]{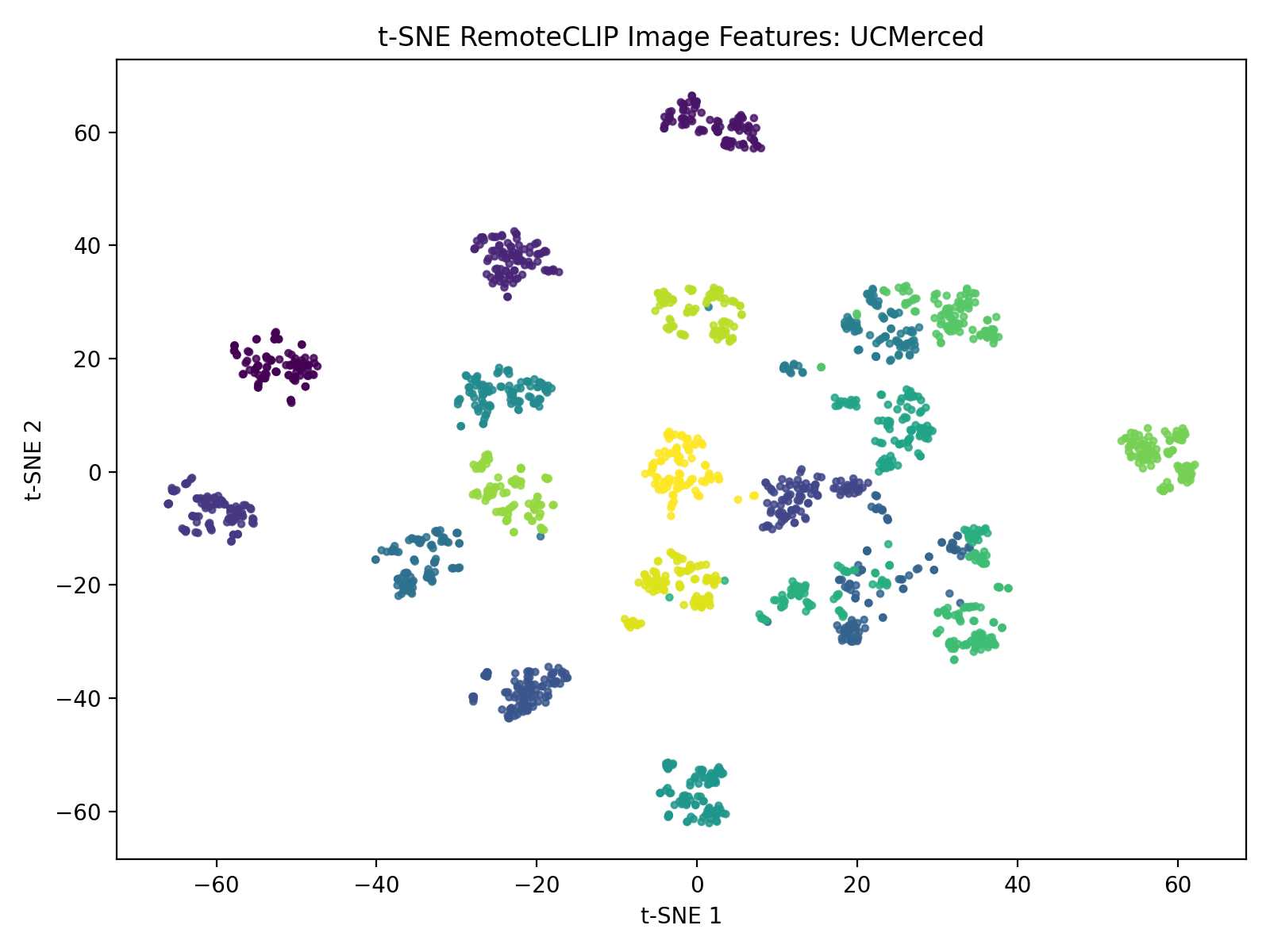}
    \caption{UC~Merced (RemoteCLIP)}
  \end{subfigure}
  \caption{t-SNE projections of RemoteCLIP features on core datasets.
  Classes are generally well-separated, confirming the quality of the
  teacher's zero-shot representations.}
  \label{fig:tsne_rc}
\end{figure*}
\begin{figure*}[!t]
  \centering
  \begin{subfigure}[b]{0.32\linewidth}
    \includegraphics[width=\linewidth]{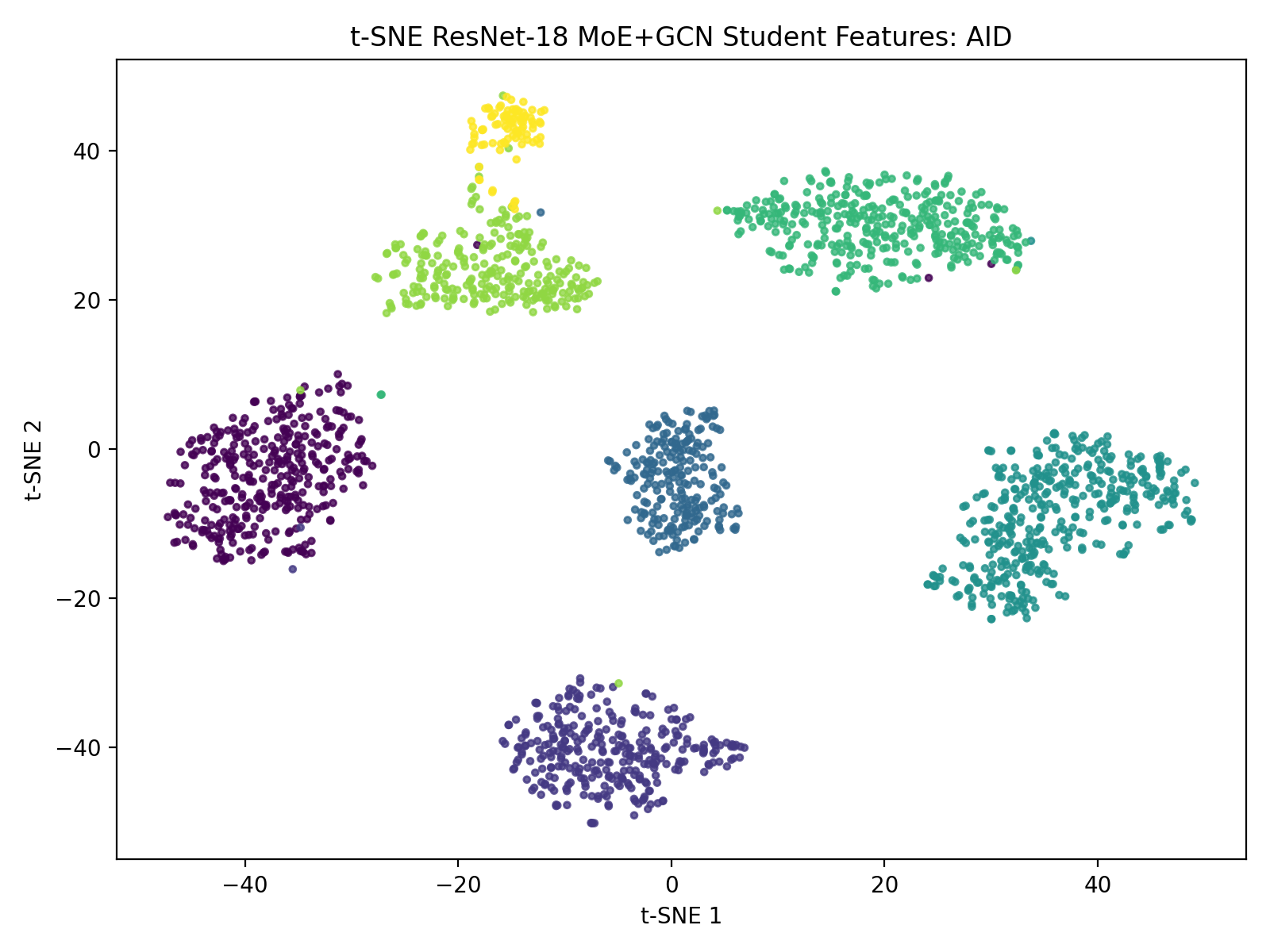}
    \caption{AID (Student)}
  \end{subfigure}
  \begin{subfigure}[b]{0.32\linewidth}
    \includegraphics[width=\linewidth]{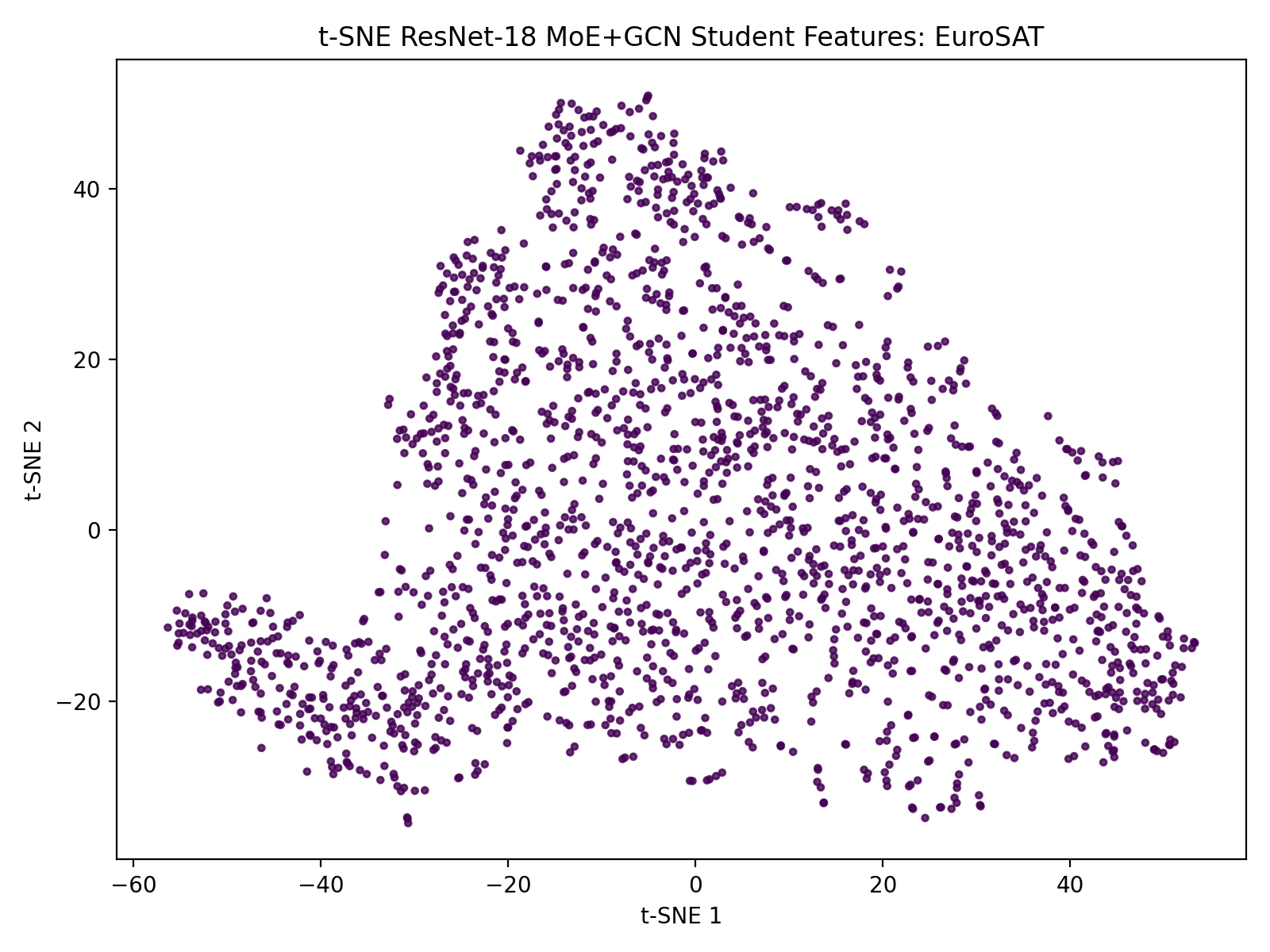}
    \caption{EuroSAT (Student)}
  \end{subfigure}
  \begin{subfigure}[b]{0.32\linewidth}
    \includegraphics[width=\linewidth]{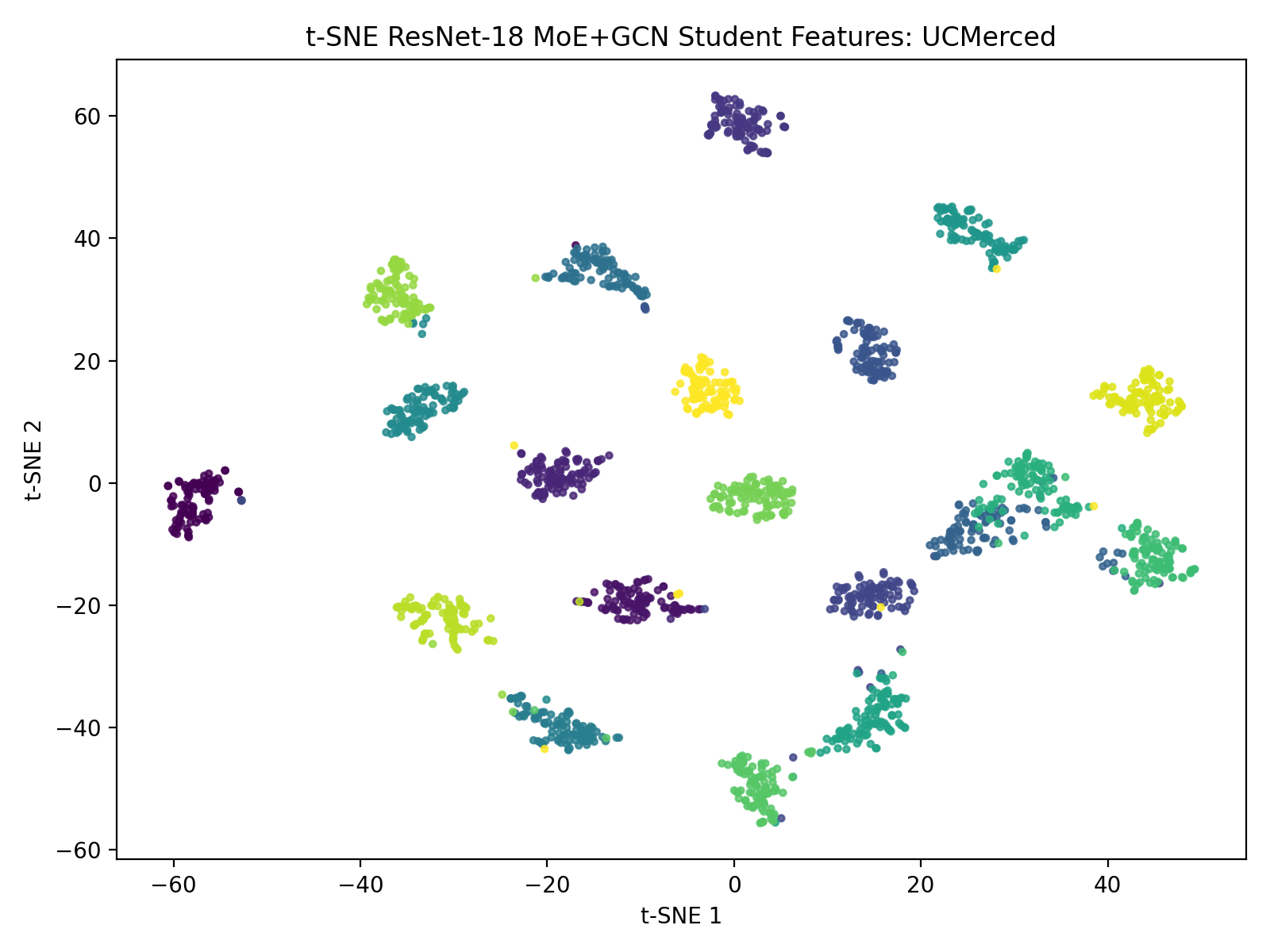}
    \caption{UC~Merced (Student)}
  \end{subfigure}
  \caption{t-SNE projections of student features on core datasets.
  The student achieves comparable cluster separation to RemoteCLIP,
  confirming successful knowledge distillation.}
  \label{fig:tsne_student}
\end{figure*}
 \begin{figure}[!t]
  \centering
  \begin{subfigure}[b]{0.32\linewidth}
    \includegraphics[width=\linewidth]{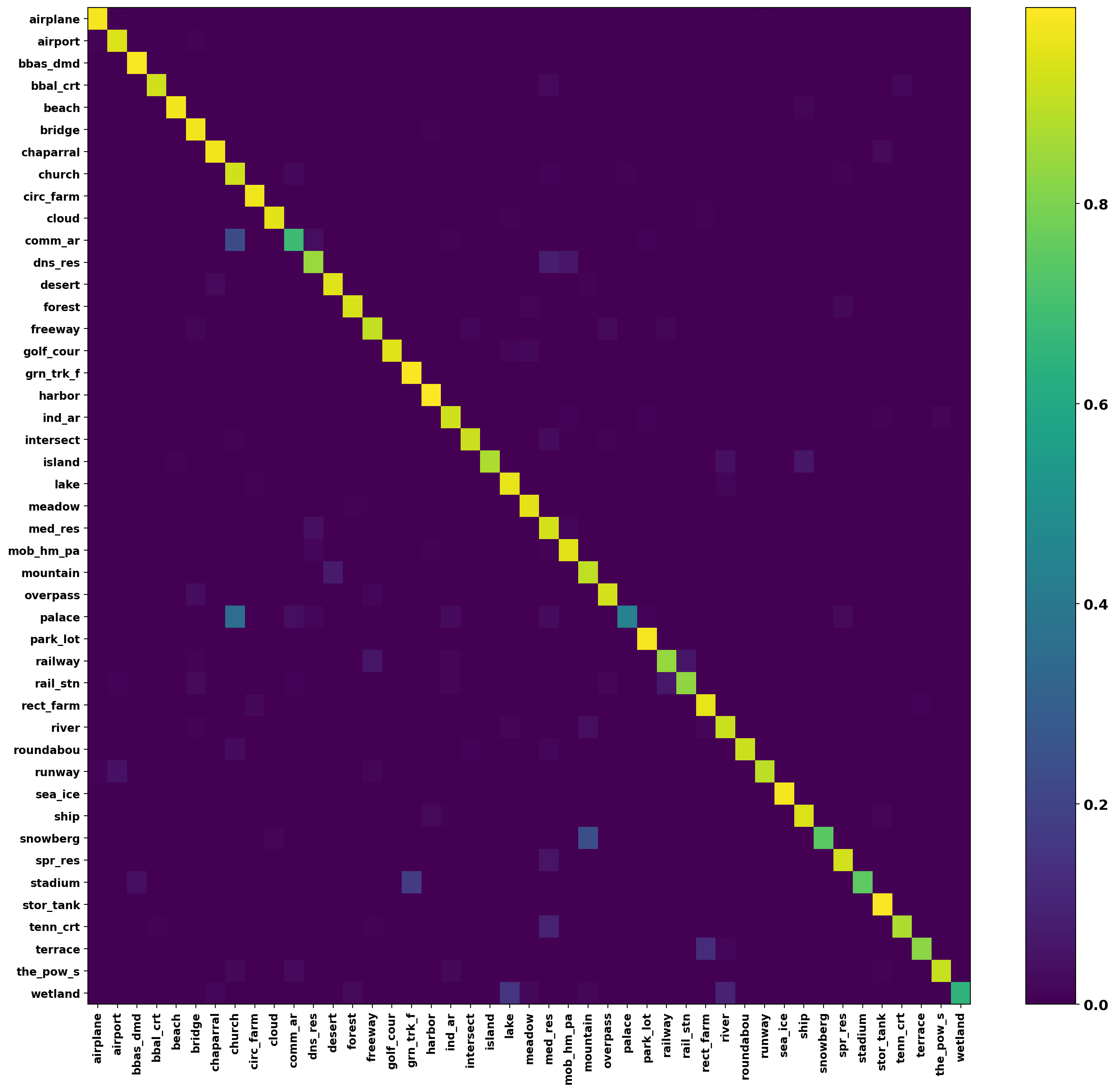}
    \caption{RESISC45}
  \end{subfigure}
  \begin{subfigure}[b]{0.32\linewidth}
    \includegraphics[width=\linewidth]{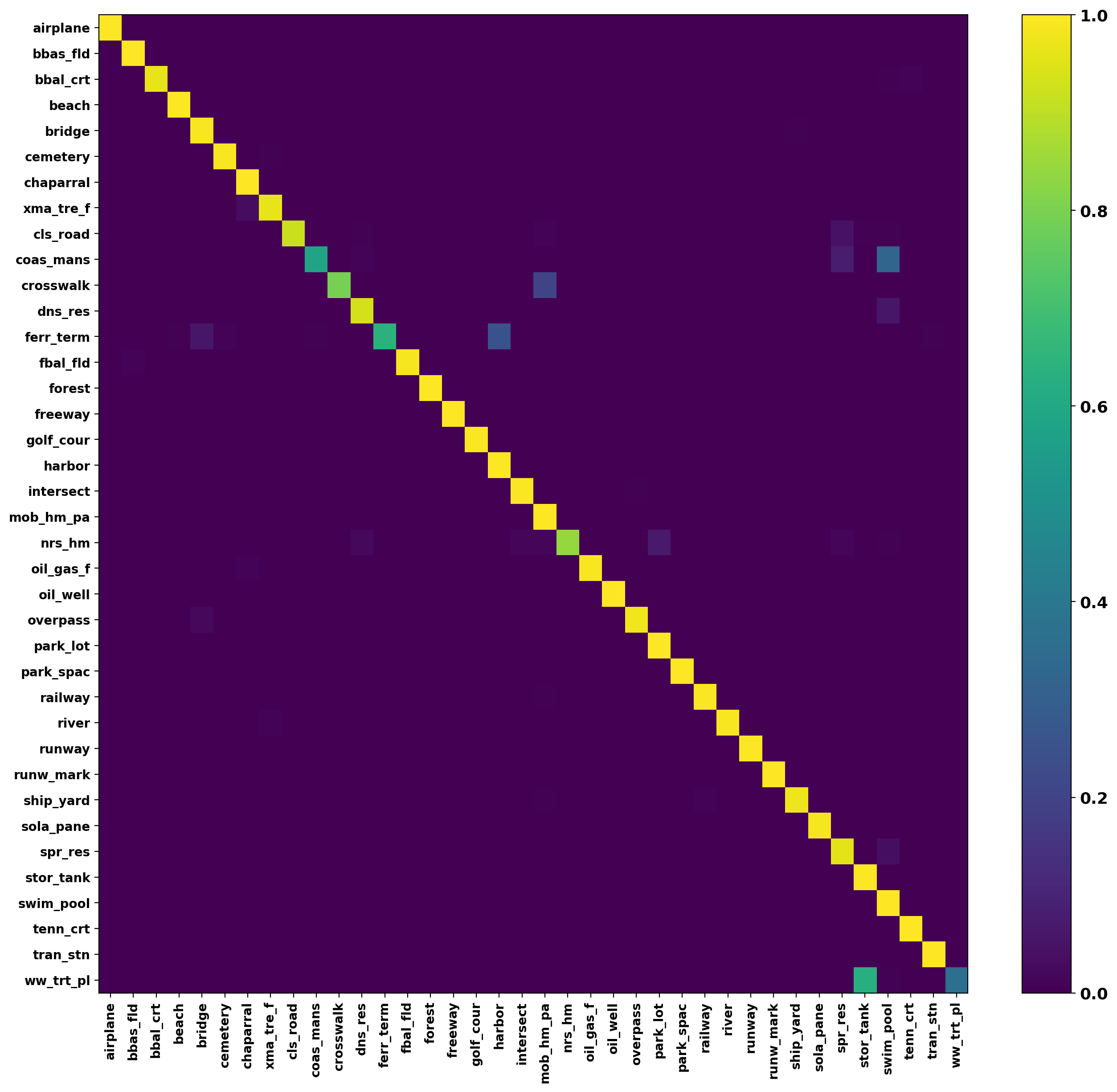}
    \caption{PatternNet}
  \end{subfigure}
  \begin{subfigure}[b]{0.32\linewidth}
    \includegraphics[width=\linewidth]{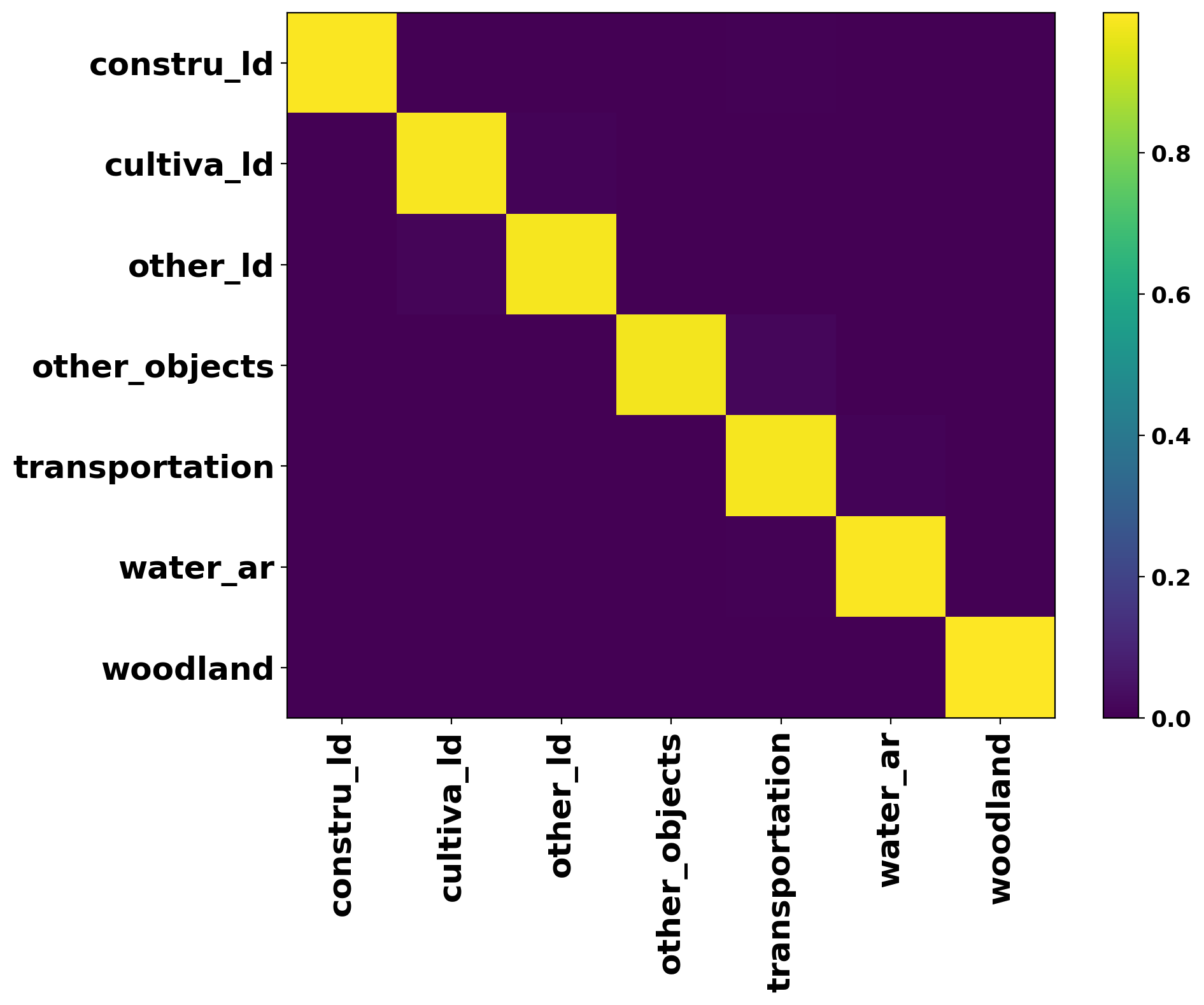}
    \caption{RSI-CB256}
  \end{subfigure}
  \caption{Normalized confusion matrices -- extended datasets.}
  \label{fig:confusion_extended}
\end{figure}
\begin{figure}[!t]
  \centering
  \begin{subfigure}[b]{0.24\linewidth}
    \includegraphics[width=\linewidth]{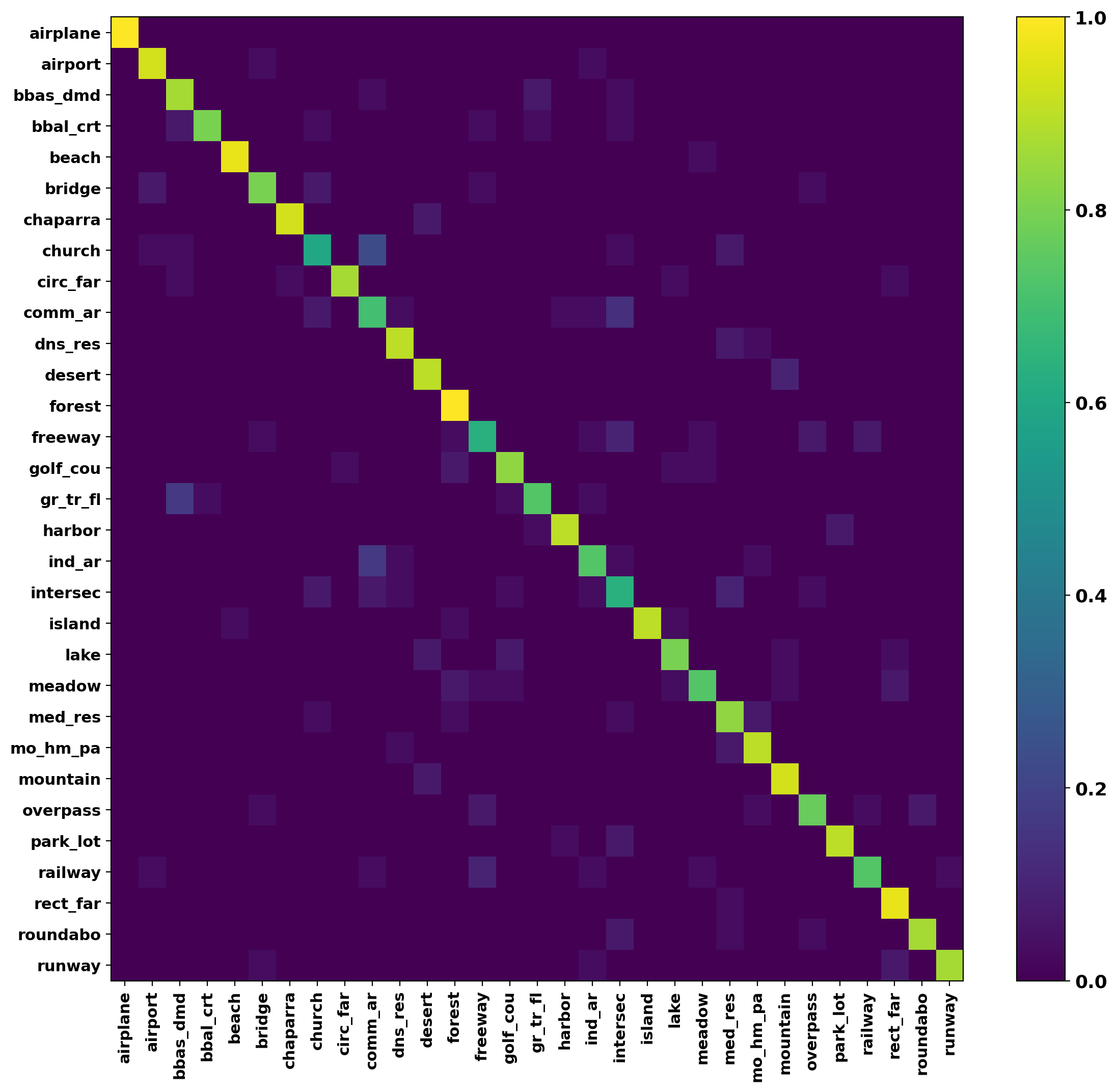}
    \caption{OPTIMAL31}
  \end{subfigure}
  \begin{subfigure}[b]{0.24\linewidth}
    \includegraphics[width=\linewidth]{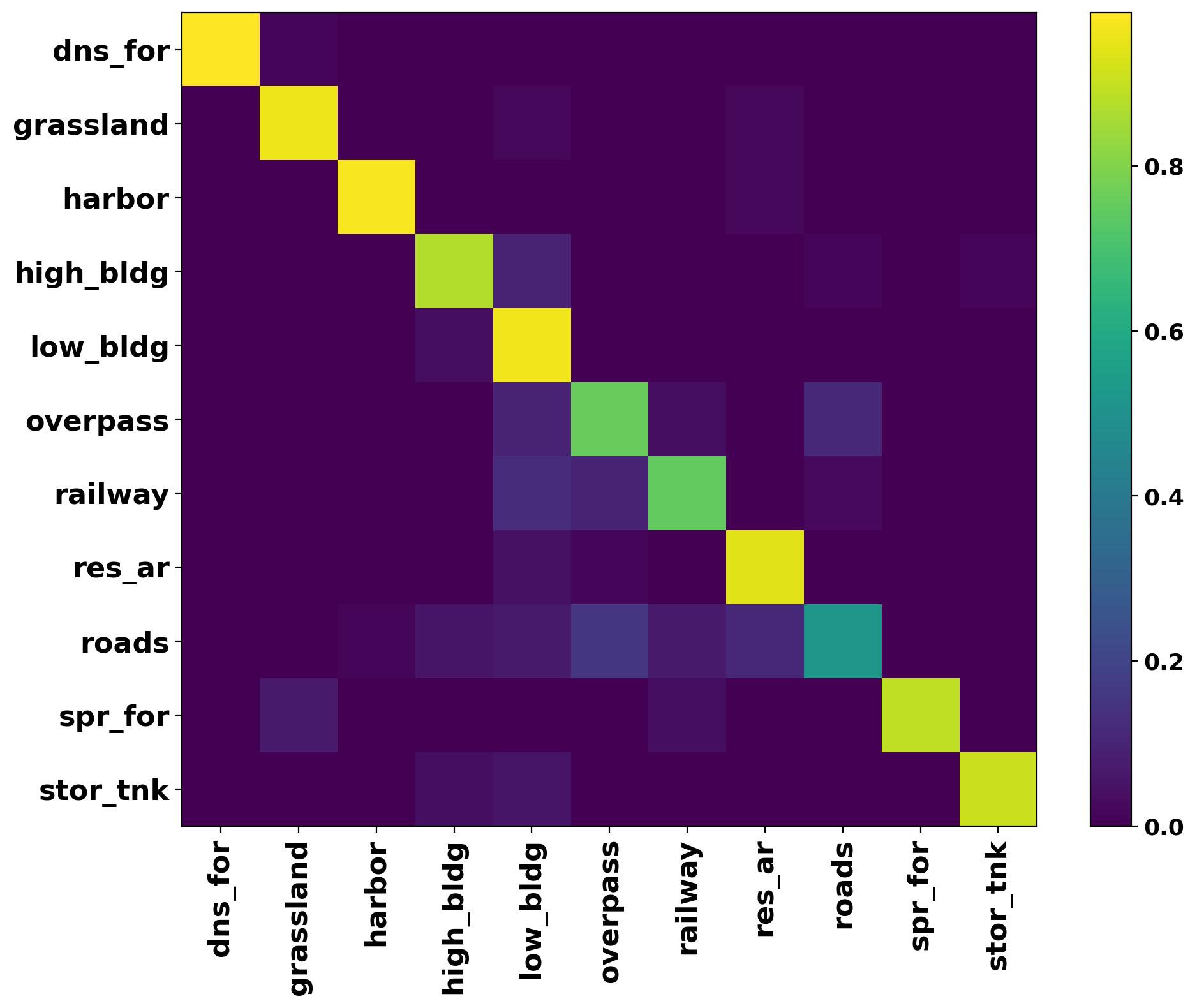}
    \caption{RSC11}
  \end{subfigure}
  \begin{subfigure}[b]{0.24\linewidth}
    \includegraphics[width=\linewidth]{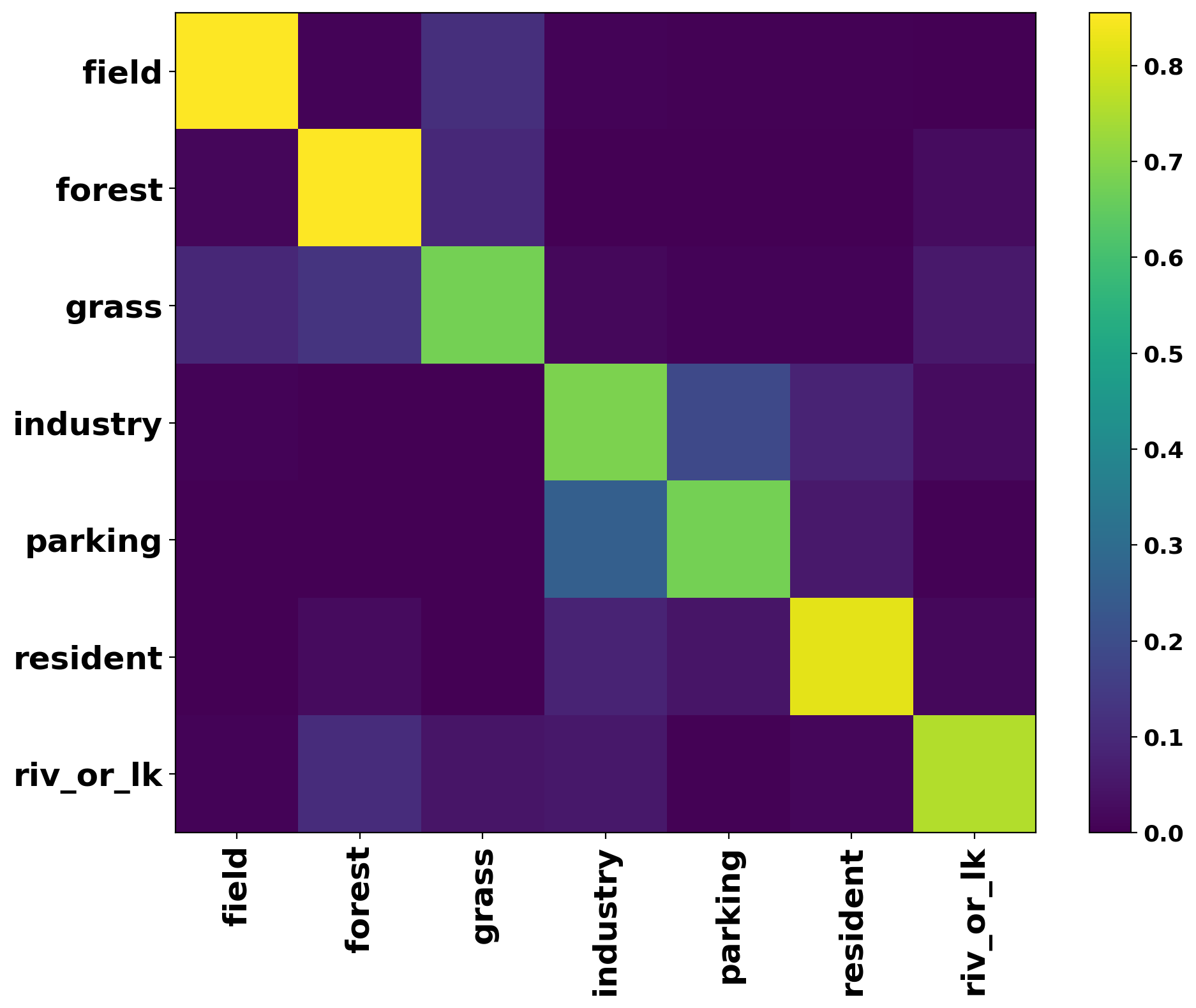}
    \caption{RS2800}
  \end{subfigure}
  \begin{subfigure}[b]{0.24\linewidth}
    \includegraphics[width=\linewidth]{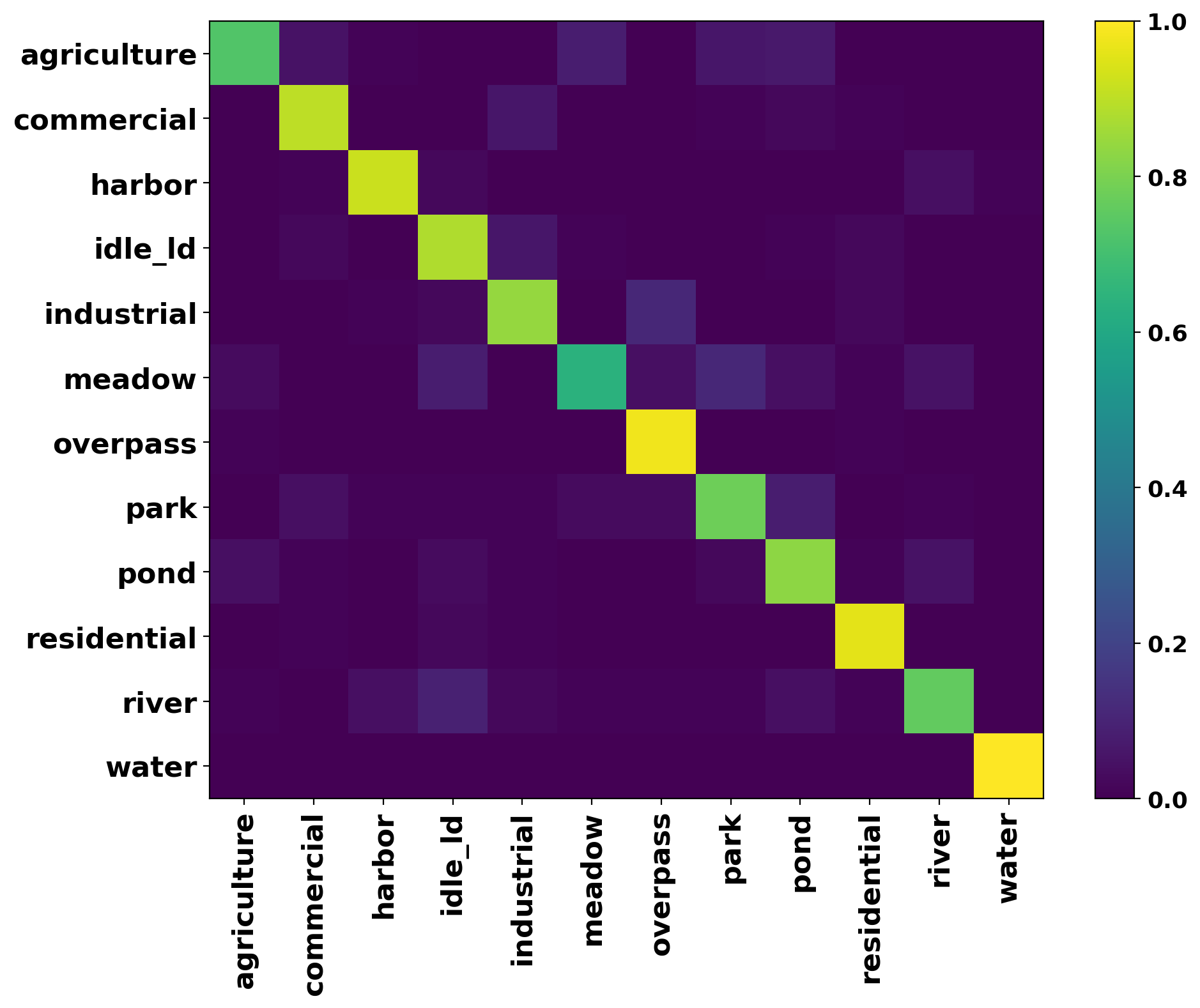}
    \caption{SIRI-WHU}
  \end{subfigure}
  \caption{Normalized confusion matrices -- unseen datasets (prototype mode).}
  \label{fig:confusion_unseen}
\end{figure}
\begin{figure*}[t]
  \centering
  \begin{subfigure}[b]{0.32\linewidth}
    \includegraphics[width=\linewidth]{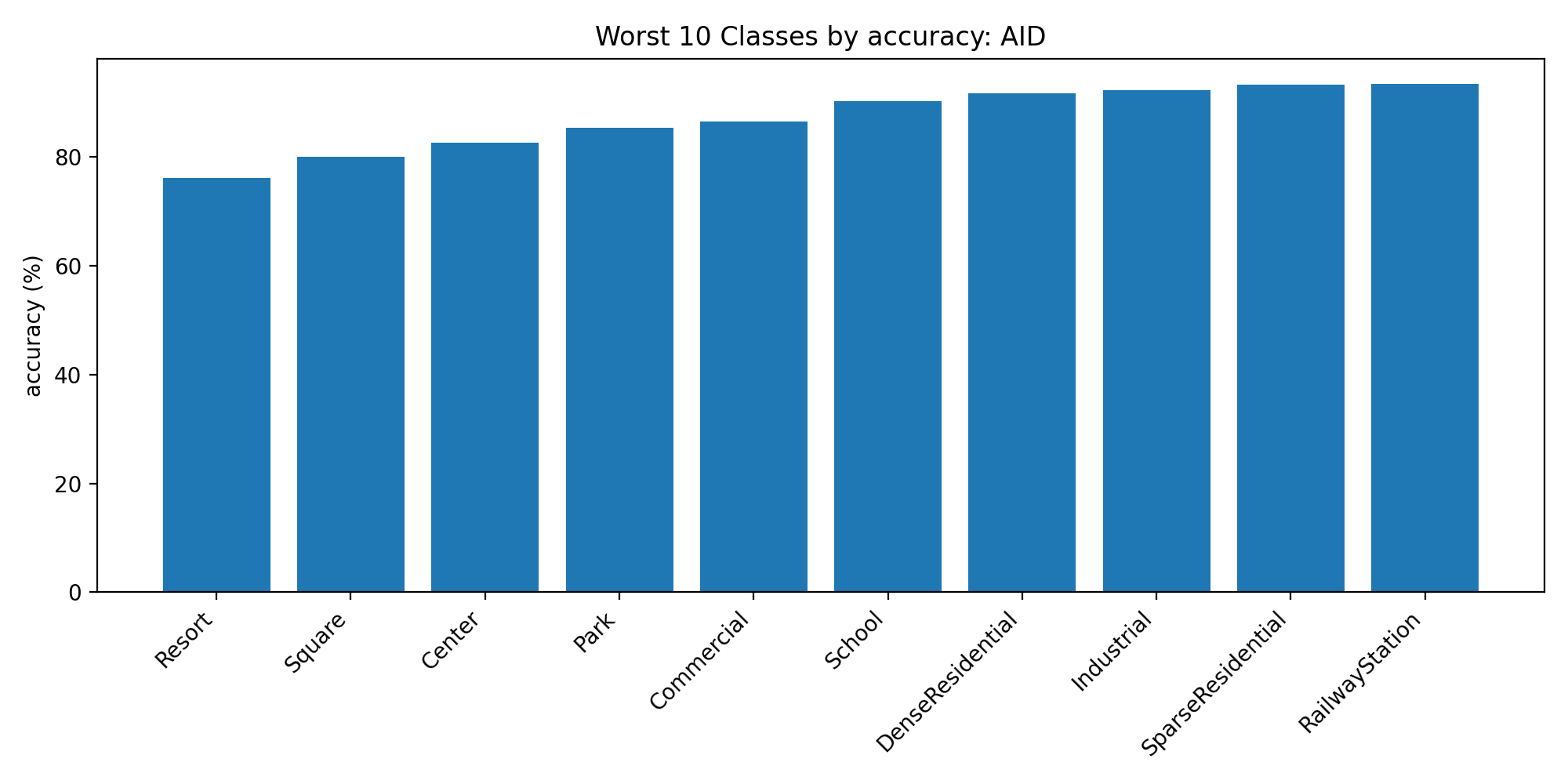}
    \caption{AID}
  \end{subfigure}
  \begin{subfigure}[b]{0.32\linewidth}
    \includegraphics[width=\linewidth]{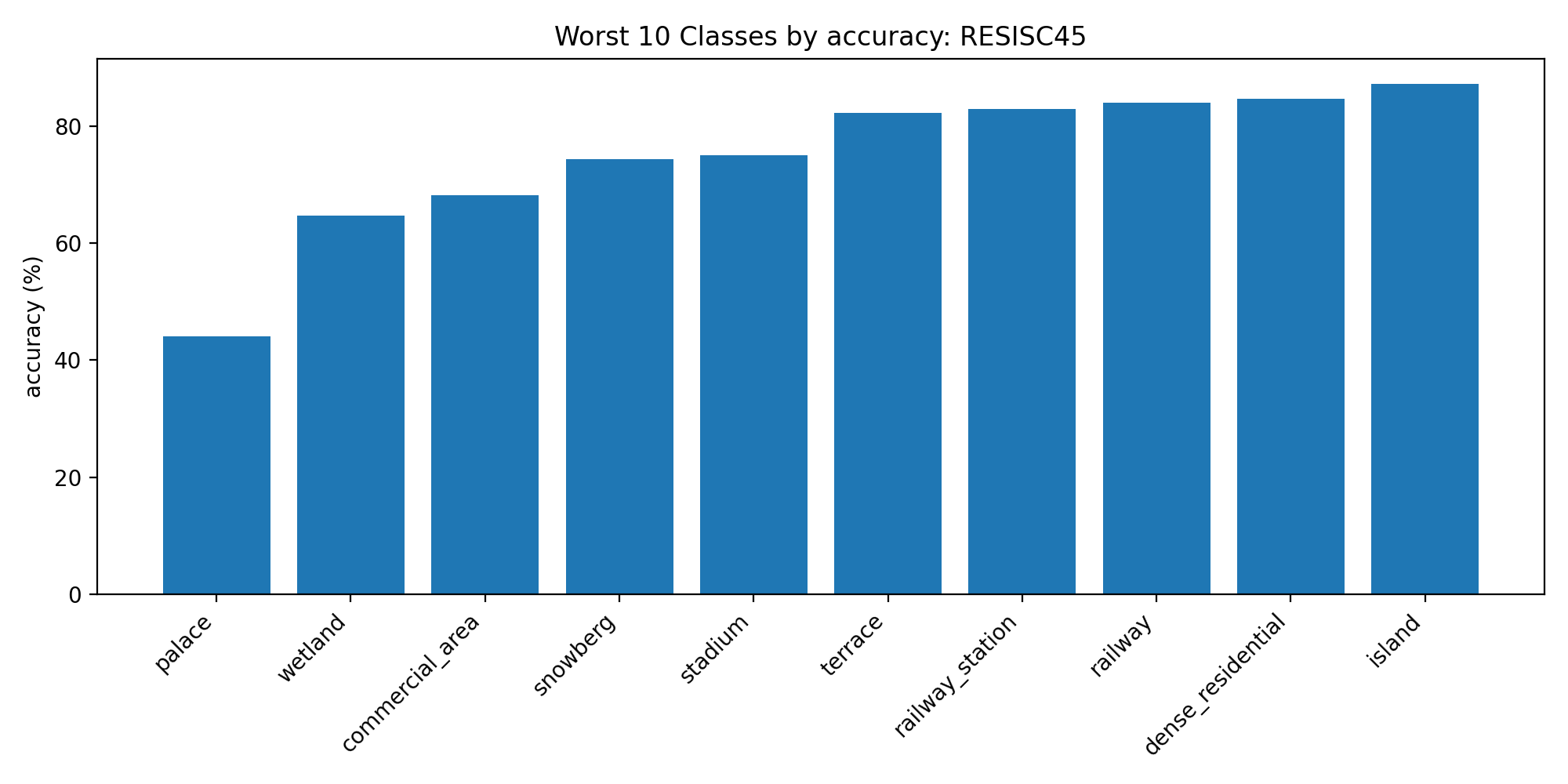}
    \caption{RESISC45}
  \end{subfigure}
  \begin{subfigure}[b]{0.32\linewidth}
    \includegraphics[width=\linewidth]{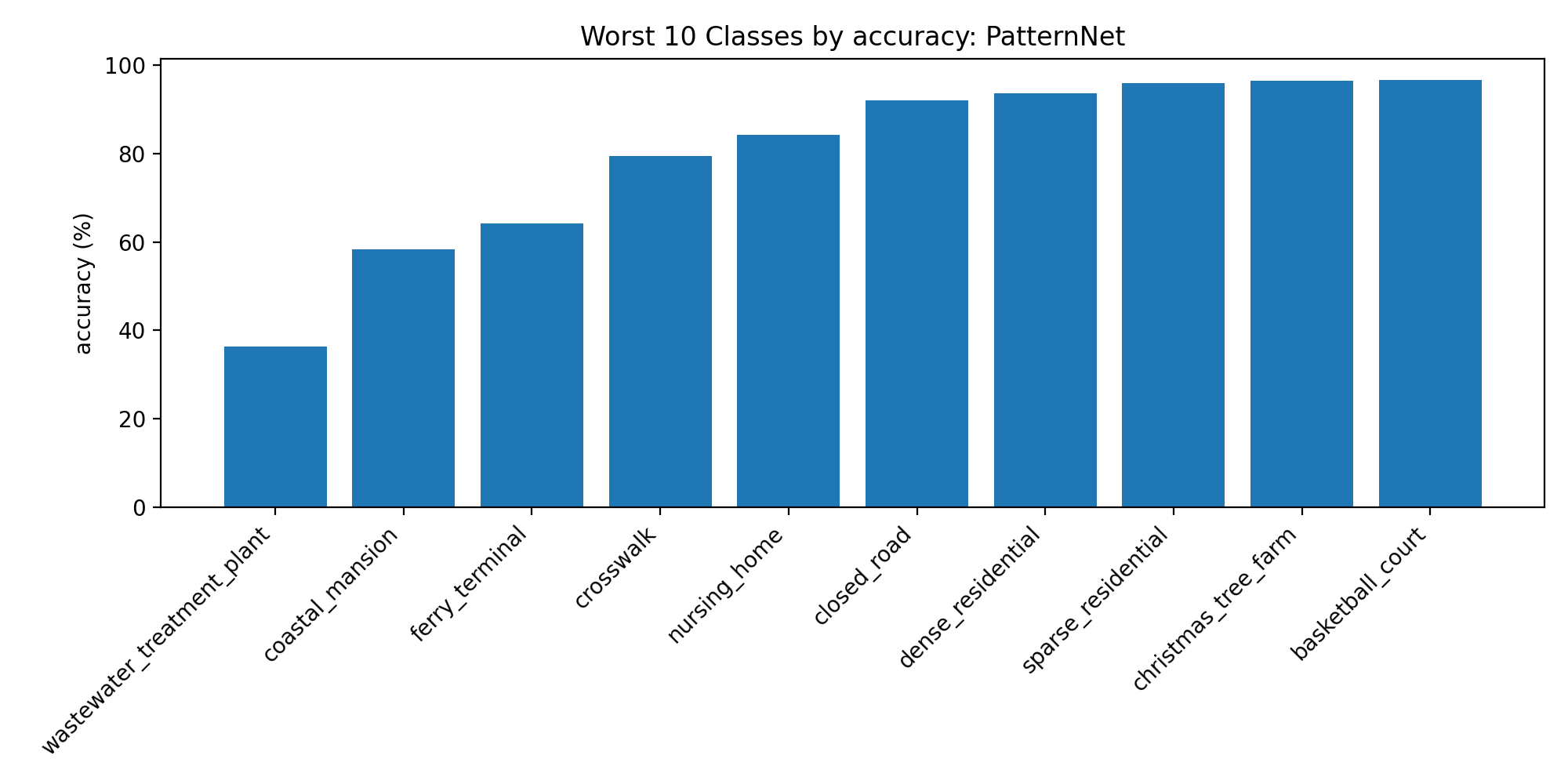}
    \caption{PatternNet}
  \end{subfigure}
  \\[6pt]
  \begin{subfigure}[b]{0.32\linewidth}
    \includegraphics[width=\linewidth]{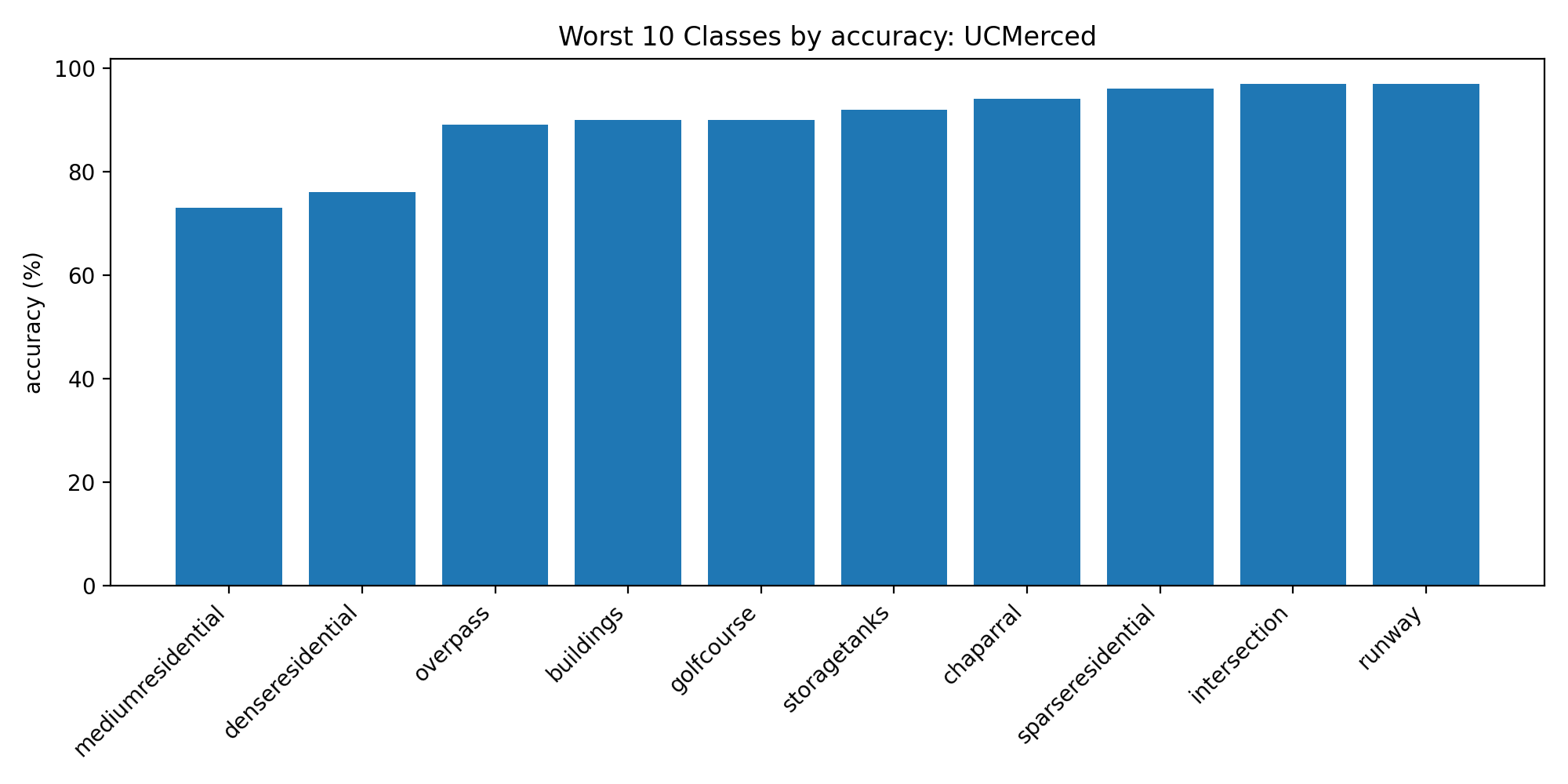}
    \caption{UC~Merced}
  \end{subfigure}
  \begin{subfigure}[b]{0.32\linewidth}
    \includegraphics[width=\linewidth]{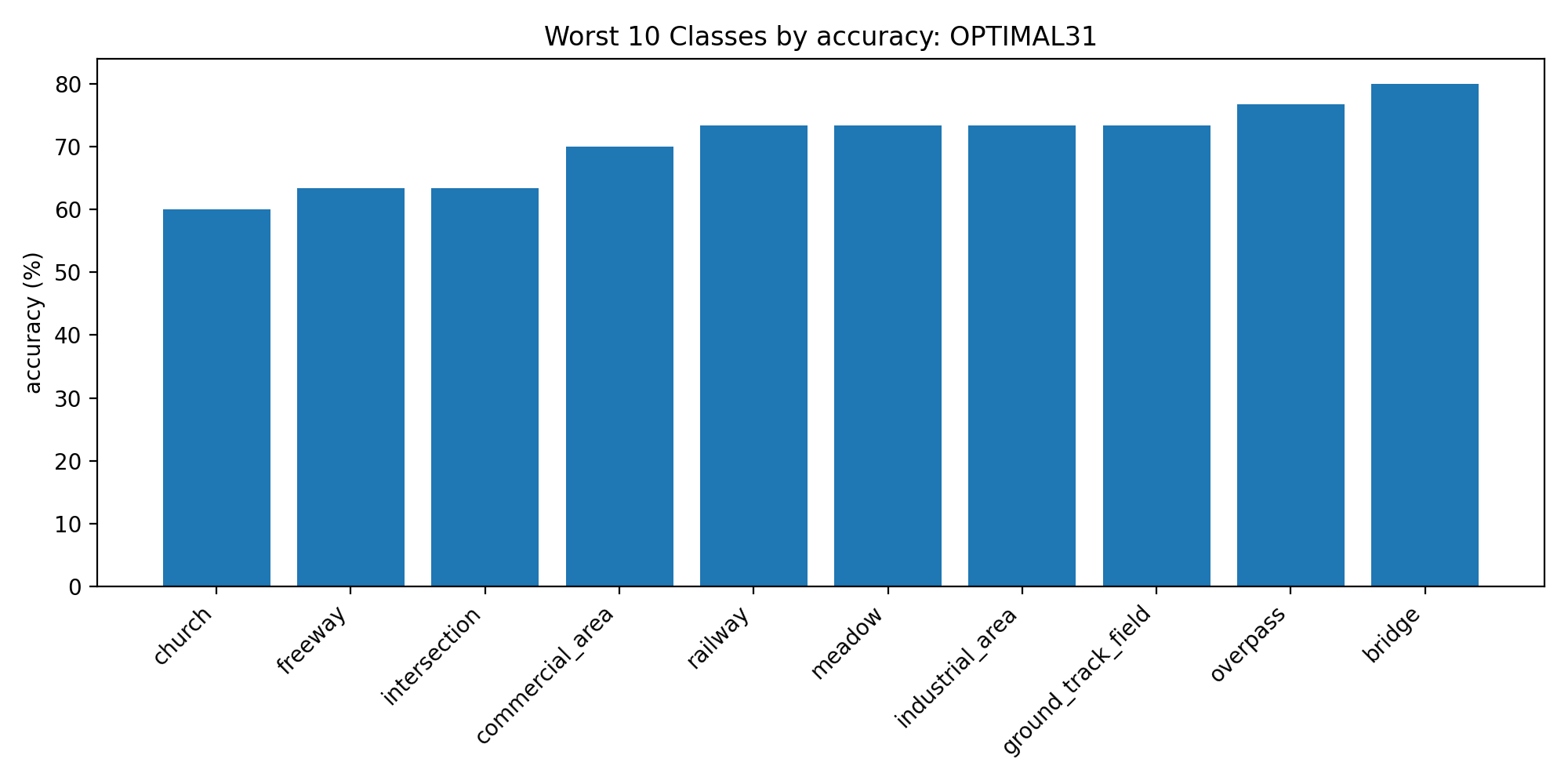}
    \caption{OPTIMAL31}
  \end{subfigure}
  \begin{subfigure}[b]{0.32\linewidth}
    \includegraphics[width=\linewidth]{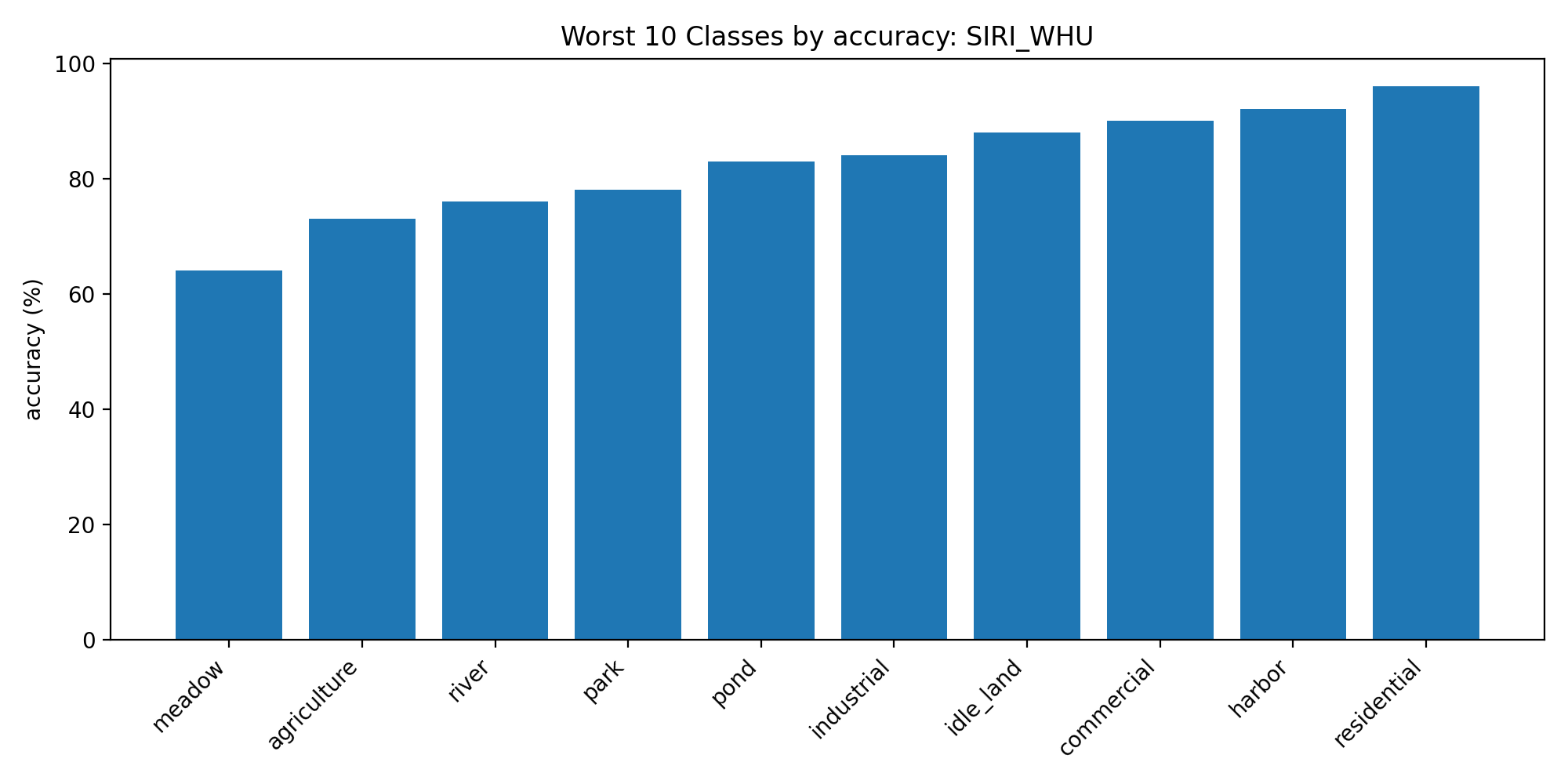}
    \caption{SIRI-WHU}
  \end{subfigure}
  \caption{Lowest-accuracy classes per dataset. Hard categories typically
  share visual patterns with neighboring classes, limiting discriminability
  under limited labeled supervision.}
  \label{fig:worst_classes}
\end{figure*}
 \begin{figure*}[!t]
  \centering
  \begin{subfigure}[b]{0.48\linewidth}
    \includegraphics[width=\linewidth]{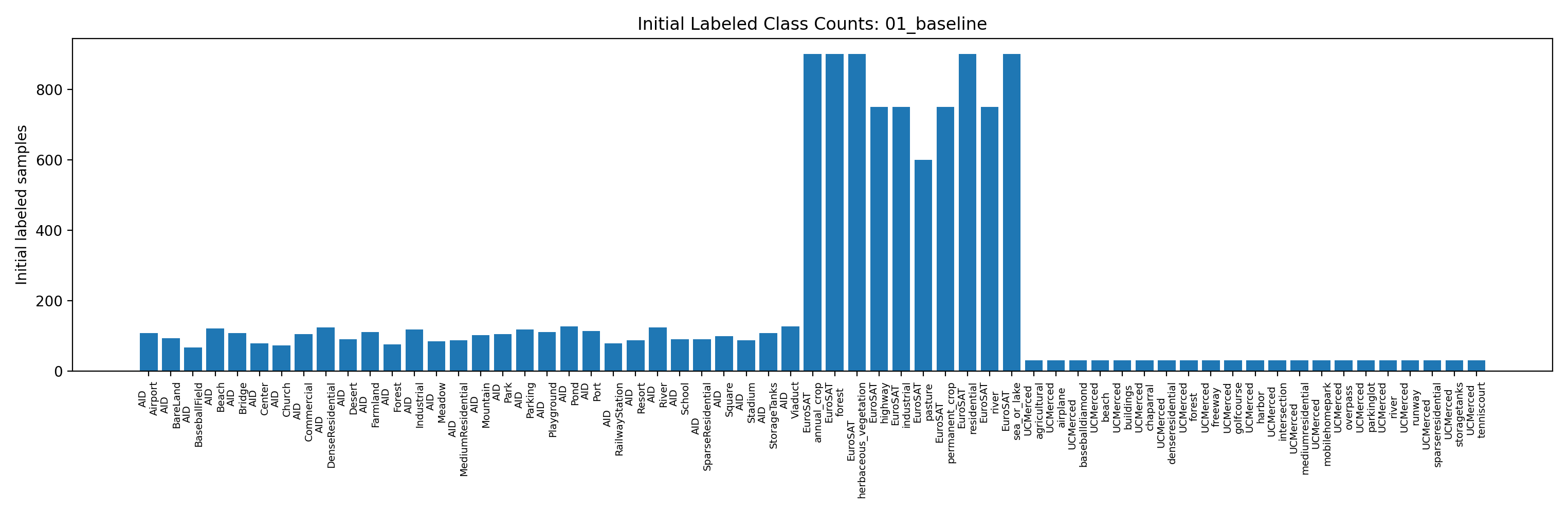}
    \caption{Config 01 (Baseline).}
  \end{subfigure}
  \hfill
  \begin{subfigure}[b]{0.48\linewidth}
    \includegraphics[width=\linewidth]{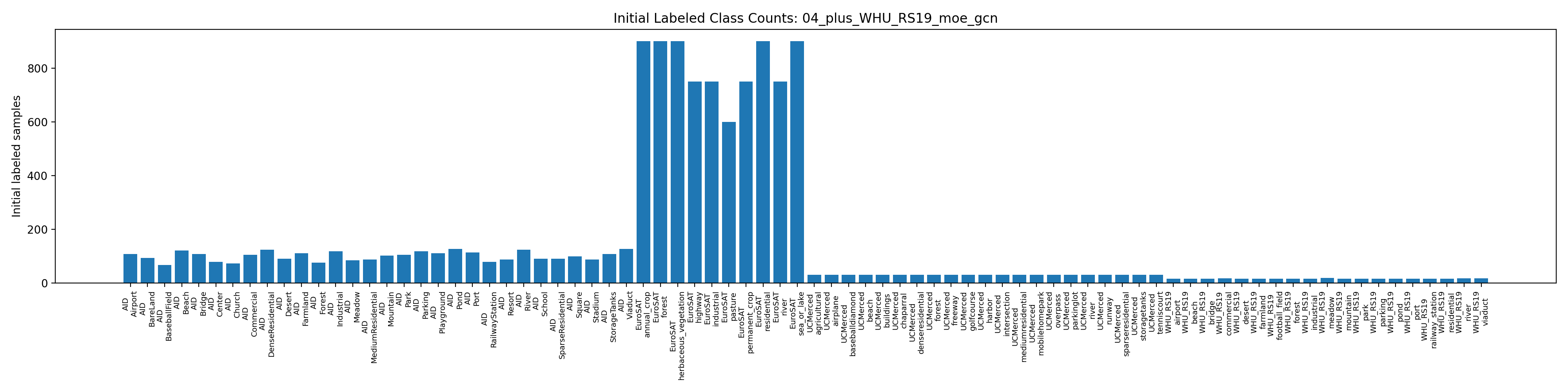}
    \caption{Config 04 (+WHU-RS19).}
  \end{subfigure}
  \caption{Initial labeled pool class distributions. WHU-RS19 (config~04)
  introduces the largest imbalance: some classes receive as few as 15
  initial labeled samples vs.\ over 200 for PatternNet or RESISC45 classes,
  motivating the stratified active-query strategy.}
  \label{fig:labeled_dist}
\end{figure*}

After one round of message passing, each image's representation has been
enriched by the features of its most similar neighbors.
The graph-refined features $Z'$ are passed to a dropout layer (rate~0.3)
followed by a fully connected classification head.
 
\section{Additional Experimental Details}
\label{supp:experiments}
 
\subsection{Dataset Statistics}
\label{supp:datasets}
 
Table~\ref{tab:dataset_stats} provides comprehensive statistics for all
eleven datasets.
The wide size range (1\,005 to 31\,500 images) and class count range
(7 to 45 classes) motivate both the dataset-balanced sampling and the
stratified active-query selection described in the main paper.

\subsection{ResNet-18 Student vs.\ RemoteCLIP}
\label{supp:vs_remoteclip}
 
Table~\ref{tab:r18_vs_rc} compares the best ResNet-18 MoE+GCN student
against RemoteCLIP zero-shot on all core and extended datasets.
The student outperforms RemoteCLIP on six of seven datasets.
WHU-RS19 is the exception: with only 1\,005 images, the foundation model's
zero-shot generalization (94.33\,\%) substantially exceeds the student's
best checkpoint (11.14\,\% accuracy), confirming that data-driven
adaptation is limited when training samples are very scarce.

\subsection{WHU-RS19 Deep Dive}
\label{supp:whu}
 
Figure~\ref{fig:supp_whu_balanced_comparison} analyzes the three model
variants on WHU-RS19.
While the balanced student without MoE/GCN achieves 80.72\,\% Macro-F1,
adding MoE+GCN improves the best checkpoint to 84.81\,\%.
RemoteCLIP retains the lead at 94.33\,\%, confirming that foundation-model
zero-shot generalization is superior when training data is very scarce.
The gap highlights a genuine limitation: dataset-balanced sampling and
active learning improve small-dataset coverage but cannot fully substitute
for richer training data.

\subsection{Per-Class Results: ResNet-18 MoE+GCN Student}
\label{supp:perclass_r18}
 
Tables~\ref{tab:r18_perclass_aid}--\ref{tab:r18_perclass_optimal} report
ResNet-18 MoE+GCN student per-class Precision, Recall, and F1-Score for
all eleven datasets, sorted by F1-Score (descending).
Per-class metrics are derived from the evaluation CSV and reveal which
categories the student finds challenging.

\section{Feature Visualizations}
\label{supp:vis}

\subsection{Hardest Classes per Dataset}
\label{supp:hardclasses}
 
Figure~\ref{fig:worst_classes} shows the lowest-accuracy classes for
selected datasets.
Hard categories consistently involve fine-grained distinctions within a
semantic group (\eg, residential density levels, crop types) or
categories with high inter-class visual overlap.

\subsection{Initial Labeled Pool Distribution}
\label{supp:labeled_dist}
 
Figure~\ref{fig:labeled_dist} shows the class distribution of the initial
labeled pool for selected configurations, illustrating the imbalance that
arises when datasets of very different sizes are combined.

\end{document}